\documentclass{article}

\PassOptionsToPackage{numbers}{natbib}
\usepackage[table,xcdraw]{xcolor}

    \usepackage[preprint]{neurips_2025}

\usepackage[utf8]{inputenc} 
\usepackage[T1]{fontenc}    
\usepackage{hyperref}       
\usepackage{url}            
\usepackage{booktabs}       
\usepackage{amsfonts}       
\usepackage{nicefrac}       
\usepackage{microtype}      
\usepackage{xcolor}         
\usepackage{graphicx}
\usepackage{textcomp}
\usepackage{xspace}
\usepackage{lipsum}
\usepackage{makecell}
\usepackage{multirow}
\usepackage{amsmath}        
\usepackage{rotating} 
\usepackage{subcaption}   
\usepackage{mwe} 

\title{FedRS-Bench: Realistic Federated Learning Datasets and Benchmarks in Remote Sensing}

\author{%
  Haodong Zhao\thanks{Equal contribution.} ,  Peng Peng$^*$, Chiyu Chen, Linqing Huang$^\dagger$, Gongshen Liu\thanks{Corresponding authors.} \\
  Shanghai Jiao Tong University, China \\
\texttt{\{zhaohaodong, nd9pengpeng, 2273202874, huanglinqing, lgshen\}@sjtu.edu.cn} \\
}

\begin{document}

\maketitle

\begin{abstract}
Remote sensing (RS) images are usually produced at an unprecedented scale, yet they are geographically and institutionally distributed, making centralized model training challenging due to data-sharing restrictions and privacy concerns. Federated learning (FL) offers a solution by enabling collaborative model training across decentralized RS data sources without exposing raw data. However, there lacks a realistic federated dataset and benchmark in RS. Prior works typically rely on manually partitioned single dataset, which fail to capture the heterogeneity and scale of real-world RS data, and often use inconsistent experimental setups, hindering fair comparison. To address this gap, we propose a realistic federated RS dataset, termed FedRS. FedRS consists of eight datasets that cover various sensors and resolutions and builds 135 clients, which is representative of realistic operational scenarios. Data for each client come from the same source, exhibiting authentic federated properties such as skewed label distributions, imbalanced client data volumes, and domain heterogeneity across clients. These characteristics reflect practical challenges in federated RS (e.g. hundreds of ground stations or satellites each observing different locales) and support evaluation of FL methods at scale. Based on FedRS, we implement 10 baseline FL algorithms and evaluation metrics to construct the comprehensive FedRS-Bench. The experimental results demonstrate that FL can consistently improve model performance over training on isolated data silos, while revealing performance trade-offs of different methods under varying client heterogeneity and availability conditions. We hope FedRS-Bench will accelerate research on large-scale, realistic FL in RS by providing a standardized, rich testbed and facilitating fair comparisons across future works. The source codes and dataset are available at \href{https://fedrs-bench.github.io/}{https://fedrs-bench.github.io/}.
\end{abstract}

\section{Introduction}
The volume of remote sensing (RS) images from satellites, aircraft, and drones has increased exponentially in recent years~\cite{huang2023evidential,li2024vrsbench,lin2025fedrsclip}. The data deluge provides unprecedented opportunities for applications crucial to societal well-being, including environmental monitoring, climate change analysis, urban planning, precision agriculture, and disaster response~\cite{bastani2023satlaspretrain}. However, RS data are often inherently distributed and reside in archives managed by different government agencies, research institutions, or private companies.
Traditional centralized training would require aggregating terabytes of sensitive images in one location, which is infeasible and often restricted by privacy, security, or policies such as the European Union’s General Data Protection Regulation (GDPR)\footnote{https://gdpr-info.eu/}. Federated learning (FL)~\cite{mcmahan2017communication} has thus emerged as a promising paradigm for RS: multiple parties (e.g., space agencies, governments, or ground stations) collaboratively train a global model by sharing model updates instead of raw images, thereby preserving data locality and privacy. Extensive research has been conducted to optimize FL, aiming to improve communication efficiency and address data heterogeneity~\cite{zhao2023fedprompt,liu2024fedlpa,liao2025privacy,fu2025learn,qi2025cross}.

Despite growing interest in applying FL to the RS field~\cite{zhu2023privacy,zhang2023federated,buyuktacs2023learning,li2024feddiff,buyuktacs2024transformer,duong2024leveraging,ben2024federated,buyuktas2024federated,moreno2024federated,buyuktacs2025multi,li2025safe,klotz2025communication}, a significant hurdle remains: currently there is no realistic standardized federated benchmark for RS data. As shown in Table~\ref{tab:existing-partition}, most prior studies design their own ad-hoc federated experiments by artificially partitioning clients from a single existing RS dataset, such as allocating data based on label distributions using Dirichlet sampling~\cite{wu2025vertical}. For example, \cite{ben2024federated} simulates FL in aerial image classification by partitioning popular datasets (including Optimal-31~\cite{wang2018scene}, UCM~\cite{yang2010bag}, and NWPU-RESISC45~\cite{alosaimi2023self}) among a small number of clients. However, these partitions often do not reflect real operational scenarios—they may assign data randomly or by class to a few clients, which fails to capture key heterogeneity properties, such as geographic domain shifts or highly imbalanced client data volumes. Moreover, different works use different datasets and experimental protocols, making it difficult to compare results or truly assess FL performance in a realistic context.

\begin{table}[t]
\centering
\caption{Summary of datasets and partitions used in existing FL-RS works. Most previous studies design their own ad-hoc experiments by simply splitting clients from a single existing RS dataset, which has limited data sources and clients. There is a gap to real scenarios.}
\label{tab:existing-partition}
\resizebox{\columnwidth}{!}{%
\begin{tabular}{@{}llll@{}}
\toprule
Research    & Dataset                                                                             & Clients                                           & Partition                                                                                                                                                          \\ \midrule
  \cite{buyuktacs2025multi}      & \begin{tabular}[c]{@{}l@{}}BigEarthNet-MM\\ Dynamic World-Expert\end{tabular}       & 14                                                & 7 clients contain Sentinel-1 images, others contain Sentinel-2 images.                                                                                             \\ \midrule
   \cite{klotz2025communication}     & BigEarthNet-S2                                                                      & 8                                                 & Each client hold data from a single country.                                                                                                                       \\ \midrule
FedDiff~\cite{li2024feddiff}  & \begin{tabular}[c]{@{}l@{}}Houston2013\\ Trento\\ MUUFL\end{tabular}                & 8                                                 & Randomly partition.                                                                                                                                                \\ \midrule
    \cite{moreno2024federated}   & RSI-CB                                                                              & 20, 50, 100                                       & Single Dataset Simulation.                                                                                                                                         \\ \midrule
FedPM~\cite{zhang2023federated}   & \begin{tabular}[c]{@{}l@{}}IAIL\\ Bh-Pools\\ GLM\end{tabular}                       & \begin{tabular}[c]{@{}l@{}}5\\ 8\\ 6\end{tabular} & \begin{tabular}[c]{@{}l@{}}Partition in a single dataset. \\
The dataset on each client comes from one region,\\ but all data comes from the same source\end{tabular} \\ \midrule
    \cite{kopidaki2025federated}    & \begin{tabular}[c]{@{}l@{}}AID\\ NWPU-RESISC45\end{tabular}                         & 10                                                & \begin{tabular}[c]{@{}l@{}}Partition in a single dataset. DS1: Balanced Sampling;\\  DS2: Skewed Sampling; DS3: Highly Skewed Sampling.\end{tabular}                \\ \midrule
  \cite{buyuktacs2023learning}      & BigEarthNet-MM                                                                      & 6                                                 & 3 clients contain Sentinel-1 images, others contain Sentinel-2 images.                                                                                             \\ \midrule
    \cite{zhu2023privacy}    & Eurosat                                                                             & 20                                                & Single dataset simulation, balanced partitioning.                                                                                                                  \\ \midrule
SAFE~\cite{li2025safe}    & \begin{tabular}[c]{@{}l@{}}PatternNet\\ NWPU-RESISC45\\ LoveDA\\ WHDLD\end{tabular} & 2, 5, 10, 20                                      & Single dataset simulation, using Dirichlet distribution partitioning.                                                                                            \\ \midrule
      \cite{buyuktacs2024transformer}  & BigEarthNet-S2                                                                      & 7                                                 & DS1: Balanced Sampling; DS2: Each client hold data from a single country.                                                                                          \\ \bottomrule
\end{tabular}%
}
\end{table}

To fill this gap, this paper first constructs a realistic federated dataset---FedRS---to comprehensively measure real-world scenarios using eight real RS datasets. FedRS contains 135 clients, and the core innovation of FedRS is that ``data from the same client comes from the same source". We further extract FedRS-5 from FedRS to accommodate simpler scenarios. For each dataset, we considered two data partitioning schemes and testing criteria to comprehensively evaluate the performance of the model from multiple perspectives. 

Based on the proposed dataset, we build FedRS-Bench to comprehensively measure the performance of commonly used FL algorithms, including FedAvg~\cite{mcmahan2017communication}, FedAvgM~\cite{hsu2019measuring}, FedProx~\cite{fedprox20}, SCAFFOLD~\cite{KarimireddyKMRS20}, FedNova~\cite{WangLLJP20}, FedDyn~\cite{acar2021federated}, FedDC~\cite{gao2022feddc}, MOON \cite{model21}, FedDisco \cite{YeXWXCW23} and user-level differential privacy (DP)~\cite{wei2020performance} on FedAvg (dubbed FedDP). 
Considering the complexity of the FL scenario, we focus on the scalability of the number of clients. We account for client availability variability during training by simulating partial participation: in our experiments, only a small fraction of clients (10 clients) are active in each communication round, which mimics real-world conditions where not all data sources are online or willing to participate at all times. The diversity and scale of FedRS-Bench make it a comprehensive testbed for developing and evaluating FL methods under realistic RS conditions. In particular, FedRS-Bench inherits several representative real-world challenges: (1) \textbf{Statistical heterogeneity}: Clients have vastly different data quantities and class distributions, challenging the performance of FL algorithms. (2) \textbf{Domain shifts}: Imagery from different regions or sensors varies in resolution, spectral bands, and sampling time, testing algorithms' robustness to domain mismatches. (3) \textbf{Scalability and efficiency}: With potentially hundreds of clients and high-dimensional image models, communication and convergence efficiency are critical issues. (4) \textbf{Partial participation}: As client availability is dynamic, methods must cope with varying client subsets per round and even long periods of client inactivity. By providing a benchmark that explicitly includes and measures these factors, FedRS-Bench aims to drive research towards FL techniques truly applicable in large-scale remote sensing deployments.

We then conduct extensive experiments on various FL algorithms and derive insights into FL behavior in realistic RS scenarios. Our experiments consistently show that: (i) Federation provides significant performance gains over training separate models on each client’s data in terms of overall accuracy and generalization, confirming the value of collaborative learning even under privacy constraints. On all clients, the federated model achieves substantially higher mean accuracy across than any local model. (ii) No single FL algorithm excels in all scenarios—the ranking of methods varies with the degree of data heterogeneity and scale. For instance, we observe SCAFFOLD can stabilize training under extreme class imbalance, whereas methods like MOON or naive FedAvg may yield better accuracy when many clients participate concurrently. These results provide the first comparative baseline for FL methods on realistic RS data. Beyond benchmarking, FedRS-Bench opens avenues for new research. Thanks to its rich diversity, researchers can explore advanced topics such as personalized federated models for different geo-clusters, domain adaptation between clients, or efficient federated learning on multimodal sensor data. In summary, our key contributions are as follows.  

$\bullet$ We present FedRS, the first realistic FL dataset for RS. FedRS includes 135 realistically partitioned clients that capture the complexities of heterogeneous remote sensing scenarios in the real world.

$\bullet$ We thoroughly document the construction and characteristics of the FedRS dataset, extract the simpler FedRS-5 dataset from FedRS, and show that it exhibits significant heterogeneous properties and scale, providing a challenging and practical testbed for FL algorithms.

$\bullet$ We benchmark the representative FL methods in FedRS. FedRS-Bench gives insight into the relative strengths of different methods under realistic conditions (heterogeneity, partial client participation) and demonstrates the necessity of federated collaboration for high-performance RS models. 

$\bullet$ FedRS is publicly available along with open source code for data processing and experiment reproducibility. We hope that this benchmark will foster fair evaluation and rapid development of FL techniques that can be deployed at scale in the RS community.
\section{Related Work}
\textbf{Federated learning in remote sensing.} Due to its effectiveness in protecting privacy, FL has been studied and applied in the field of RS. Early works have applied FL to RS in a limited manner: for example, by distributing classical scene classification datasets across a few clients to demonstrate the feasibility of FL on aerial images~\cite{zhu2023privacy,zhang2023federated}. These studies confirmed that FL can train competitive models without centralizing sensitive geospatial data~\cite{buyuktacs2024transformer,li2025safe}. However, they generally considered only small numbers of clients and used simplistic data splits, failing to represent the true scale and diversity of operational remote sensing environments. A concurrent line of research addresses the challenges of heterogeneity and multimodal fusion in FL~\cite{li2024feddiff,buyuktacs2025multi}, but RS evaluations have been limited to proprietary or toy setups.

\textbf{Federated learning benchmarks.} The importance of benchmarks in FL has been recognized. LEAF~\cite{caldas2018leaf} introduce datasets such as FEMNIST and Shakespeare to simulate federated scenarios in mobile devices and encourage comparative studies. Subsequent efforts include FLAIR~\cite{song2022flair} and FLamby~\cite{ogier2022flamby} which provide multi-label image and medical data benchmarks, respectively, under federated settings. Most recently, FedLLM-Bench~\cite{ye2024fedllm} targets large language models, introducing realistic cross-user splits for federated language tasks. FEDMEKI~\cite{wang2024fedmeki} is designed to address the unique challenges of integrating medical knowledge into foundation models under privacy constraints. Pfl-research~\cite{granqvist2024texttt} focuses on the efficiency required to simulate FL on large and realistic FL datasets. These benchmarks highlight the value of covering a range of tasks, client counts, and data heterogeneity factors under a unified framework. To our knowledge, no analogous benchmark existed for RS before FedRS-Bench. A very recent study, FedRSClip~\cite{lin2025fedrsclip}, constructs a federated dataset from three existing RS image sets for evaluation. However, dataset in FedRSClip (Fed-RSIC) is limited due to data source and data partition, which primarily aimed to demonstrate a novel FL algorithm, while FedRS-Bench provides a broader dataset (multiple sources, multiple resolutions, and significantly larger client counts) and is designed as a general benchmark for the community. Compared with prior benchmarks and approaches, FedRS-Bench differentiates itself through its focus on realistic RS scenarios and comprehensive coverage of FL challenges within this domain.

\section{Descriptions of FedRS}

\subsection{Constituent Datasets and Properties}

We briefly reiterate the key characteristics of the eights datasets. The overview is shown in Table~\ref{tab:dataset}. 

$\bullet$ \textbf{NaSC-TG2}~\cite{zhou2021nasc} is a natural scene classification dataset that contains 20,000 images (128x128, 10m spatial resolution) in 10 classes, exclusively sourced from the Chinese Tiangong-2 satellite (TG-2).

$\bullet$ \textbf{WHU-RS19}~\cite{dai2010satellite} is a single region (China) scene classification dataset with 1,005 high-resolution (0.5m) RGB images (600x600) across 19 classes, sourced from Google Earth (circa 2010). Known for fairly uniform characteristics but challenging class similarities.

$\bullet$ \textbf{EuroSAT}~\cite{helber2019eurosat} contains 27,000 Sentinel-2 satellite patches (64x64, 10m spatial resolution) covering 10 land cover classes across Europe. We use the RGB version for compatibility.

$\bullet$ \textbf{AID}~\cite{xia2017aid} is a aerial image dataset for scene classification with 10,000 images (600x600) covering 30 diverse classes. All images are sourced from Google Earth across different countries and times, featuring high diversity (0.5-8m resolution). Each class has 220 to 420 images.

$\bullet$ \textbf{NWPU-RESISC45}~\cite{cheng2017remote} is a scene classification dataset with 31,500 images (256x256) across 45 classes. Collected from Google Earth, it features varying resolutions (0.2-30m).

$\bullet$ \textbf{UCM}~\cite{yang2010bag}, the UC Merced Land Use dataset includes 2,100 aerial images (256x256, 0.3m resolution) evenly covering 21 land use classes from USGS Urban Area Orthoimagery across the US.

$\bullet$ \textbf{OPTIMAL-31}~\cite{wang2018scene} is a VHR scene classification dataset with 1,860 images (256x256, 0.26-8.85m resolution) across 31 classes evenly, sourced from Google Earth. 

$\bullet$ \textbf{RSD46-WHU}~\cite{long2017accurate} is a scene classification dataset that covers 46 categories (500-3000 images per class). Sourced from Google Earth and Tianditu (0.5-2m resolution, typically 256x256).

\begin{table}[!htbp]
\centering
\caption{Information and client partitioning of the original RS datasets used to build the FedRS.}
\label{tab:dataset}
\resizebox{\textwidth}{!}{%
\begin{tabular}{@{}lllllll@{}}
\toprule
Dataset       & \makecell[l]{Image \\ resolution} & \makecell[l]{Spatial \\resolution} & Classes & Images & Clients & Source Characteristics \\ \midrule
NaSC-TG2      & 128×128          & 10                     & 10        & 20,000   &     34      & Single source (China, TG-2)                       \\
WHU-RS19      & 600×600          & 0.5                    & 19        & 1,005    &     1      &  Single region (China)                      \\
EuroSAT  & 64×64            & 10                     & 10        & 27,000   &     45      & European Satellite Imagery                       \\
AID           & 600×600          & 0.5$\sim$8                  & 30        & 10,000   &     5      & Multi-source diversity                       \\
NWPU-RESISC45 & 256×256          & 0.2$\sim$30                 & 45        & 31,500   &      12     & Multi-source diversity                       \\
UCM           & 256×256          & 0.3                    & 21        & 2,100    &     2      & Single region (USGS)                       \\
OPTIMAL-31    & 256×256          & 0.26$\sim$8.85              & 31        & 1,860    &     1      &  Google Earth Imagery
                      \\
RSD46-WHU     & 256×256          & 0.5$\sim$2                  & 46        & 34,549   &     35      &  Multi-source diversity                      \\ \bottomrule
\end{tabular}
}
\end{table}

\begin{figure}[!htbp] 
    \centering 
    \begin{subfigure}[b]{0.19\textwidth} 
        \centering
        \includegraphics[width=\linewidth]{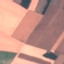} 
        \caption{Agriculture}
    \end{subfigure}
    \hfill 
    \begin{subfigure}[b]{0.19\textwidth}
        \centering
        \includegraphics[width=\linewidth]{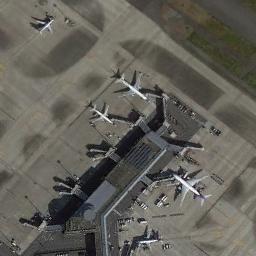} 
        \caption{Airport}
    \end{subfigure}
        \hfill 
    \begin{subfigure}[b]{0.19\textwidth} 
        \centering
        \includegraphics[width=\linewidth]{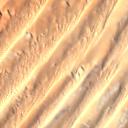} 
        \caption{Bareland}
    \end{subfigure}
            \hfill 
    \begin{subfigure}[b]{0.19\textwidth} 
        \centering
        \includegraphics[width=\linewidth]{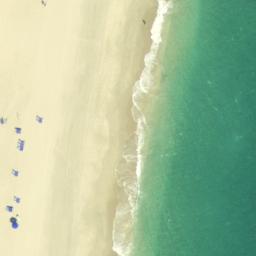} 
        \caption{Beach}
    \end{subfigure}
                \hfill 
    \begin{subfigure}[b]{0.19\textwidth} 
        \centering
        \includegraphics[width=\linewidth]{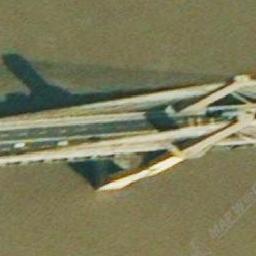} 
        \caption{Bridge}
    \end{subfigure}
    \begin{subfigure}[b]{0.19\textwidth} 
        \centering
        \includegraphics[width=\linewidth]{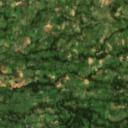} 
        \caption{Forest}
    \end{subfigure}
    \hfill 
    \begin{subfigure}[b]{0.19\textwidth} 
        \centering
        \includegraphics[width=\linewidth]{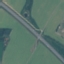}
        \caption{Highway}
    \end{subfigure}
    \hfill 
    \begin{subfigure}[b]{0.19\textwidth} 
        \centering
        \includegraphics[width=\linewidth]{figures/samples15/Industrial.jpg}
        \caption{Industrial}
    \end{subfigure}
    \hfill 
        \begin{subfigure}[b]{0.19\textwidth} 
        \centering
        \includegraphics[width=\linewidth]{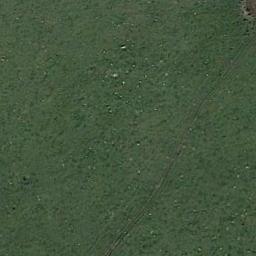}
        \caption{Meadow}
    \end{subfigure}
    \hfill 
        \begin{subfigure}[b]{0.19\textwidth} 
        \centering
        \includegraphics[width=\linewidth]{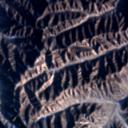}
        \caption{Mountain}
    \end{subfigure}
        \begin{subfigure}[b]{0.19\textwidth} 
        \centering
        \includegraphics[width=\linewidth]{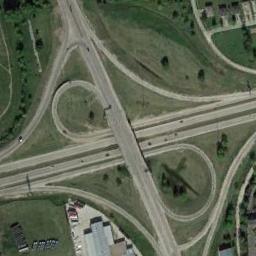}
        \caption{Overpass}
    \end{subfigure}
    \hfill 
        \begin{subfigure}[b]{0.19\textwidth} 
        \centering
        \includegraphics[width=\linewidth]{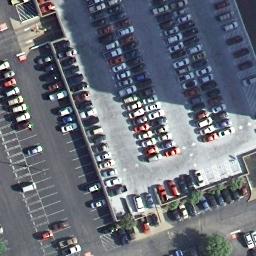}
        \caption{Parkinglot}
    \end{subfigure}
    \hfill 
            \begin{subfigure}[b]{0.19\textwidth} 
        \centering
        \includegraphics[width=\linewidth]{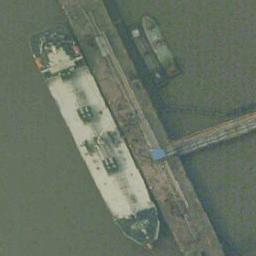}
        \caption{Port}
    \end{subfigure}
    \hfill 
            \begin{subfigure}[b]{0.19\textwidth} 
        \centering
        \includegraphics[width=\linewidth]{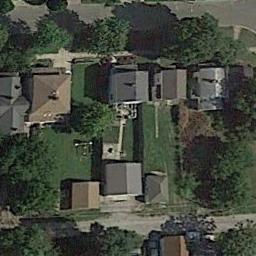}
        \caption{Residential}
    \end{subfigure}
    \hfill 
            \begin{subfigure}[b]{0.19\textwidth} 
        \centering
        \includegraphics[width=\linewidth]{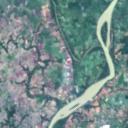}
        \caption{River}
    \end{subfigure}
    \caption{Samples of FedRS.}
    \label{fig:category-samples} 

\end{figure}

These eight datasets do not have overlapping data. They have many differences in composition and cannot be directly divided. (i) Table~\ref{tab:dataset} shows that the original image resolutions of these datasets are different. (ii) As shown in Table~\ref{tab:label-map}, there are inconsistencies in the granularity of the category labels between different datasets, including overlap and ambiguity. We solve the first challenge by unifying all images to a resolution of 64x64 when pre-processing. We carefully check and establish a unified classification scheme with 15 semantic categories designed to harmonize the varied content and annotations present in the source datasets. Table~\ref{tab:label-map} shows the label mapping relationship that distinct labels from the original datasets corresponding to similar semantic concepts are mapped into a single FedRS category. For example, the unified \textit{Agriculture} category in FedRS encompasses images originally labeled as \textit{Rectangular Farmland}, \textit{Farmland}, \textit{Annual Crop} and \textit{Permanent Crop}, depending on the source datasets. Some samples of the 15 semantic categories are shown in Figure~\ref{fig:category-samples}.

\begin{figure}[!htbp] 
    \centering 

    \begin{subfigure}[b]{0.31\textwidth} 
        \centering
        \includegraphics[width=\linewidth]{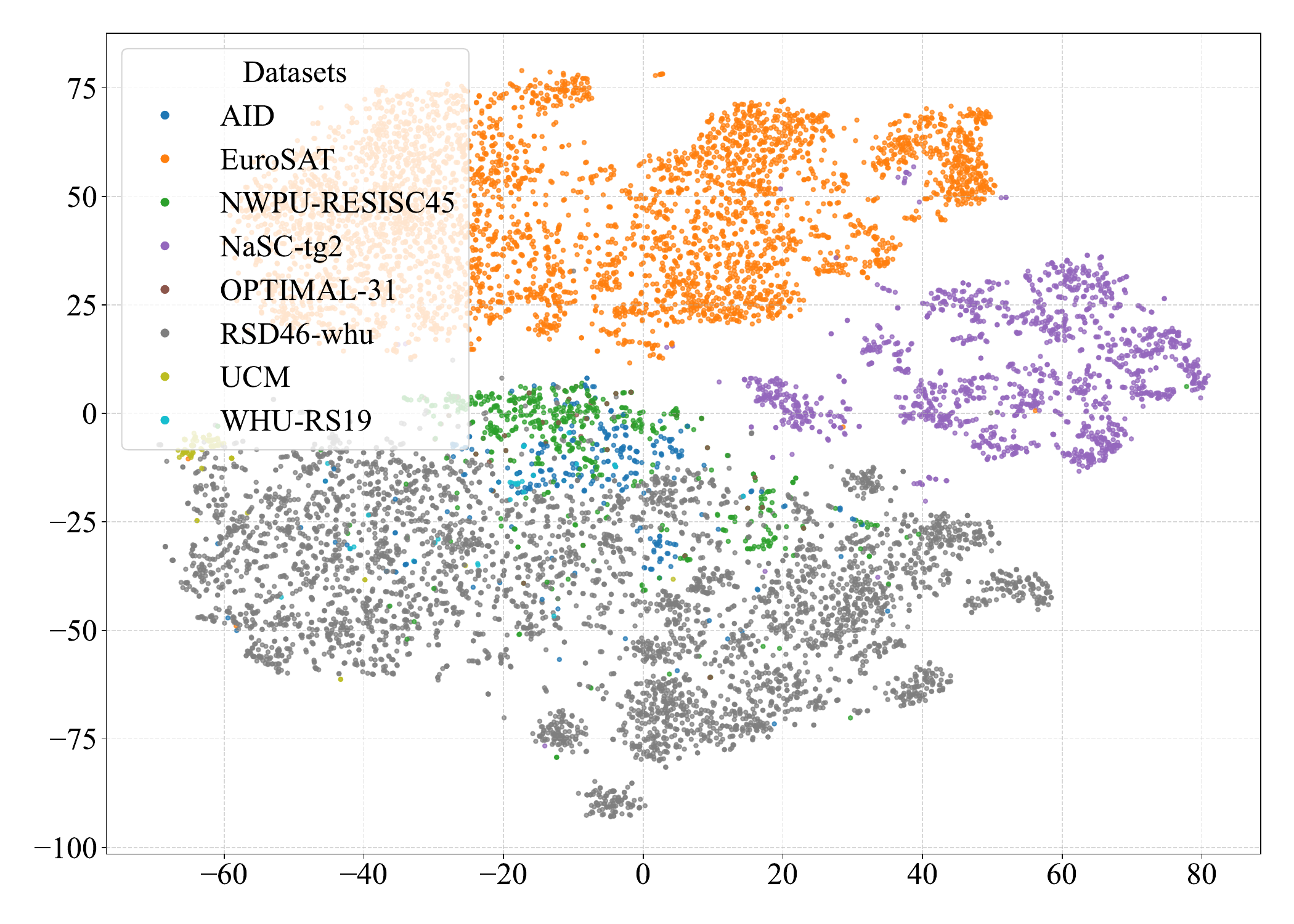} 
        \caption{Agriculture (0.07)}
        \label{fig:subfig1} 
    \end{subfigure}
    \hfill 
    \begin{subfigure}[b]{0.31\textwidth}
        \centering
        \includegraphics[width=\linewidth]{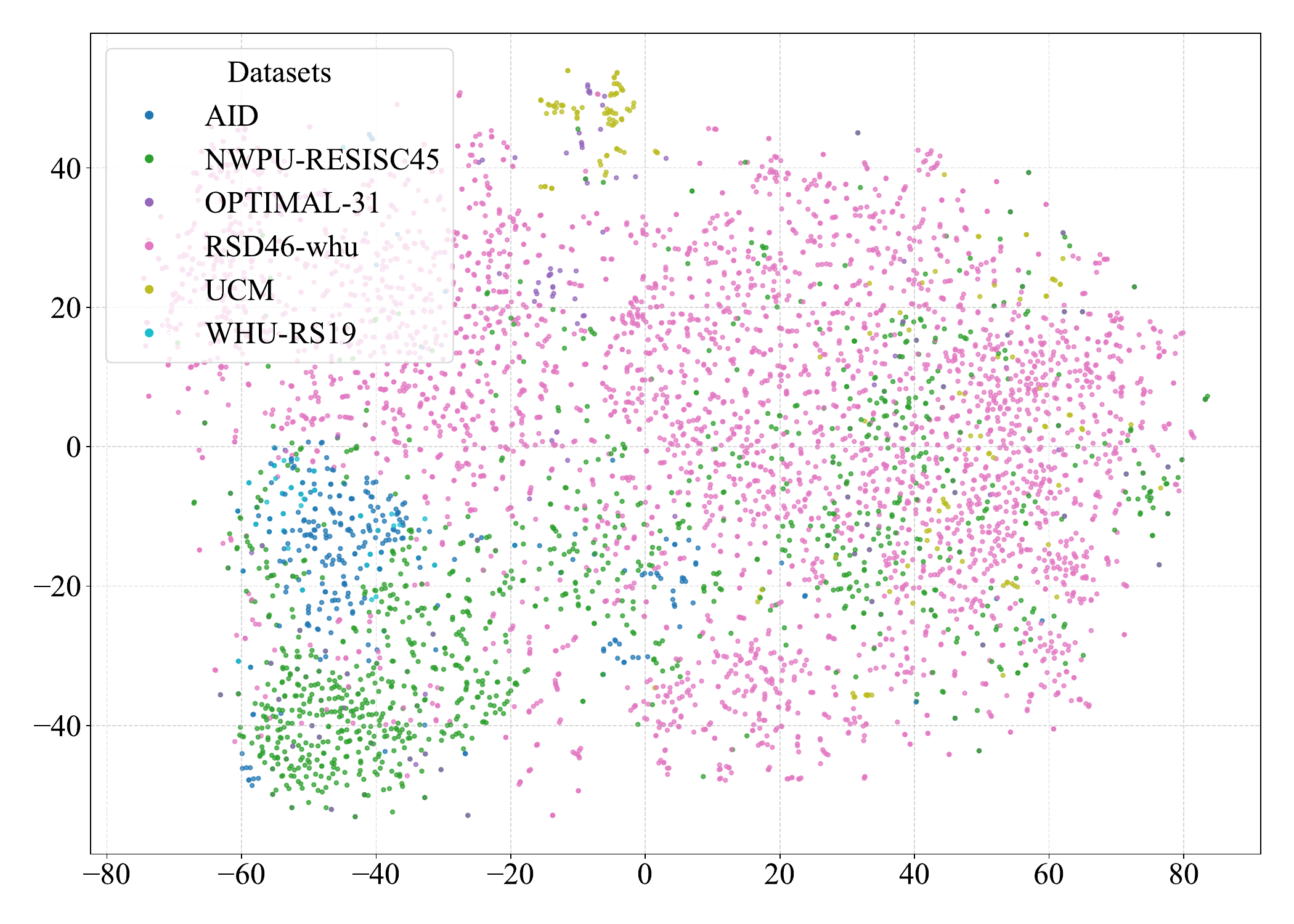} 
        \caption{Airport (-0.21)}
        \label{fig:subfig2} 
    \end{subfigure}
        \hfill 
    \begin{subfigure}[b]{0.31\textwidth} 
        \centering
        \includegraphics[width=\linewidth]{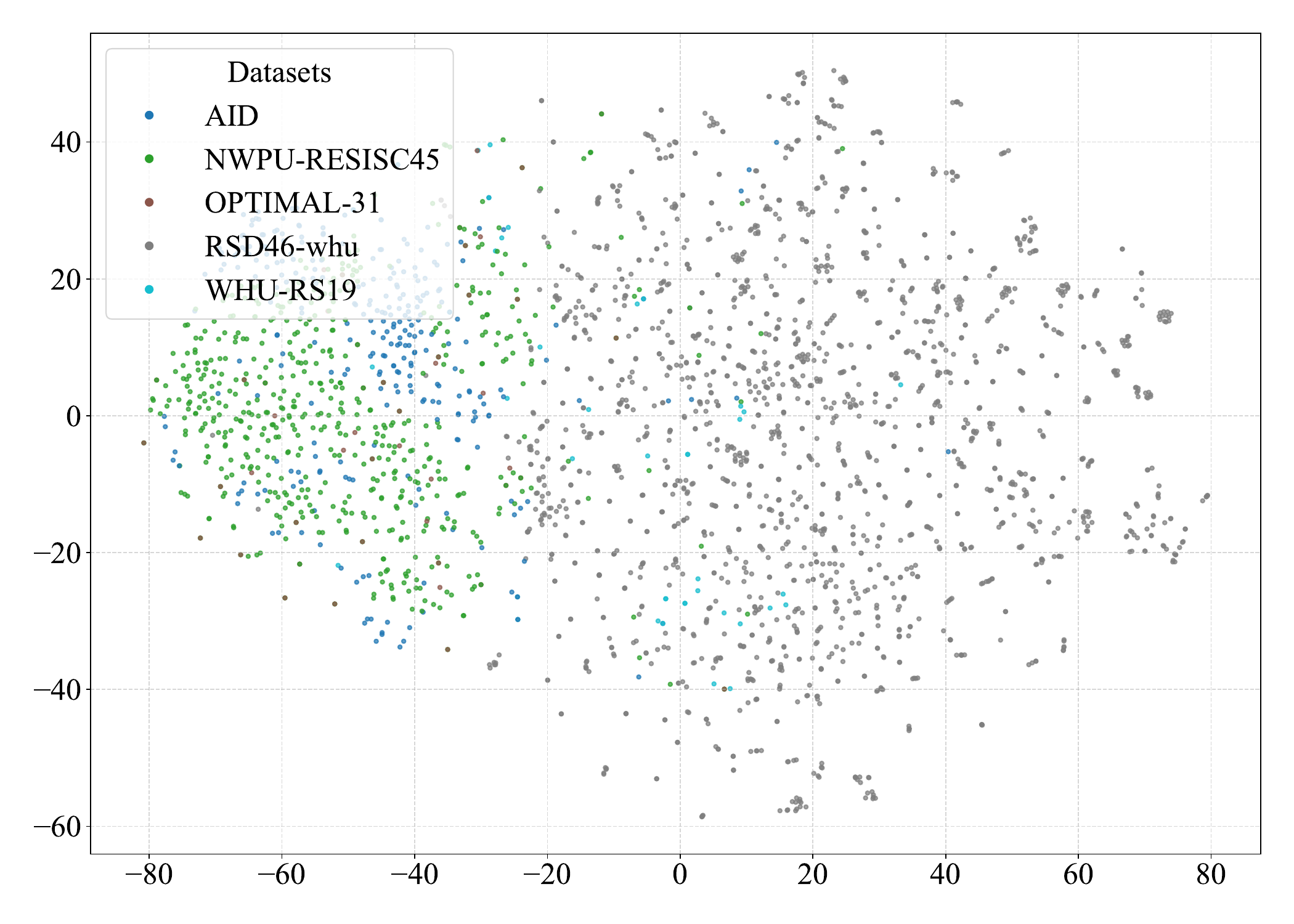} 
        \caption{Bridge (0.06)}
        \label{fig:subfig3} 
    \end{subfigure}
    \begin{subfigure}[b]{0.31\textwidth} 
        \centering
        \includegraphics[width=\linewidth]{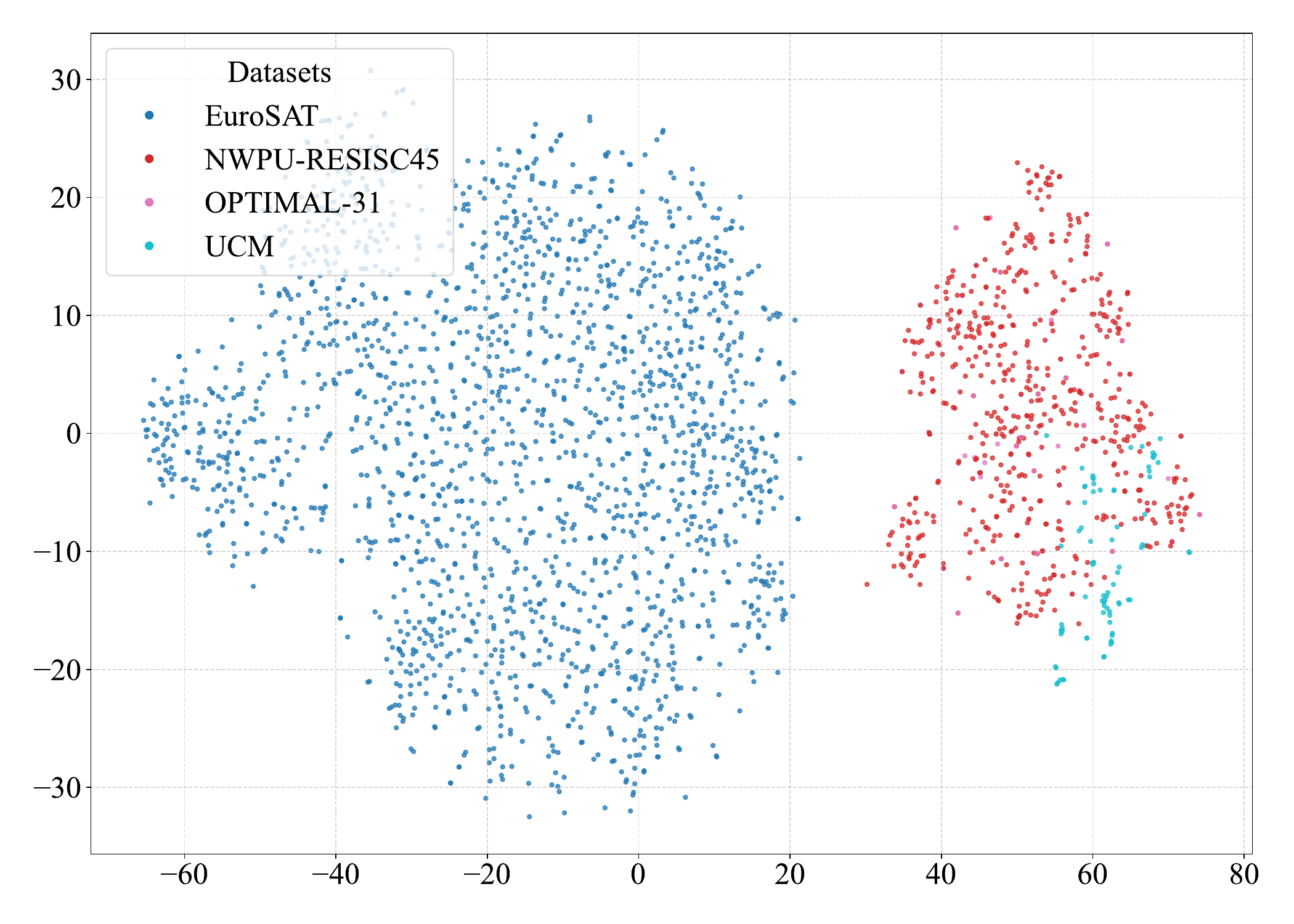} 
        \caption{Highway (0.37)}
        \label{fig:subfig4} 
    \end{subfigure}
    \hfill 
    \begin{subfigure}[b]{0.31\textwidth} 
        \centering
        \includegraphics[width=\linewidth]{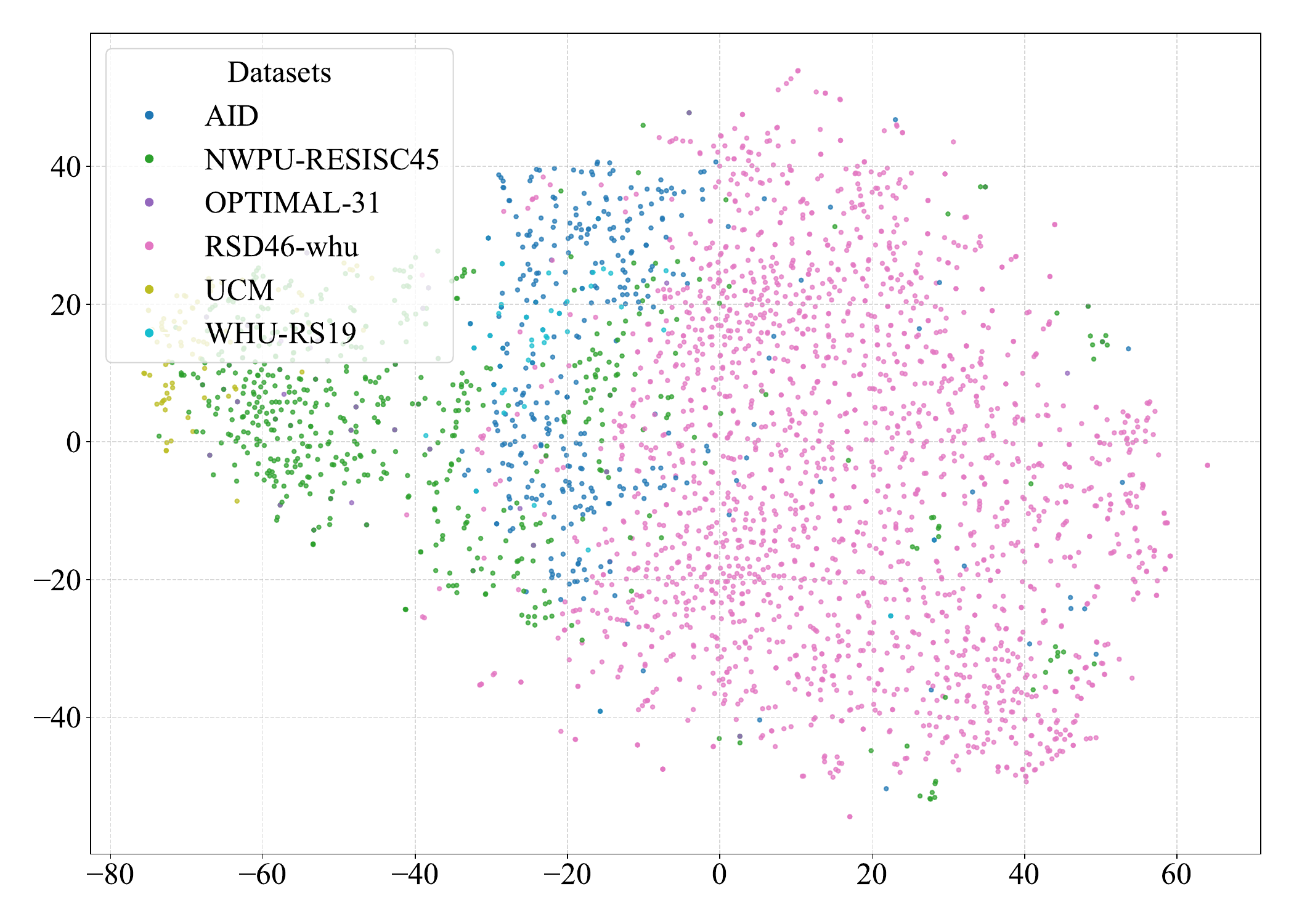} 
        \caption{Overpass (0.03)}
        \label{fig:subfig5} 
    \end{subfigure}
        \hfill 
    \begin{subfigure}[b]{0.31\textwidth} 
        \centering
        \includegraphics[width=\linewidth]{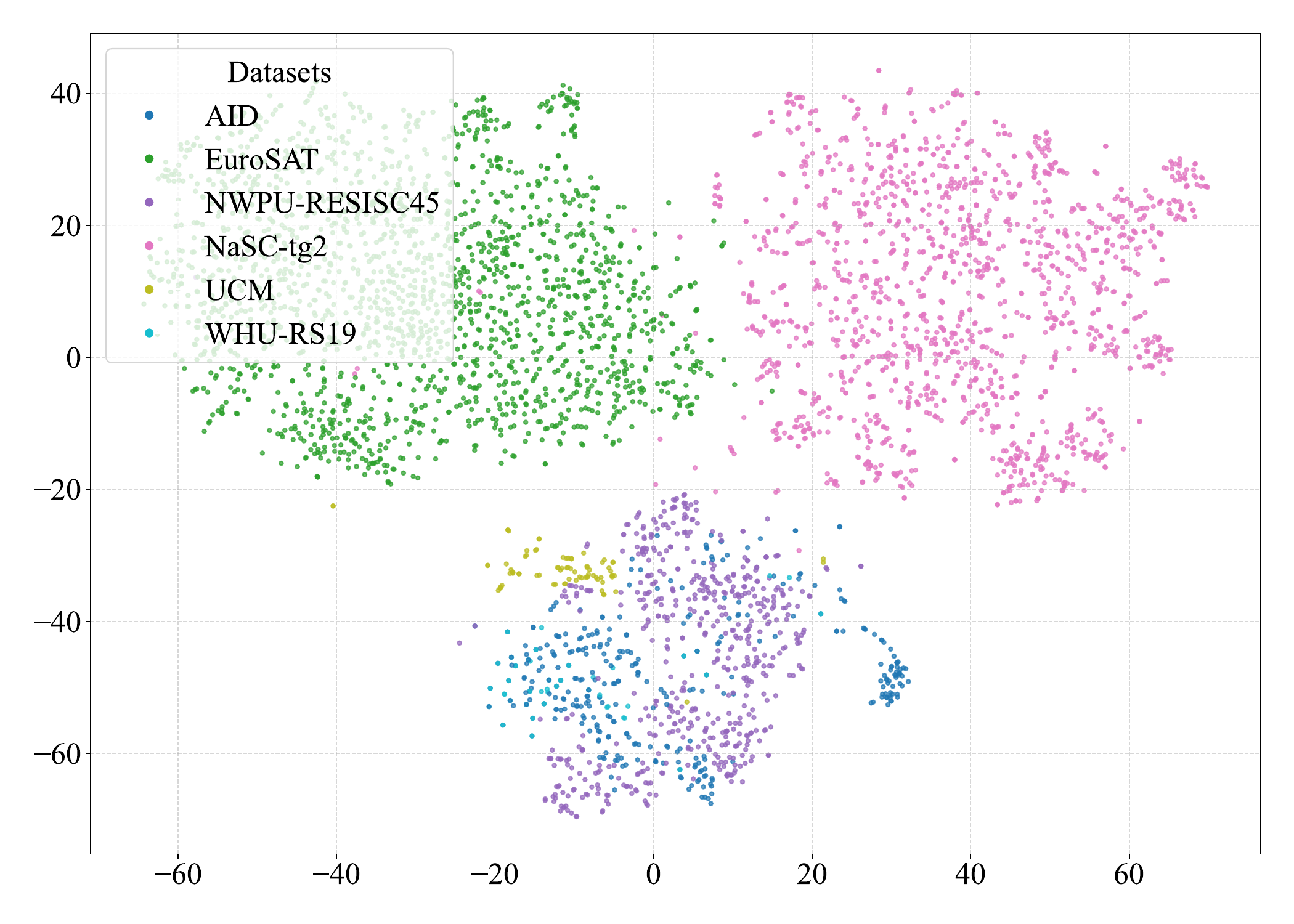} 
        \caption{River (0.33)}
        \label{fig:subfig6}
    \end{subfigure}
    \caption{Visualization of features of from different sources using t-SNE. The corresponding categories and silhouette scores~\cite{shahapure2020cluster} after dimensionality reduction are shown in the subtitles, reflecting the intra-category feature deviation from different sources. Please refer to Figure~\ref{fig:tsne_sources_all} for all results.}
    \label{fig:tsne_sources} 
\end{figure}

Figure~\ref{fig:tsne_sources} shows that even for the same category, there are obvious differences between images from different sources in quantity and feature distribution, which is more in line with the real scenario, that is, different clients often mean different regions, devices, and periods.

\subsection{Federated Training Data Partitioning (135 Clients)}
The core of building a federated RS dataset is how we partition these datasets into distinct clients to emulate real-world data silos. Our key principle is that \textbf{ each client should have data from a single ``source''}, where a source is defined in a way that reflects a coherent distribution (the same sensor, region, etc.). This ensures intra-client consistency and inter-client heterogeneity. For datasets with a single source and limited samples (WHU-RS19, UCM and OPTIMAL-31), we use its data to build one or two clients. For datasets with more samples and sources, we build multiple clients using a dataset, and avoid mixing disparate classes on one client to keep the``single source'' concept. For datasets that are inherently multi-source, we try to use known metadata, including regions/cities/countries to split by source. 
\begin{figure}[!tp] 
    \centering 
    \begin{subfigure}[b]{0.48\textwidth} 
        \centering
        \includegraphics[width=\linewidth]{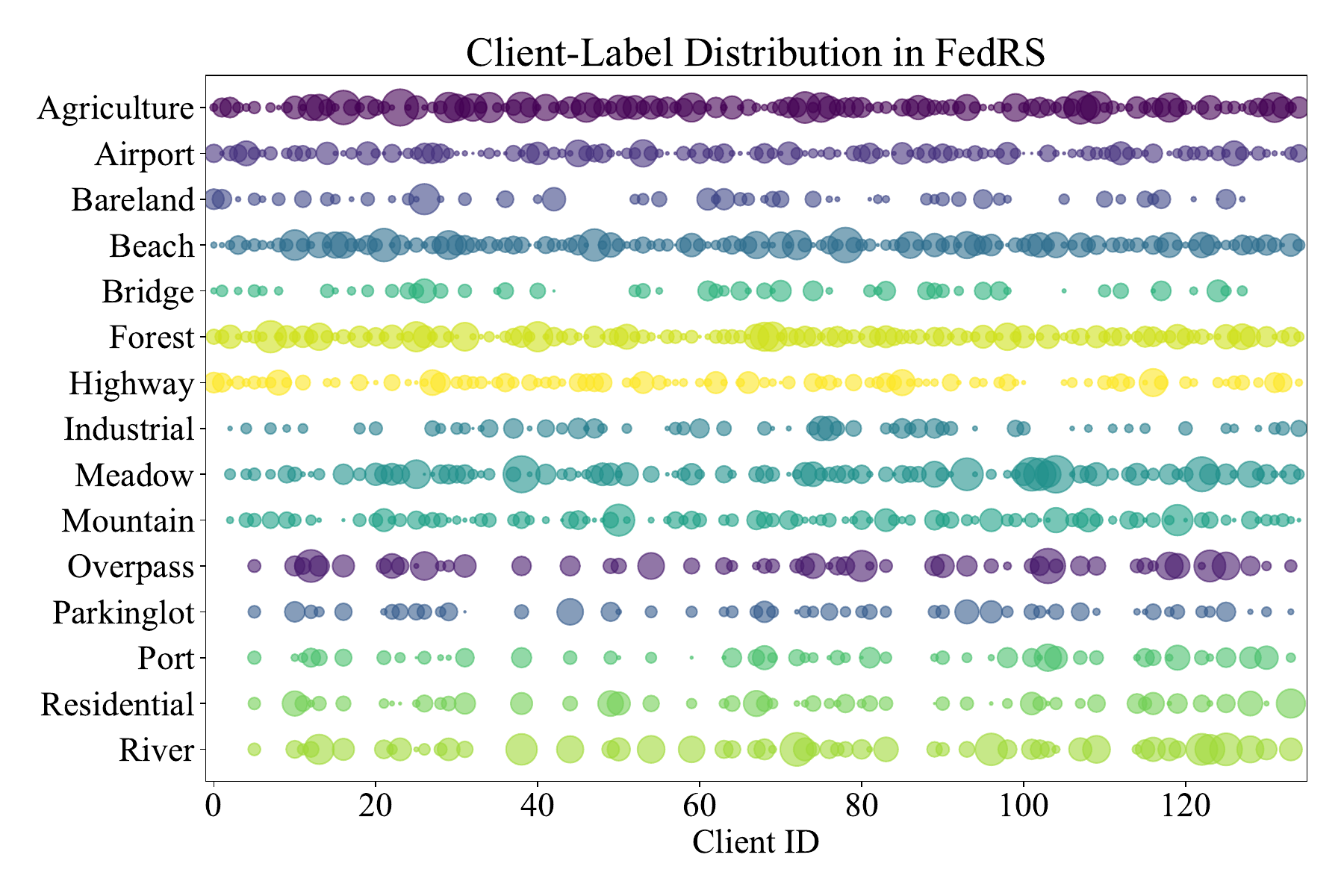} 
        \caption{Category distribution between different clients}
        \label{fig:label} 
    \end{subfigure}
        \hfill 
    \begin{subfigure}[b]{0.48\textwidth} 
        \centering
        \includegraphics[width=\linewidth]{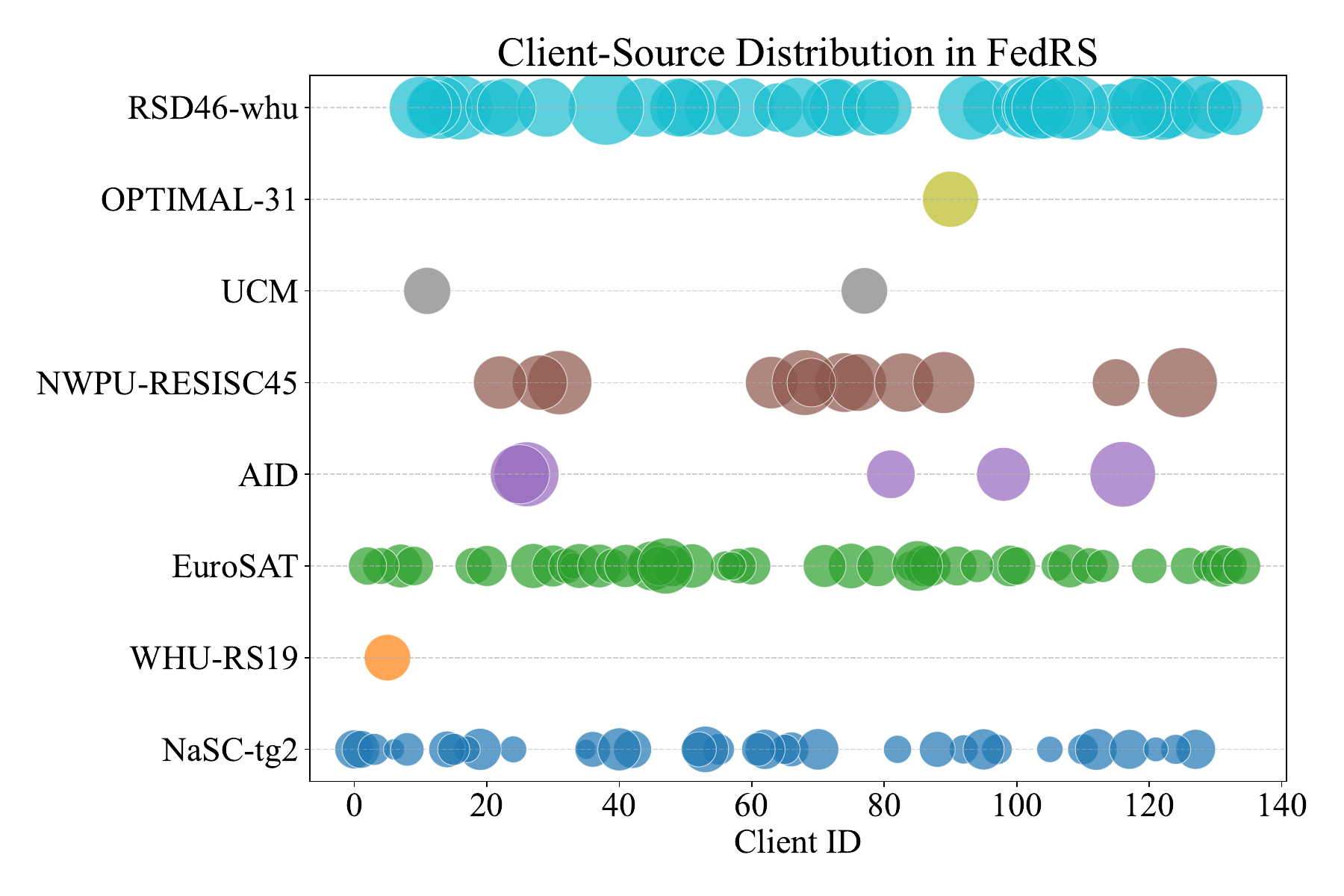} 
        \caption{Source distribution between different clients}
        \label{fig:source}
    \end{subfigure}
    \caption{Data distribution between different clients under NIID-1 partition. (a) Category distribution, each client only has a part of the 15 categories; (b) Source distribution, each client's data comes from only one source, and the number of clients contributed by each source varies greatly.}
    \label{fig:distribution} 
\end{figure}

As shown in Figure~\ref{fig:distribution}, we obtain 135 clients in total. Among these clients, EuroSAT contributes 45 clients, AID 15 clients, RSD46-WHU 35 clients, NaSC-TG2 34 clients, NWPU-RESISC45 12 clients, AID 5 clients, UCM 2 clients and the remaining 2 clients are from WHU-RS19 and OPTIMAL-31. Each client has a non-overlapping subset of the data. The number of images per client ranges widely, from as few as 114 to 1559, mirroring the unbalanced nature of real federated data.

Considering the complexity of real-world scenarios, we try to fully simulate data partitioning under different heterogeneities and comprehensively measure the performance of existing algorithms when faced with data distributions of different difficulties. Therefore, based on the 15 unified semantic categories, we first construct \textbf{FedRS} with complete 15 data categories. We then select five categories with relatively balanced and sufficient data and construct a simple \textbf{FedRS-5} dataset, including \textit{Agriculture, Bareland, Residential, River and Forest}. Under each division, we further divide it into two non independent and identically distributions, \textbf{NIID-1} and \textbf{NIID-2}, based on how to divide the data from each source to the client. NIID-1 uses Dirichlet distribution for data partitioning, and NIID-2 uses uniform partitioning. As shown in Figure~\ref{fig:tsne-clients}, there is a clear shift in characteristics between different clients. 

\subsection{Training Datasets Statistics}
Notably, the data distribution on each client is significantly skewed relative to the global distribution. Many clients possess data for only a subset of the total classes. For example, clients derived from EuroSAT only contain images of 8 classes, and clients derived from NaSC-TG2 only contain images of 7 classes. This means that the global classification task (with 15 unique classes) faces serious label distribution heterogeneity: most clients do not have examples of all classes. Moreover, the feature distributions differ significantly. For example, EuroSAT has a much coarser spatial resolution (10m per pixel for RGB) compared to aerial photos in UCM (0.3m per pixel). Additionally, the resolution of images varies across sources: EuroSAT images are 64×64, while UCM images are 256×256. As a result, a forest in EuroSAT appears markedly different from a forest in UCM, NaSC-TG2, and other source datasets. This domain shift is a critical aspect of FL-RS and a global model must address these differences effectively.

\section{Experiments}
\begin{table}[]
\centering
\caption{Experimental evaluation on the FedRS dataset using CNN and ResNet18. FL methods generally perform better than local training. All results are shown in percentages (\%). \textbf{Bold} and \underline{underline} values highlight the best and and second-best accuracy, respectively.}
\label{tab:main}
\resizebox{\columnwidth}{!}{%
\begin{tabular}{l|l|cccc|c|cccc|c}
\toprule
\multirow{3}{*}{Model}   & Dataset  & \multicolumn{5}{c|}{FedRS}                                                                 & \multicolumn{5}{c}{FedRS-5}                                                                \\ \cmidrule(l){2-12} 
                         & Training & \multicolumn{2}{c}{NIID-1}                  & \multicolumn{2}{c|}{NIID-2}  &  \multirow{2}{*}{Average}               & \multicolumn{2}{c}{NIID-1}                  & \multicolumn{2}{c|}{NIID-2}                  & \multirow{2}{*}{Average}                            \\ \cmidrule(lr){2-6} \cmidrule(lr){8-11}
                         & Testing  & $T_I$                 & $T_B$                 & $T_I$                 & $T_B$   &               & $T_I$                 & $T_B$                 & $T_I$                 & $T_B$                  &                          \\ \midrule
\multirow{12}{*}{CNN}  & Centralized    & 62.20                & 57.34                & 62.20                & 57.34 & \cellcolor{gray!10}59.77$\pm$2.81                & 80.31                & 71.58                & 80.31                & 71.58                 & \cellcolor{gray!10}75.95$\pm$5.04                  \\  

& Local Avg    & 13.86                & 11.91               & 15.82                & 12.74 & \cellcolor{gray!10}13.58$\pm$1.69                &    27.50             & 26.19                & 31.40                & 28.81                 & \cellcolor{gray!10}28.48$\pm$2.22                    \\ \cmidrule(l){2-12} 
                         & FedAvg   & \underline{54.94}                     &  \textbf{57.44}                    & 55.70                     &  57.55  & \cellcolor{gray!10}\underline{56.41}$\pm$1.29                    &  \underline{75.56}                    & \underline{75.87}                     &        75.22              &  \textbf{78.55}                     &  \cellcolor{gray!10}\textbf{76.30}$\pm$1.52                         \\
                         & FedAvgM  & 54.01                     & \underline{57.13}                     &              56.02        &   56.27       & \cellcolor{gray!10}55.86$\pm$1.32             &  74.82                    &           74.80           &  \underline{75.85}                    &   77.50                    & \cellcolor{gray!10}75.74$\pm$1.27                         \\
                         & FedProx  & 54.48                     &  55.96                    &  56.18                    &  \underline{57.87}       & \cellcolor{gray!10}56.12$\pm$1.39               & 75.41                     &  74.80                    &       75.41               &  77.21                     &  \cellcolor{gray!10}75.71$\pm$1.04                        \\
                         & SCAFFOLD & 53.08                     & 48.40                     & 54.69                     &  48.62 & \cellcolor{gray!10}51.20$\pm$3.17                     &  73.73                    & 69.17                     &       \textbf{79.71}               &  76.14                     & \cellcolor{gray!10}74.69$\pm$4.42                         \\
                         & FedDC    & 52.47                     & 52.98                     & 53.27                     & 52.13  & \cellcolor{gray!10}52.71$\pm$0.51                     &  73.74                    &  73.99                    &       74.94               &  76.94                     & \cellcolor{gray!10}74.90$\pm$1.45                         \\
                         & FedDyn   & \textbf{55.95}                     & 55.96                     & \underline{56.20}                     & 56.91  & \cellcolor{gray!10}56.26$\pm$0.45                  & 75.39                     &  \underline{75.87}                    &        75.67              &  \underline{77.75}                     & \cellcolor{gray!10}\underline{76.17}$\pm$1.07                         \\
                         & FedNova  & 54.46                     & 54.36                     &               56.12       &   56.81  & \cellcolor{gray!10}55.44$\pm$1.22                  & 74.74                     &            75.60          &  74.94                    &  75.07                     & \cellcolor{gray!10}75.23$\pm$0.29                         \\
                         & MOON     & 54.86                     & \underline{57.13}                     & \textbf{56.44}                     & \textbf{58.19} & \cellcolor{gray!10}\textbf{56.66}$\pm$1.40                      &  \underline{75.56}                    & \textbf{77.50}                     &      74.71                &  75.60                     &  \cellcolor{gray!10}75.84$\pm$1.18                        \\
                         & FedDisco & 54.80                     & \underline{57.13}                     &  55.90                    & 56.28         & \cellcolor{gray!10}56.13$\pm$0.97              &  \textbf{76.10}                    &  75.07                    &        75.59              &   76.68                    & \cellcolor{gray!10}75.86$\pm$0.69                                           \\
                        & FedDP & 45.68                     & 39.68                     & 47.78                     & 44.15  & \cellcolor{gray!10}44.32$\pm$3.43                     &  74.52                    &               70.51       &  74.98                    &   72.92                    & \cellcolor{gray!10}73.23$\pm$2.02                                           \\
                         \midrule
\multirow{12}{*}{ResNet18} & Centralized    & 79.18                & 75.32                & 79.18                & 75.32  & \cellcolor{gray!10}77.25$\pm$2.23              & 83.73                & 78.28                & 83.73                & 78.28                 & \cellcolor{gray!10}81.01$\pm$3.15                    \\    
& Local Avg   & 20.57                & 18.07                & 23.14                & 19.54 & \cellcolor{gray!10}20.33$\pm$2.14               & 37.20                & 37.20                & 40.54                & 40.30                 & \cellcolor{gray!10}38.81$\pm$1.86                    \\ \cmidrule(l){2-12} 
                         & FedAvg   & 62.26                     & 66.49                     &  62.71                    & 66.17  & \cellcolor{gray!10}64.41$\pm$2.23                   &          73.48            &  72.92                    & 73.77                     & 75.87                      & \cellcolor{gray!10}74.01$\pm$1.29                         \\
                         & FedAvgM  &  61.65                    &  66.49                    &             \underline{63.55}         &  66.17  & \cellcolor{gray!10}\underline{64.47}$\pm$2.29                  & 73.14                     &         74.26             & 74.09                     &  75.34                     & \cellcolor{gray!10}74.21$\pm$0.90                         \\
                         & FedProx  &  \underline{62.87}                    & \underline{67.45}                     &  63.31                    &  64.47    & \cellcolor{gray!10}64.26$\pm$2.06                &        72.91              &  \underline{76.41}                    & 73.87                     &  \underline{76.94}                     & \cellcolor{gray!10}75.03$\pm$1.95                         \\
                         & SCAFFOLD &  \textbf{68.76}                    & \textbf{71.06}                     &  \textbf{68.22}                    &  \textbf{71.60}  & \cellcolor{gray!10}\textbf{69.91}$\pm$1.67                   &        \textbf{74.20}              &   \textbf{77.50}                   & \textbf{76.38}                     & \textbf{78.28}                      & \cellcolor{gray!10}\textbf{76.59}$\pm$1.77                         \\
                         & FedDC    & 10.87                     &  13.19                    &        10.71              &  10.96 & \cellcolor{gray!10}11.43$\pm$1.18                    &                  39.15    &  51.47                    & 38.88                     &  45.31                     & \cellcolor{gray!10}43.70$\pm$5.97                         \\
                         & FedDyn   & 15.53                     &  14.79                    & 15.21                     &  15.43  & \cellcolor{gray!10}15.24$\pm$0.33                 &       45.65               &  54.16                    & 38.47                     &   46.11                    & \cellcolor{gray!10}46.10$\pm$6.41                         \\
                         & FedNova  & 62.11                     &  64.89                    &             63.18         &  65.32  & \cellcolor{gray!10}63.88$\pm$1.50                 &  72.67                    &         75.87             &  \underline{74.13}                    &  76.41                     & \cellcolor{gray!10}74.77$\pm$1.70                         \\
                         & MOON     &  62.31                    &  66.28                    & 62.75                     &  65.32  & \cellcolor{gray!10}64.17$\pm$1.94                   &         \underline{73.59}             &  76.14                    & 73.99                     &  76.68                     & \cellcolor{gray!10}\underline{75.10}$\pm$1.54                         \\
                         & FedDisco & 62.84                     &  64.68                    &       63.18               &  \underline{66.60}  & \cellcolor{gray!10}64.33$\pm$1.71                   &                   72.92   &   73.19                   & 73.24                     &   74.26                    & \cellcolor{gray!10}73.40$\pm$0.59                                                \\ 
                        & FedDP & 52.82                     & 50.85                     &       53.88               &   54.89 & \cellcolor{gray!10}53.11$\pm$1.73                    &  70.89                    &             71.58         &  72.49                    &  72.39                     & \cellcolor{gray!10}71.84$\pm$0.75                                           \\ \bottomrule
\end{tabular}%
}
\end{table}

\subsection{Experimental Setup}
\textbf{Metric.}
We evaluate the accuracy on the global test set. 
To establish a benchmark that rigorously evaluates model performance against realistic data distribution shifts, we constructed two distinct test sets derived from the original test sets. Recognizing the significant variations in category representation across different data sources present in training data, which is a common real-world challenge, our test set is designed to assess performance under different distributional assumptions. The first imbalanced test set termed ${T}_{I}$ is sampled to reflect the proportional representation of categories of each original source found in the original test data, thus reflecting naturally occurring imbalances. Specifically, we extracted 20\% from each source dataset to compose ${T}_{I}$. For example, ${T}_{I}$ contains UCM images (300 images, 15 classes, 20 per class), EuroSAT images (4,800 images, 7 classes, 400–1,000 per class) from the EuroSAT. The second test set, termed $T_{B}$, is a balanced test set constructed by uniformly sampling exactly 10 instances for each category from every original data source represented in the test split. This balanced configuration allows for an assessment of model performance across all category-source combinations, mitigating the influence of varying sample sizes present in the imbalanced set. Together, these test sets facilitate a comprehensive evaluation of model generalization capabilities under heterogeneous data conditions. In addition to FL training, we also perform local training on each client using their private data (dubbed ``Local Training'') and merge all client training data for centralized training (dubbed ``Centralized Training'') to compare with the results of federated learning.

\textbf{FL settings.} Among the 135 clients, we sample 10 clients for training each round, which mimics real-world conditions where not all parties are online or willing to participate at all times due to network, energy and computing power. A total of 500 rounds of global training are conducted.

\textbf{Implementation details.} The number of local epochs and batch size are 3 and 32, respectively. We use a simple CNN network~\cite{model21} and ResNet18~\cite{he2016deep} (training from scratch) as the model. We use SGD optimizer with a learning rate of 0.005. 

\textbf{Hyper-parameters.} Following~\cite{YeXWXCW23}, we use the best hyper-parameter for each method: $\beta = 0.5$ in FedAvgM~\cite{hsu2019measuring}, $\mu = 0.01$ in FedProx~\cite{fedprox20}, $\alpha = 0.01$ in FedDyn~\cite{acar2021federated}, $\mu = 0.1$ in MOON~\cite{model21}, $\alpha = 0.01$ in FedDC~\cite{gao2022feddc}, $a = 0.5$ and $b = 0.1$ in FedDisco~\cite{YeXWXCW23}.

\subsection{FedRS-Bench Results}

\begin{figure}[!tp] 
    \centering 
    \begin{subfigure}[b]{0.48\textwidth} 
        \centering
        \includegraphics[width=\linewidth]{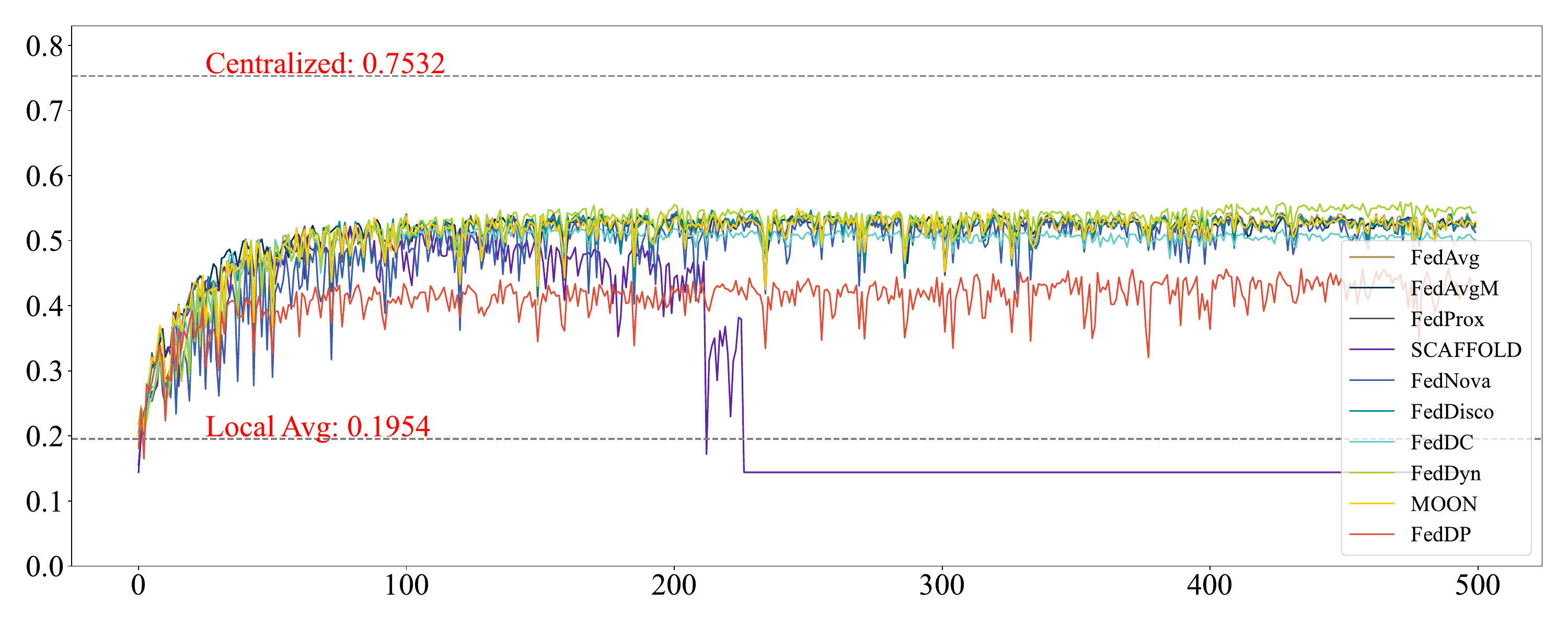} 
        \caption{Accuracy curve of CNN on FedRS NIID-1 partition. Testing is performed on $T_I$.}
        \label{fig:curve-a} 
    \end{subfigure}
        \hfill 
    \begin{subfigure}[b]{0.48\textwidth} 
        \centering
        \includegraphics[width=\linewidth]{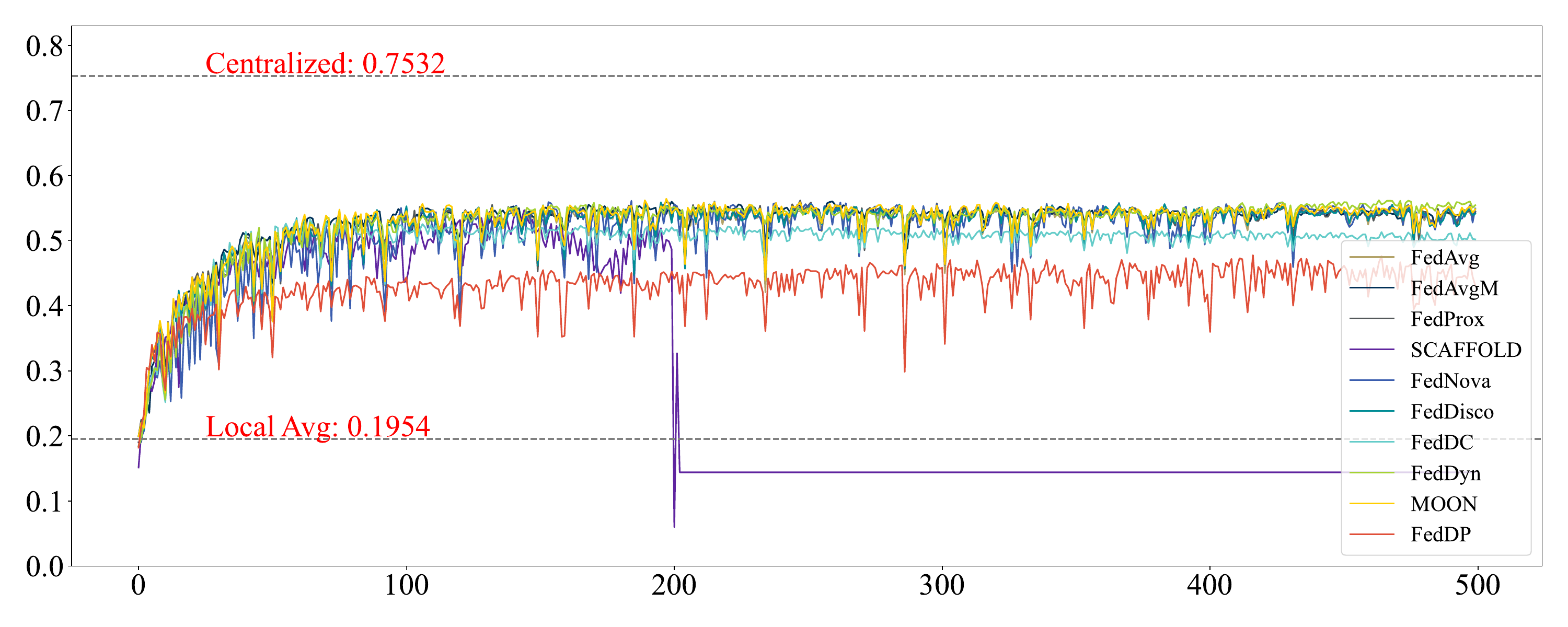} 
        \caption{Accuracy curve of CNN on FedRS NIID-2 partition. Testing is performed on $T_I$.}
        \label{fig:curve-b}
    \end{subfigure}
    \begin{subfigure}[b]{0.48\textwidth} 
        \centering
        \includegraphics[width=\linewidth]{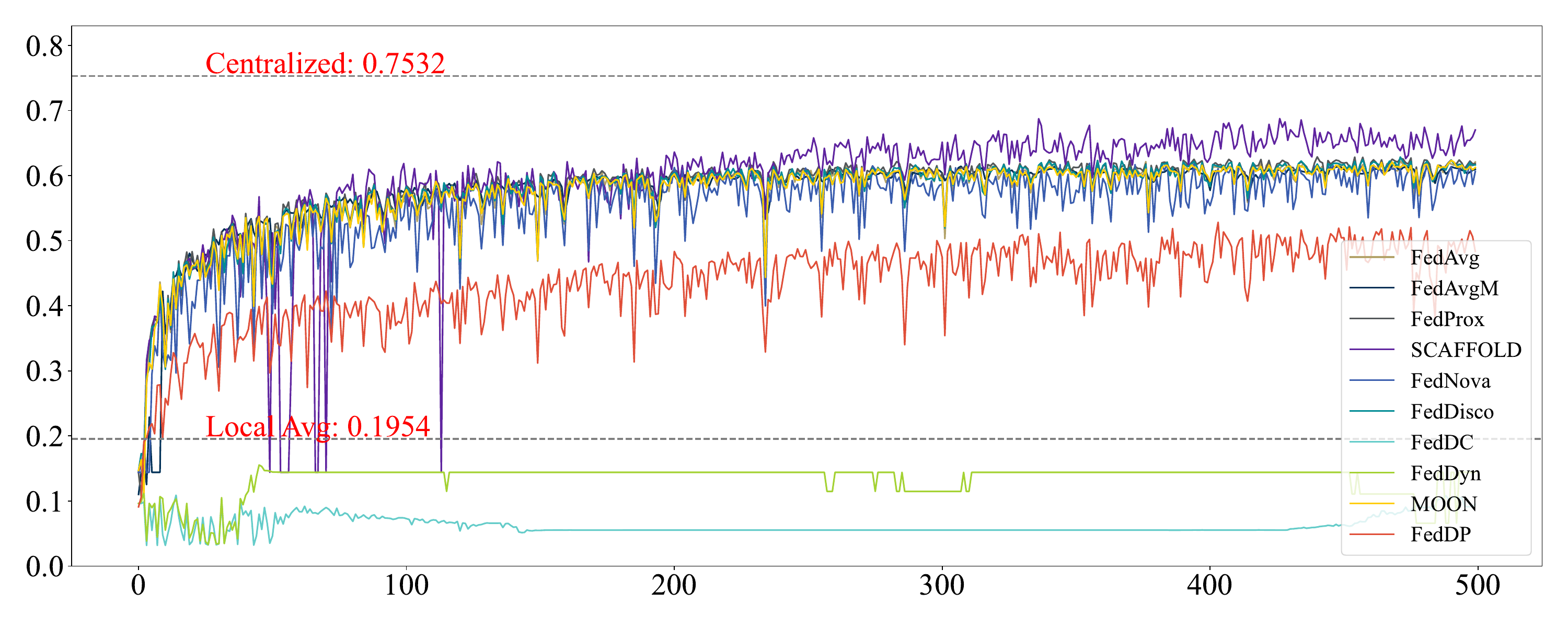} 
        \caption{Accuracy curve of ResNet18 on FedRS NIID-1 partition. Testing is performed on $T_I$.}
        \label{fig:curve-c} 
    \end{subfigure}
        \hfill 
    \begin{subfigure}[b]{0.48\textwidth} 
        \centering
        \includegraphics[width=\linewidth]{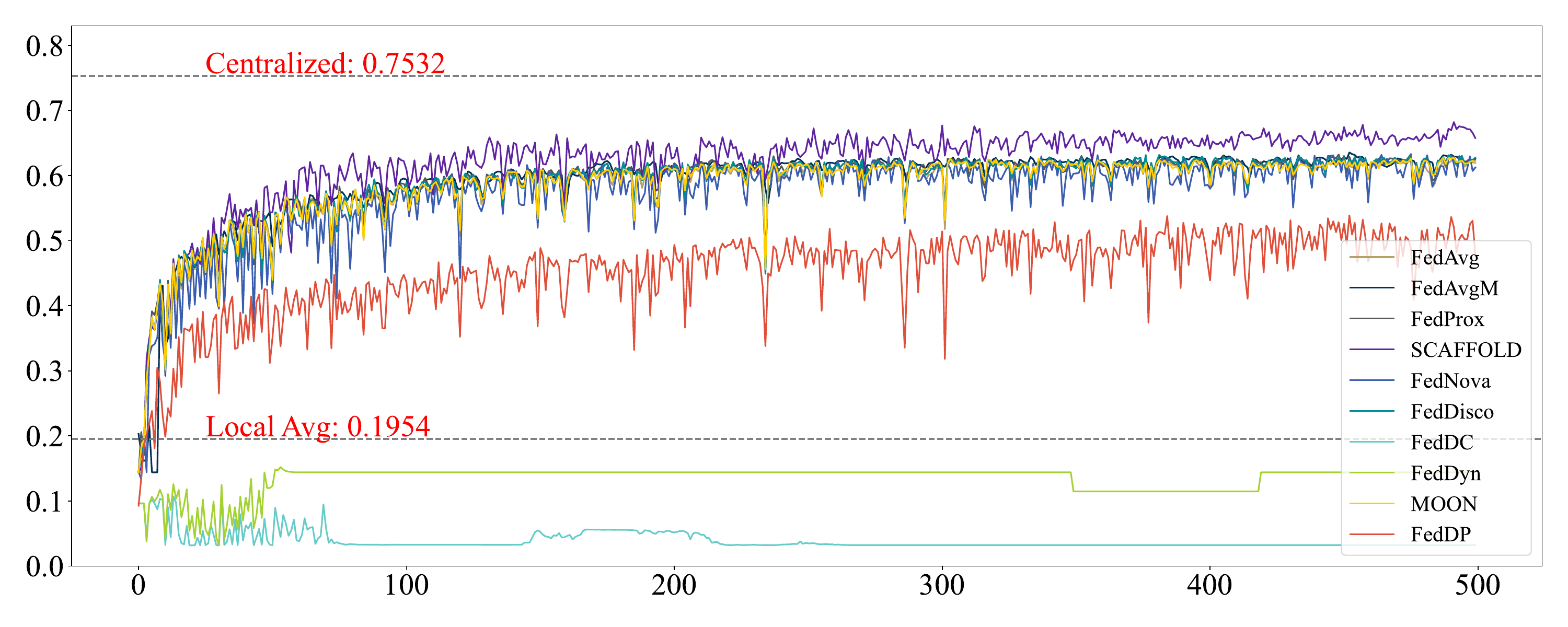} 
        \caption{Accuracy curve of ResNet18 on FedRS NIID-2 partition. Testing is performed on $T_I$.}
        \label{fig:curve-d}
    \end{subfigure}
    \caption{Test accuracy curve of FL baseline method during the communication round of training. More results on $T_B$ and FedRS-5 are shown in Appendix~\ref{sec:curves}.}
    \label{fig:curve} 
\end{figure}

We now present the comprehensive evaluation results of various FL algorithms benchmarked on FedRS and FedRS-5 dataset. Table~\ref{tab:main} reports the final best accuracy of each method on each test split. The average performance of ``Local Training'' (Local Avg, detailed in Figure~\ref{fig:local-rs15},~\ref{fig:local-rs5}) and performance of ``Centralized Training'' (Centralized) are also listed to compare with. Figure~\ref{fig:curve} illustrates the convergence curves (accuracy vs. communication rounds) for a representative subset of algorithms.

$\bullet$ \textbf{Federated Learning vs. Local Training vs. Centralized Training}: FL methods consistently and significantly outperformed the ``Local Avg'' baseline across all settings. For example, using ResNet18 on FedRS NIID-1 ($T_I$), FedAvg achieved 62.26\% accuracy compared to only 20.57\% for Local Avg. This underscores the benefit of collaborative training enabled by FL, even with the challenges of data heterogeneity inherent in FedRS. Generally, combining all client data for centralized training is considered the performance ceiling of federated learning. Combining Figures~\ref{fig:curve},~\ref{fig:curve-balance},~\ref{fig:curve-RS5}, the overall performance of FL is slightly lower than that of centralized training. However, in some of the settings in Figure~\ref{fig:curve-RS5}, the results of FL training reach or even exceed those of centralized training. We believe this may be because in relatively simple cases, the process of FL changes the direction of gradient descent, helping the global convergence to get rid of some local optimal points and reach a better global optimal point.

$\bullet$ \textbf{Performance Variation Across FL Methods}: No single FL algorithm consistently achieved the top performance across all experimental settings. The relative ranking of the methods varied depending on the dataset, model, and specific heterogeneity condition. For example, SCAFFOLD consistently achieves the best performance in various settings when using ResNet18, but does not perform well when using CNN. Methods like MOON also show strong performance, particularly when CNN is used. In contrast, some methods such as FedDC and FedDyn demonstrated notably lower performance with the ResNet18 model.

$\bullet$ \textbf{Impact of Heterogeneity (NIID-1 vs. NIID-2)}: Performance differences are observed between the NIID-1 (Dirichlet partitioning) and NIID-2 (uniform partitioning) settings, although overall trends often remained similar. This is consistent with many previous works and reflects the general challenges that heterogeneity brings to FL algorithms. This also highlights the benchmark's ability to test algorithm robustness under different types of data distribution skew.

$\bullet$ \textbf{Model Architecture Influence}: The choice of model architecture (CNN vs. ResNet18) influenced the absolute accuracy values and sometimes the relative performance ranking of the FL algorithms. Generally, choosing a more powerful base model can simultaneously enhance the model performance obtained by FL training.

$\bullet$ \textbf{Convergence Behavior}: Figures~\ref{fig:curve},~\ref{fig:curve-balance},~\ref{fig:curve-RS5} illustrate the convergence dynamics of the FL algorithms during communication rounds. Most methods demonstrate learning progress, converging towards their final accuracy levels, although convergence speed and stability varied among algorithms.
It is worth noting that many optimization algorithms proposed based on FedAvg, including SCAFFOLD, FedDC, and FedDyn, collapse during training in many scenarios and fail to converge to good performance. These phenomena do not occur when these algorithms are first proposed, which reminds us that when data heterogeneity reaches a certain level, the robustness and stability of the FL algorithm will decrease, and users need to be more cautious in choosing algorithms.

$\bullet$ \textbf{Privacy Protection}: 
Privacy is an important factor to consider for RS data. Although FL protects the privacy of training data by keeping the original data within the domain, some work has pointed out that some information about training data can still be obtained through attack methods. The FedDP method that uses user-level DP provides provable privacy on this basis, but its performance is significantly lower than that of other algorithms. This also reminds us that we need to strike a balance between performance and privacy when using FL.

\subsection{Further Exploration}
\begin{table}[]
\centering
\caption{The best accuracy results when using the pre-trained models to train on FedRS and test on $T_I$. All results are obtained with partial parameter updates, that only the last classification layer parameters are updated locally and globally aggregated. All results are shown in percentages (\%). \textbf{Bold} and \underline{underline} values highlight the best and and second-best accuracy, respectively.}
\label{tab:partial}
\resizebox{\textwidth}{!}{%
\begin{tabular}{@{}lllllllllll@{}}
\toprule
Model & FedAvg & FedAvgM & FedProx & SCAFFOLD & FedDC & FedDyn & FedNova & MOON & FedDisco  \\ \midrule
RseNet18 &  63.71     & 64.03        &  63.74       & \underline{65.56}         & 58.37      &  \textbf{67.59}      & 64.33        &  63.63    &  63.94        \\ 
RseNet50 &  72.04     & 72.10        & \underline{72.11}        & 67.98         & 62.39      & 61.32       & 71.75        & 72.04     & \textbf{72.34}          \\  
\bottomrule
\end{tabular}%
}
\end{table}
\textbf{Partial parameters updates.} With the widespread use of powerful base models with more parameters, the FL training process also tends to use more powerful base models. However, larger parameters usually bring great challenges to the local training and global update process of FL. At the same time, directly training all parameters of the pre-trained model may also lead to the loss of the original capabilities (catastrophic forgetting). Based on this, the method of updating and communicating only partial parameters has become increasingly popular~\cite{zhao2023fedprompt,bian2025survey}. In order to study the performance of more powerful models and newer training modes on FedRS, we select several commonly used pre-trained ResNet models with different parameter sizes, including ResNet18 and ResNet50.

Table~\ref{tab:partial} and Figure~\ref{fig:curve-pretrained} show the best accuracy results and accuracy curves of using partial parameters updated on pre-trained models. For most FL algorithms, convergence is achieved in fewer rounds with partial parameter updates than with full parameter updates in Table~\ref{tab:main}, and the best accuracy is comparable to that of full parameter updates. This fully demonstrates the advantage of using public pre-trained models in FL training. At the same time, it can be seen that for SCAFFOLD, FedDC, and FedDyn, their instability still exists, which needs further attention.

\section{Conclusion}
This paper presents FedRS, the first realistic FL dataset for RS. FedRS is crafted from eight real-world RS datasets, reflecting the data heterogeneity of practical multi-source scenarios. Our detailed FedRS-Bench results with ten FL algorithms confirm the advantages of federation over isolated training and reveal that most existing methods struggle to achieve centralized-level performance under realistic NIID conditions. By making FedRS publicly available, we hope to encourage the community to test new FL methods against truly challenging heterogeneous data. This resonates with concepts from domain adaptation and multi-domain learning – bridging FL with those fields could be fruitful. Our work takes a significant step toward evaluating FL in a realistic RS context. By providing a novel dataset and extensive benchmark results, we aim to establish a reference for future research.


\newpage
{
\small
\bibliographystyle{unsrt}
\bibliography{refs}
}
\appendix
\section{Limitations}
With the continuous introduction of new remote sensing data sets and the continuous optimization of federated learning algorithms, FedRS and FedRS-Bench will need to be continuously updated in the future to adapt to the latest needs. Future benchmarks could include more multi-label scenarios and long-tailed distributions to push development in this area.
\section{Dataset Statistics}
\label{sec:dataset}
This section provides more detailed information about the construction and characteristics of the FedRS dataset.
\subsection{Data sources}
\label{sec:sources}
\begin{figure}[!h] 
    \centering 
    \begin{subfigure}[b]{0.23\textwidth} 
        \centering
        \includegraphics[width=\linewidth]{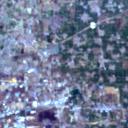} 
        \caption{NaSC-TG2.} 
    \end{subfigure}
        \hfill 
    \begin{subfigure}[b]{0.23\textwidth} 
        \centering
        \includegraphics[width=\linewidth]{figures/samples/WHU-RS19.png} 
        \caption{WHU-RS19.}
    \end{subfigure}
            \hfill 
    \begin{subfigure}[b]{0.23\textwidth} 
        \centering
        \includegraphics[width=\linewidth]{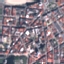} 
        \caption{EuroSAT.}
    \end{subfigure}
    \hfill 
    \begin{subfigure}[b]{0.23\textwidth} 
        \centering
        \includegraphics[width=\linewidth]{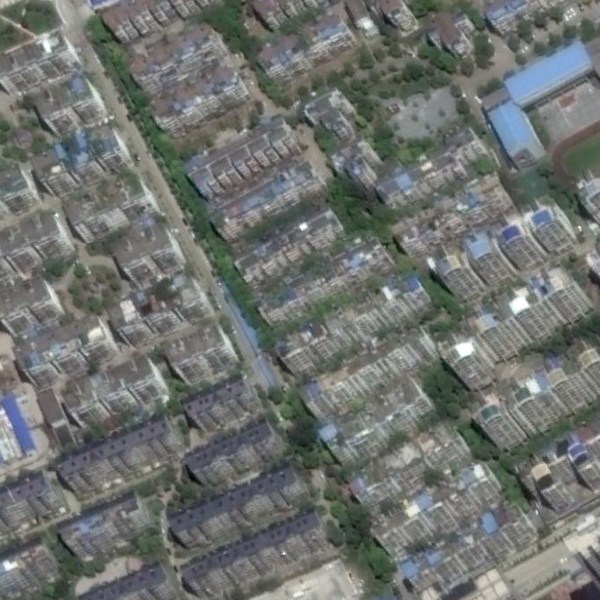} 
        \caption{AID.}
    \end{subfigure}
    \vspace{\baselineskip} 
    \begin{subfigure}[b]{0.23\textwidth} 
        \centering
        \includegraphics[width=\linewidth]{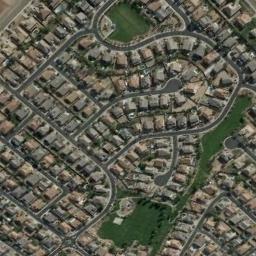} 
        \caption{NWPU-RESISC45.} 
    \end{subfigure}
        \hfill 
    \begin{subfigure}[b]{0.23\textwidth} 
        \centering
        \includegraphics[width=\linewidth]{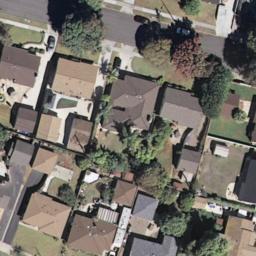} 
        \caption{UCM.}
    \end{subfigure}
            \hfill 
    \begin{subfigure}[b]{0.23\textwidth} 
        \centering
        \includegraphics[width=\linewidth]{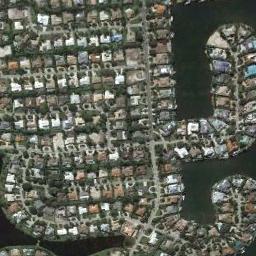} 
        \caption{Optimal-31.}
    \end{subfigure}
    \hfill 
    \begin{subfigure}[b]{0.23\textwidth} 
        \centering
        \includegraphics[width=\linewidth]{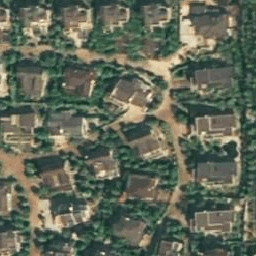} 
        \caption{RSD46-WHU.}
    \end{subfigure}
    \caption{Taking the Residential class as an example, there are obvious differences in the style of data in different data sources.}
    \label{fig:case} 
\end{figure}

FedRS is constructed from eight different existing remote sensing datasets, ensuring that there is no data overlap between them.
A key characteristic of real-world federated scenarios is the heterogeneity of data from different sources. Even within the same semantic category (e.g., "Residential"), images from different original datasets (NaSC-TG2, WHU-RS19, EuroSAT, AID, NWPU-RESISC45, UCM, Optimal-31, RSD46-WHU) exhibit noticeable variations in style, resolution, and viewpoint. This is illustrated for the "Residential" class in Figure~\ref{fig:case}.

The original datasets employ varying category labels with differing granularities, leading to overlaps and ambiguities. To address this, a manual review is conducted to select 15 conflict-free semantic categories with an appropriate number of samples. The labels of the original datasets are then mapped to these unified FedRS categories. The detailed mapping is presented in Table~\ref{tab:label-map}, showing how multiple labels from various source datasets might contribute to a single category in FedRS (e.g., "Agriculture" encompassing "Rectangular Farmland," "Farmland," "Annual Crop," etc.).

\begin{figure}[htbp] 
    \centering 

    \begin{subfigure}[b]{0.31\textwidth} 
        \centering
        \includegraphics[width=\linewidth]{figures/tsne_Agriculture.pdf} 
        \caption{Agriculture (0.07)}
        \label{fig:s1} 
    \end{subfigure}
    \hfill 
    \begin{subfigure}[b]{0.31\textwidth}
        \centering
        \includegraphics[width=\linewidth]{figures/tsne_Airport.pdf} 
        \caption{Airport (-0.21)}
        \label{fig:s2} 
    \end{subfigure}
        \hfill 
    \begin{subfigure}[b]{0.31\textwidth} 
        \centering
        \includegraphics[width=\linewidth]{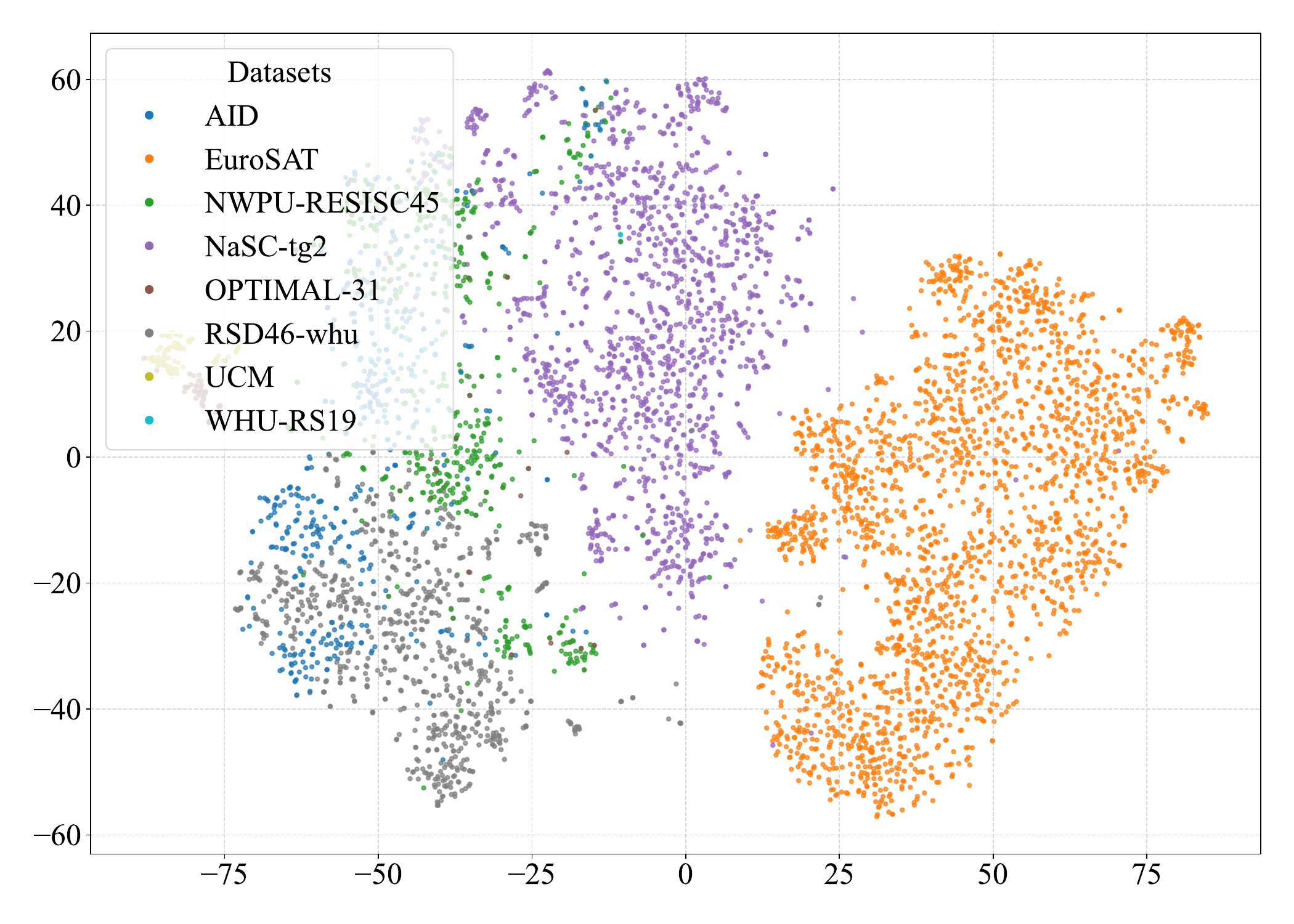} 
        \caption{Bareland (0.25)}
        \label{fig:s3} 
    \end{subfigure}

    \vspace{\baselineskip} 

    \begin{subfigure}[b]{0.31\textwidth} 
        \centering
        \includegraphics[width=\linewidth]{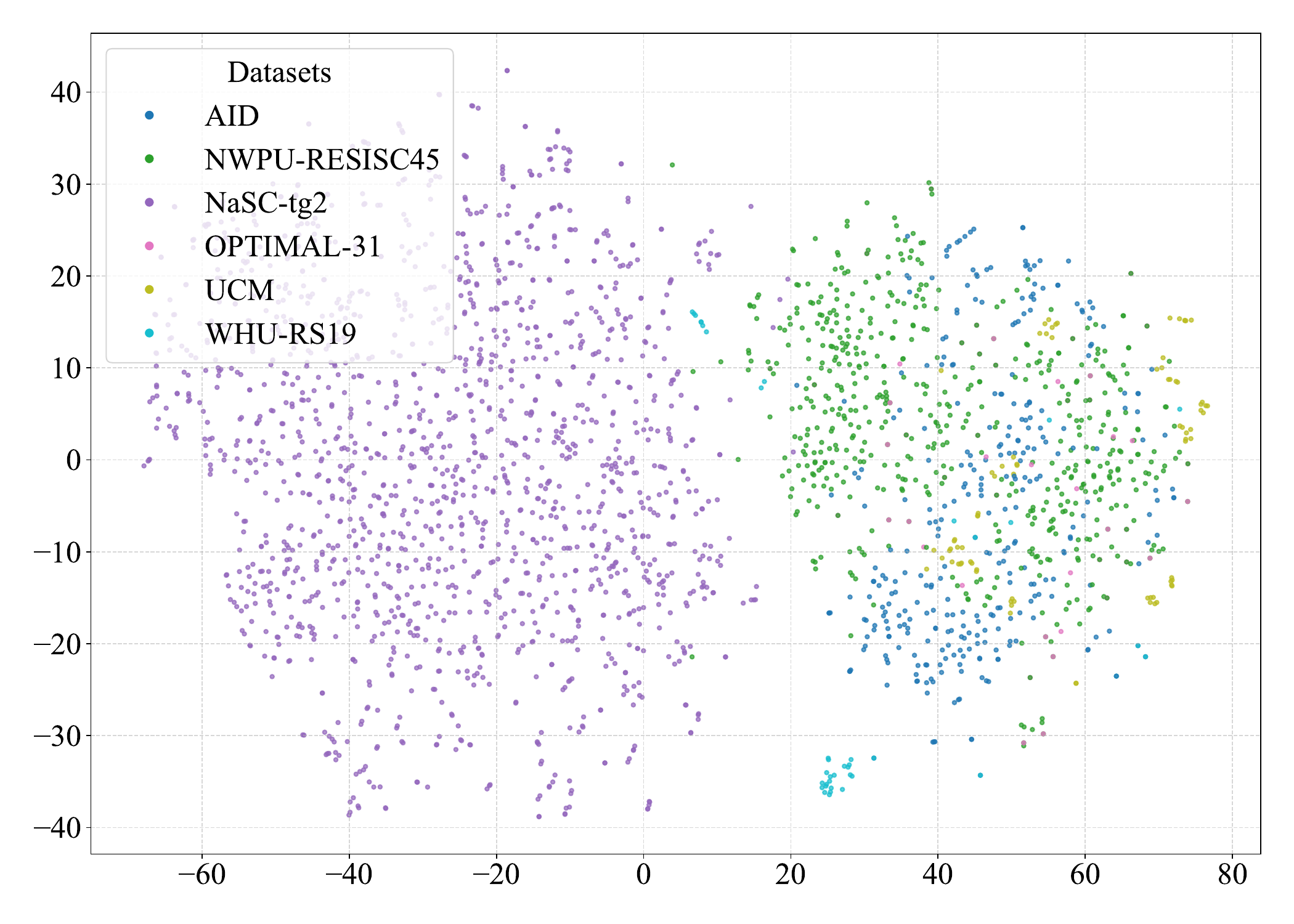} 
        \caption{Beach (0.26)}
        \label{fig:s4} 
    \end{subfigure}
    \hfill 
    \begin{subfigure}[b]{0.31\textwidth} 
        \centering
        \includegraphics[width=\linewidth]{figures/tsne_Bridge.pdf} 
        \caption{Bridge (0.06)}
        \label{fig:s5} 
    \end{subfigure}
        \hfill 
    \begin{subfigure}[b]{0.31\textwidth} 
        \centering
        \includegraphics[width=\linewidth]{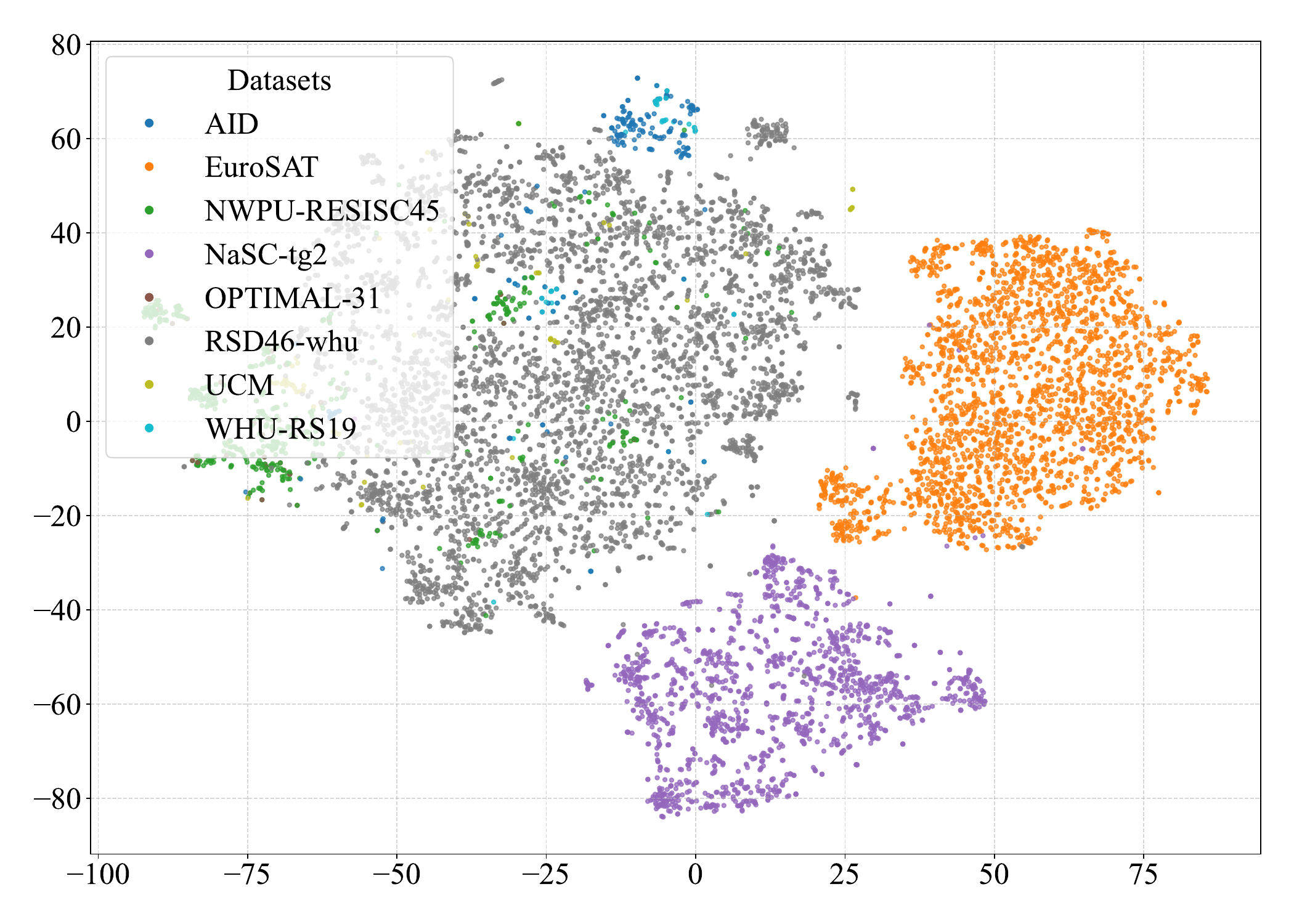} 
        \caption{Forest (0.17)}
        \label{fig:s6}
    \end{subfigure}

    \vspace{\baselineskip} 

    \begin{subfigure}[b]{0.31\textwidth} 
        \centering
        \includegraphics[width=\linewidth]{figures/tsne_Highway.pdf} 
        \caption{Highway (0.37)}
        \label{fig:s7} 
    \end{subfigure}
    \hfill 
    \begin{subfigure}[b]{0.31\textwidth} 
        \centering
        \includegraphics[width=\linewidth]{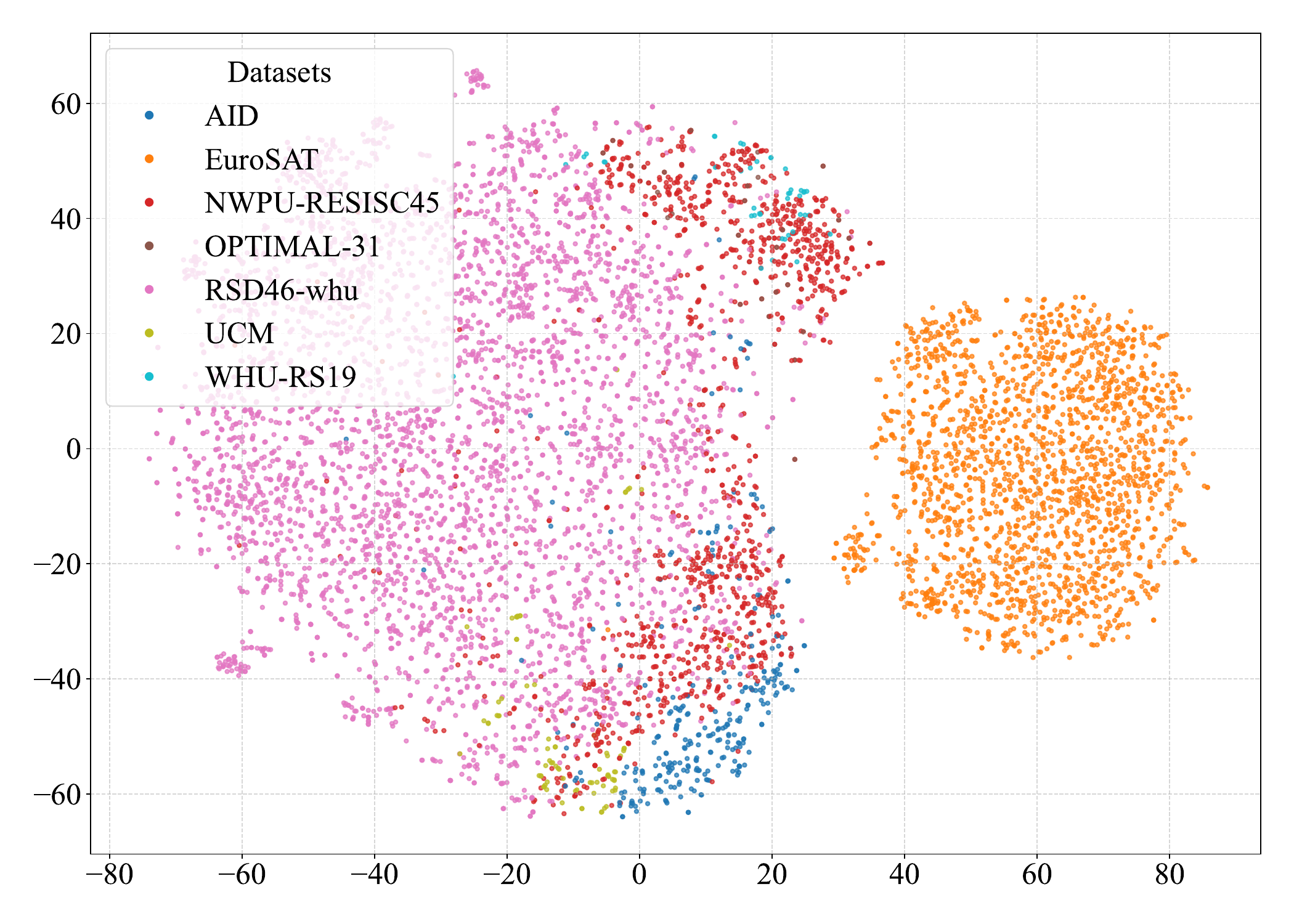} 
        \caption{Industrial (0.05)}
        \label{fig:s8} 
    \end{subfigure}
        \hfill 
    \begin{subfigure}[b]{0.31\textwidth} 
        \centering
        \includegraphics[width=\linewidth]{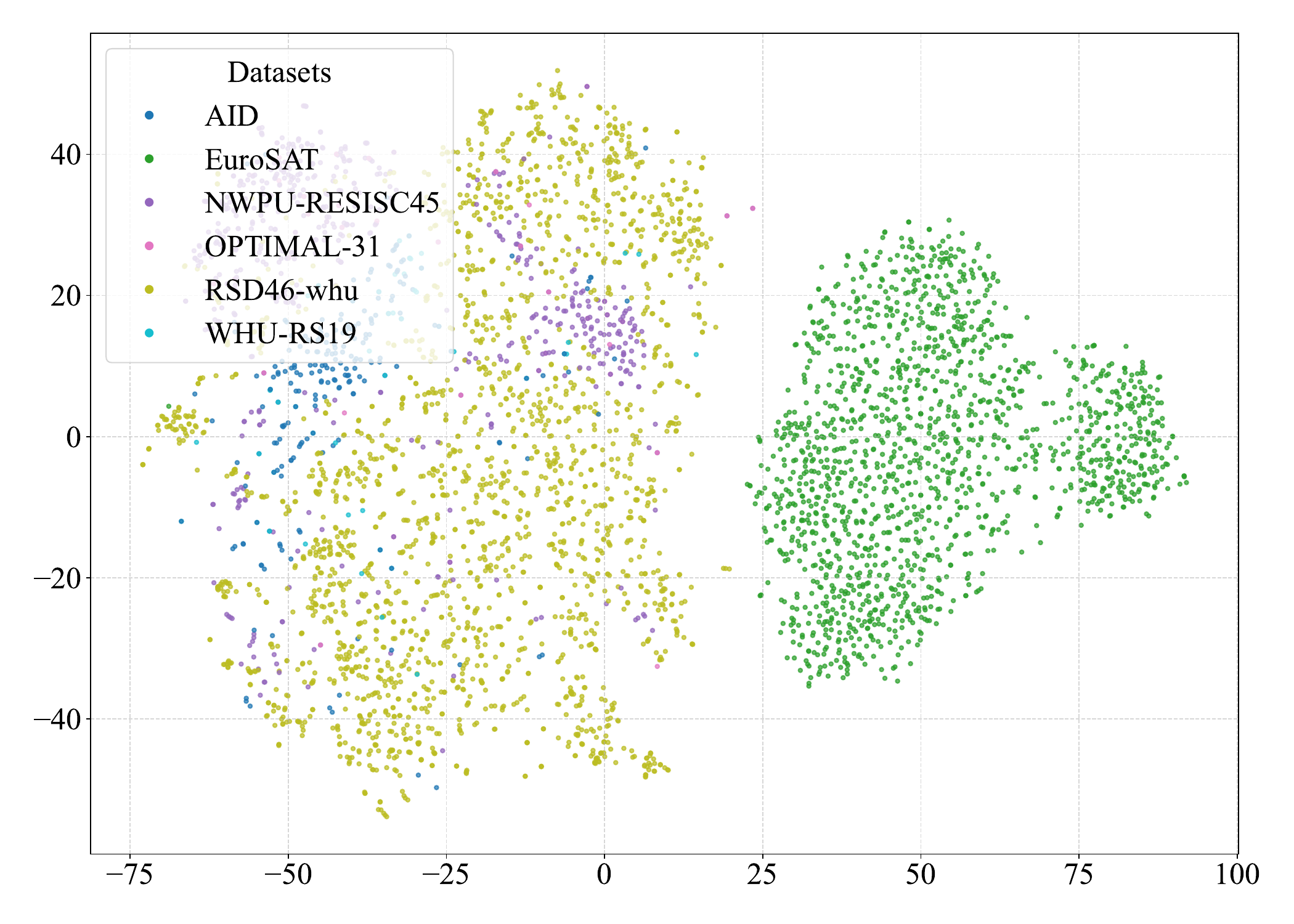} 
        \caption{Meadow (0.16)}
        \label{fig:s9}
    \end{subfigure}
        \vspace{\baselineskip} 

    \begin{subfigure}[b]{0.31\textwidth} 
        \centering
        \includegraphics[width=\linewidth]{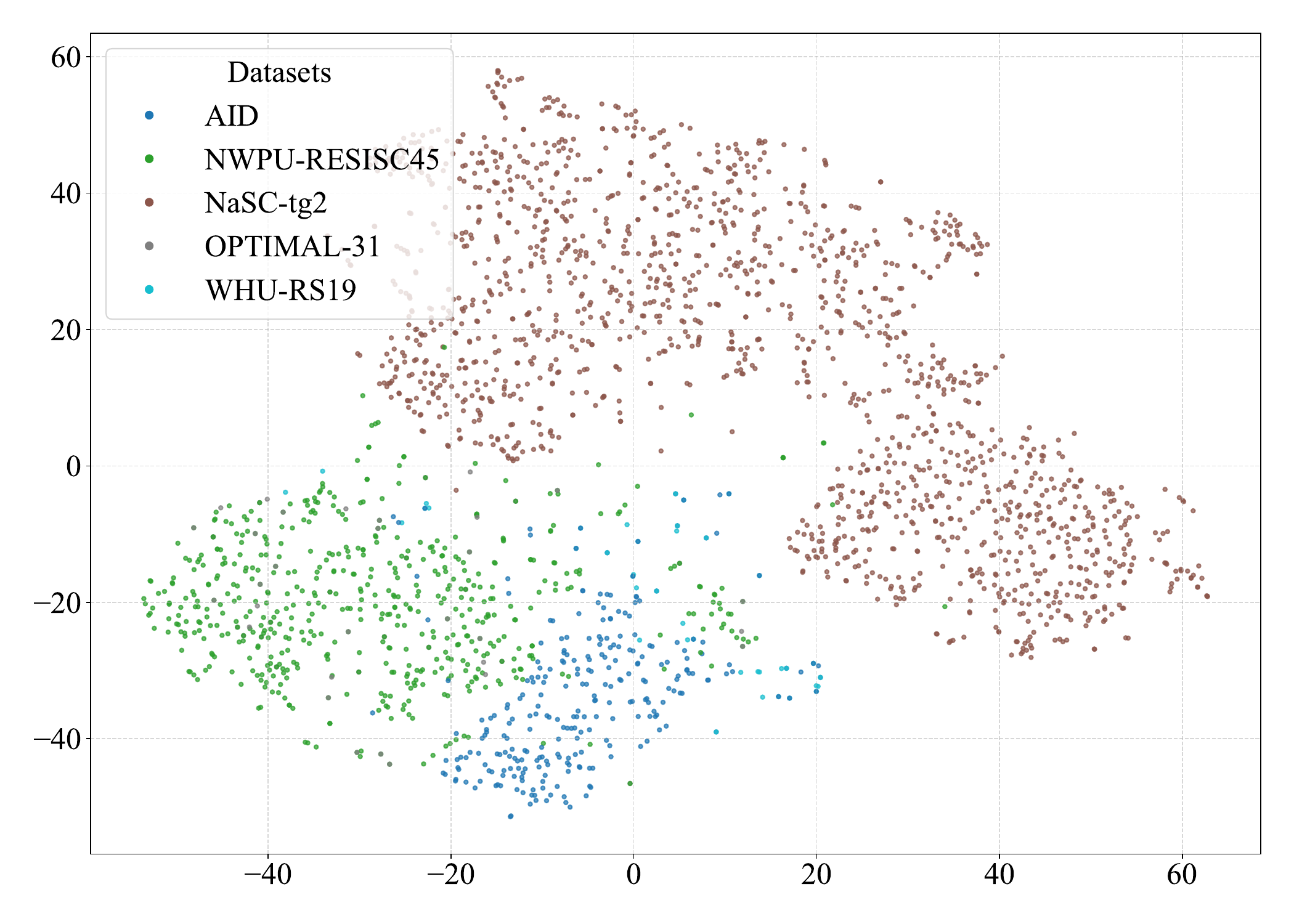} 
        \caption{Mountain (0.10)}
        \label{fig:s10} 
    \end{subfigure}
    \hfill 
    \begin{subfigure}[b]{0.31\textwidth} 
        \centering
        \includegraphics[width=\linewidth]{figures/tsne_Overpass.pdf} 
        \caption{Overpass (0.03)}
        \label{fig:s11} 
    \end{subfigure}
        \hfill 
    \begin{subfigure}[b]{0.31\textwidth} 
        \centering
        \includegraphics[width=\linewidth]{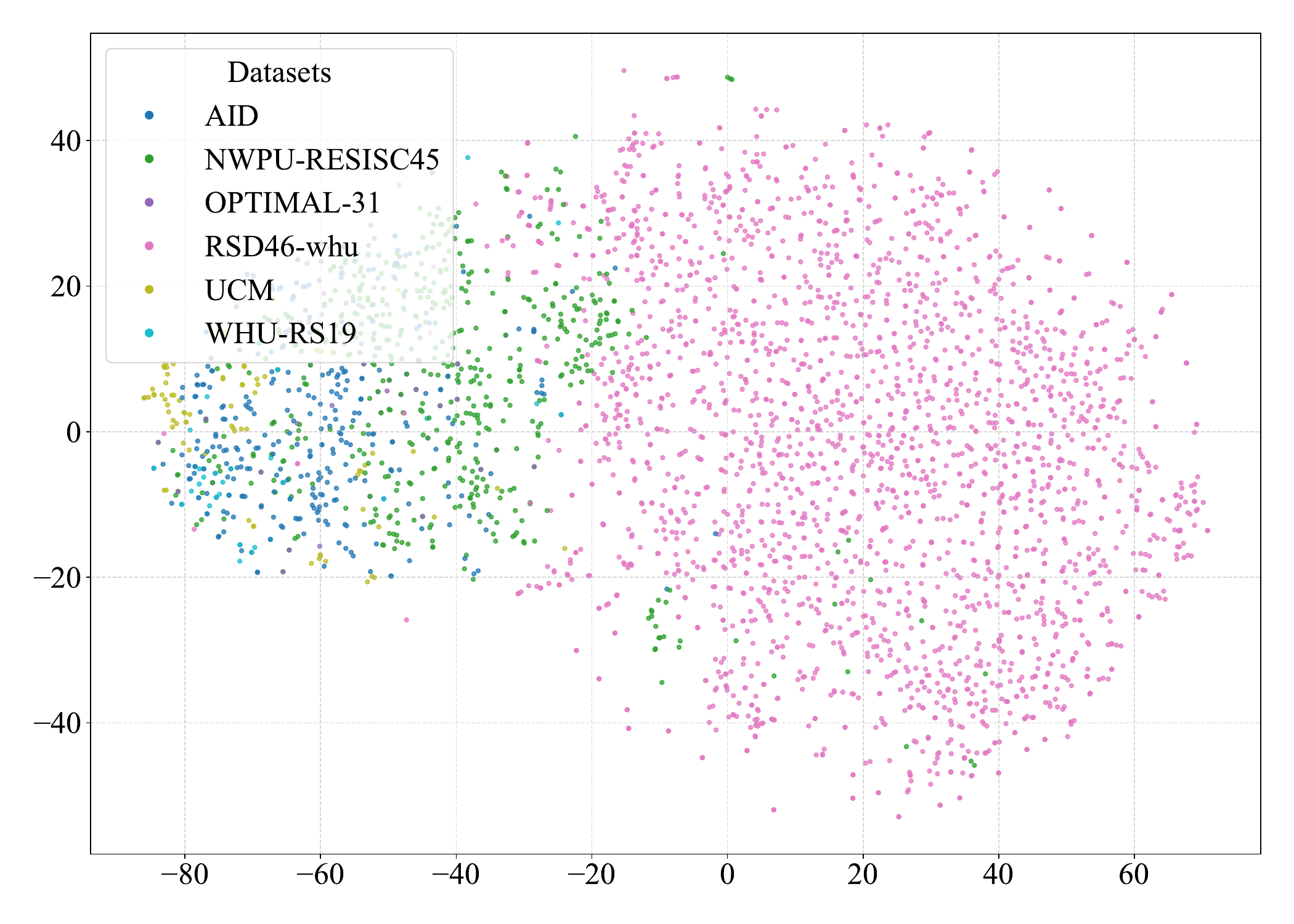} 
        \caption{Parkinglot (0.21)}
        \label{fig:s12}
    \end{subfigure}
        \vspace{\baselineskip} 

    \begin{subfigure}[b]{0.31\textwidth} 
        \centering
        \includegraphics[width=\linewidth]{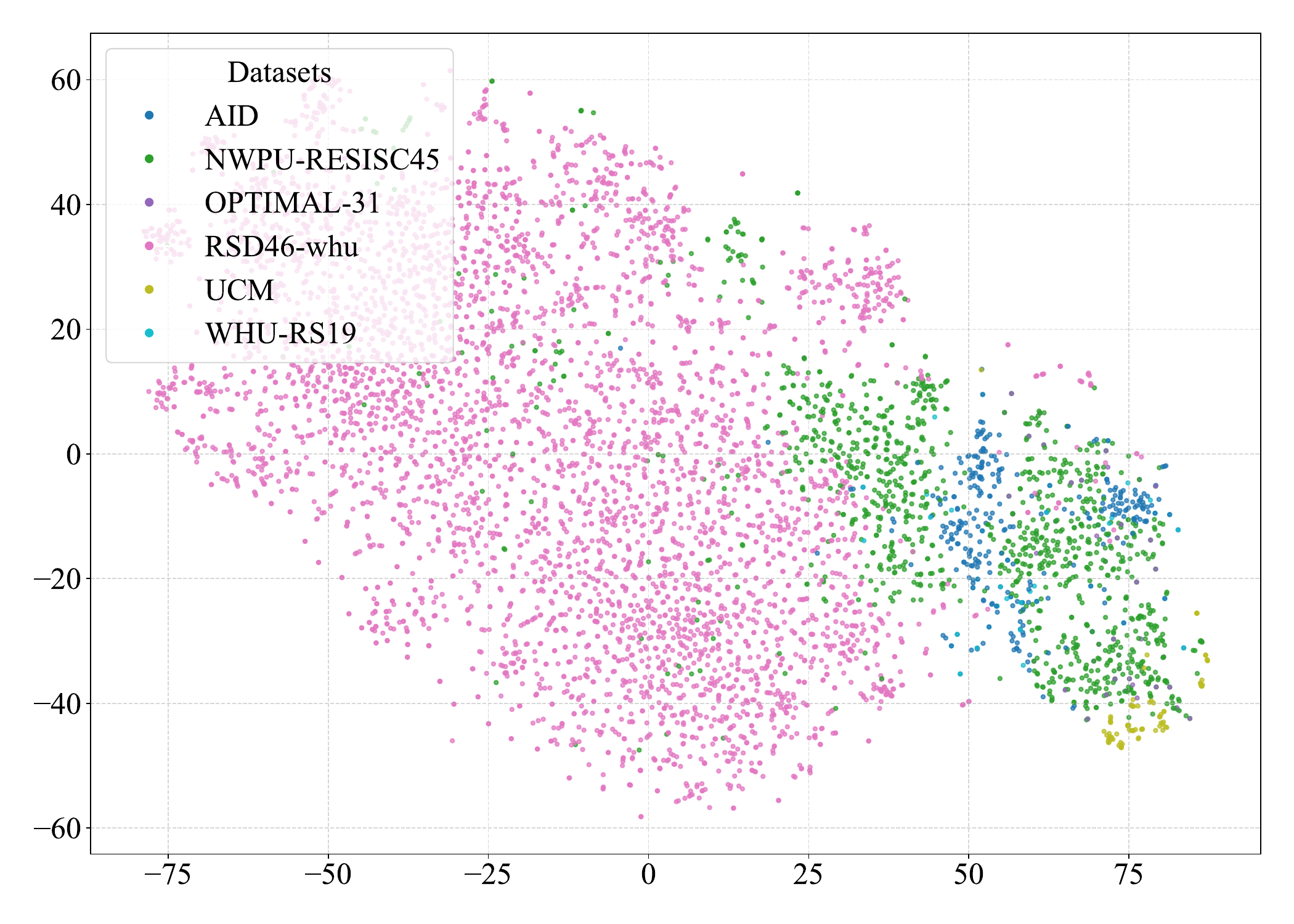} 
        \caption{Port (0.14)}
        \label{fig:s13} 
    \end{subfigure}
    \hfill 
    \begin{subfigure}[b]{0.31\textwidth} 
        \centering
        \includegraphics[width=\linewidth]{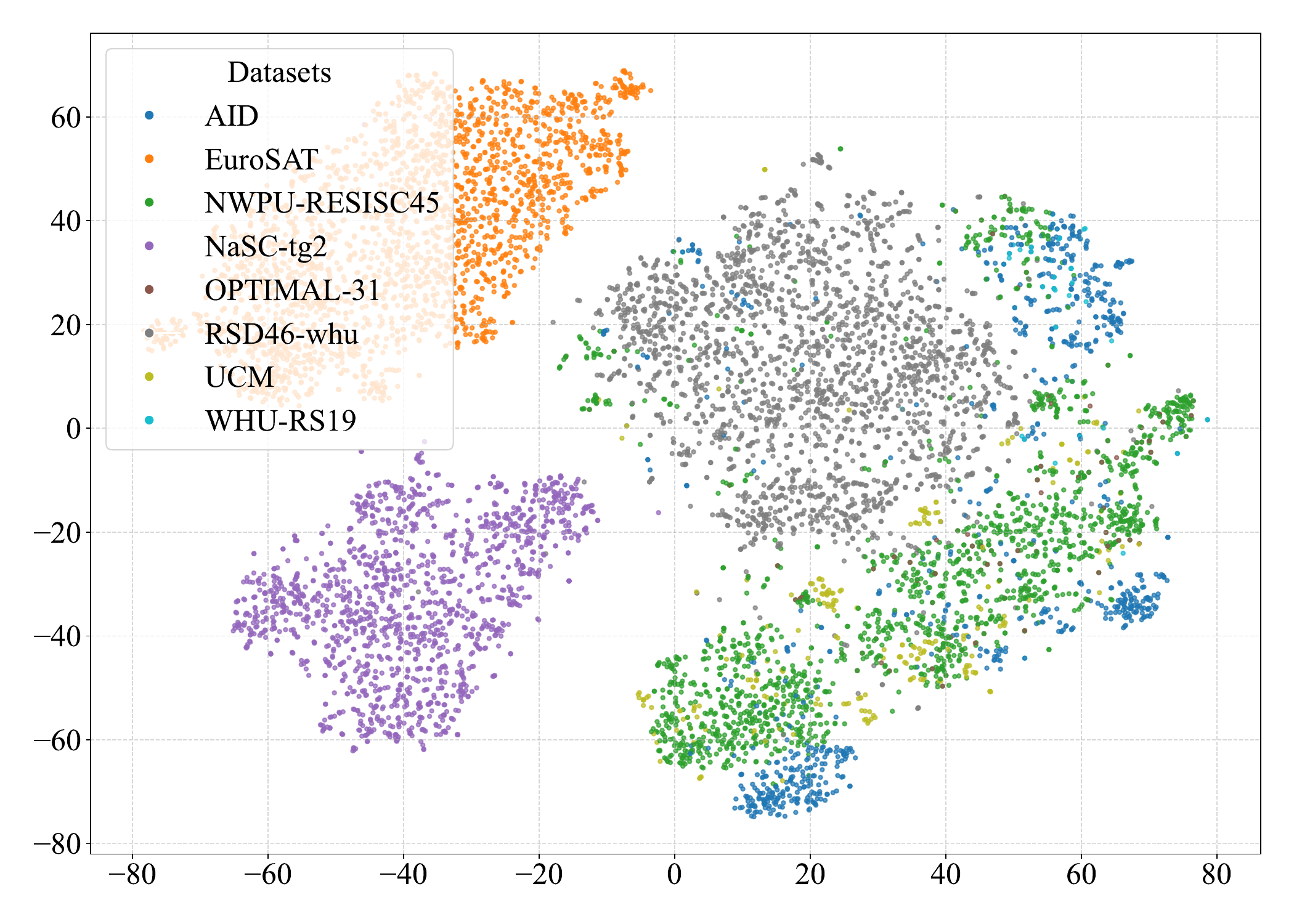} 
        \caption{Residential (0.21)}
        \label{fig:s14} 
    \end{subfigure}
        \hfill 
    \begin{subfigure}[b]{0.31\textwidth} 
        \centering
        \includegraphics[width=\linewidth]{figures/tsne_River.pdf} 
        \caption{River (0.33)}
        \label{fig:s15}
    \end{subfigure}
    \caption{Visualization of features of from different sources using t-SNE. The corresponding categories and silhouette scores after dimensionality reduction are shown in the subtitles, reflecting the intra-category feature deviation  from different sources.}
    \label{fig:tsne_sources_all} 
\end{figure}

Following label unification, the feature distributions of images from different sources still show significant differences in both quantity and characteristics. Figure~\ref{fig:tsne_sources_all} provides t-SNE visualizations for all 15 unified categories, illustrating the intra-category feature deviations and silhouette scores for data originating from different sources. This heterogeneity reflects realistic scenarios in which clients naturally possess data with distinct underlying distributions.

\begin{figure}[!tp] 
    \centering 
        \includegraphics[width=0.7\linewidth]{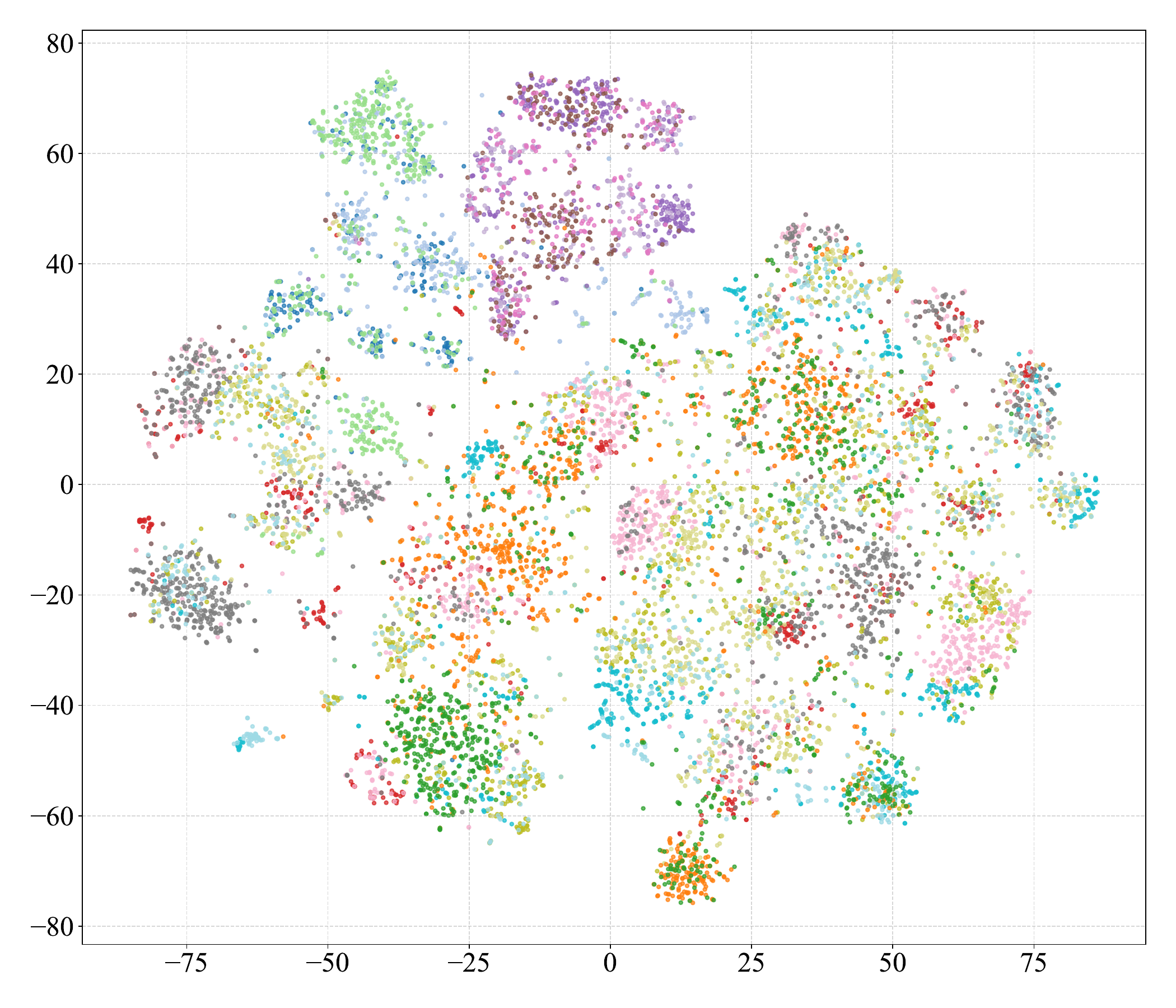} 
    \caption{Visualization of features of different clients in FedRS. Several clients are randomly selected to plot in the figure, and each color represents a client.}
    \label{fig:tsne-clients} 
\end{figure}

The client-level data heterogeneity is further visualized in Figure~\ref{fig:tsne-clients} using t-SNE. This figure shows feature distributions for randomly selected clients, where each color represents a distinct client. Two types of data shifts are evident:
\begin{itemize}
    \item Intra-client shift: Many clients exhibit multiple clusters, corresponding to the different data categories they hold. 
    \item Inter-client shift: Clear separations and shifts exist between the feature spaces of different clients.  These characteristics underscore FedRS's realistic simulation of federated environments where clients often have differing data sources and characteristics.
\end{itemize}

\begin{sidewaystable}
\centering
\caption{The data source and label mapping of each category in the FedRS dataset. Due to different annotations in different datasets, a category in FedRS may contain multiple labels from multiple datasets.}
\label{tab:label-map}
\resizebox{\textwidth}{!}{%
\begin{tabular}{@{}lllllllll@{}}
\toprule
\textbf{Category}    & \textbf{NaSC-TG2}             & \textbf{WHU-RS19}    & \textbf{EuroSAT}                                                              & \textbf{AID}                                                                                                 & \textbf{NWPU-RESISC45}                                                                                       & \textbf{UCM}                                                                                                  & \textbf{OPTIMAL-31}                                                                     & \textbf{RSD46-WHU}                                                                                        \\ \midrule
Agriculture & Rectangular Farmland & Farmland    & \begin{tabular}[c]{@{}l@{}}Annual Crop\\ Permanent Crop\end{tabular} & Farmland                                                                                            & Rectangular Farmland                                                                                & Agricultural                                                                                         & Rectangular Farmland                                                           & \begin{tabular}[c]{@{}l@{}}Irregular Farmland\\ Regular Farmland\end{tabular}                    \\ \midrule
Bareland    & Desert               & Desert      & Herbaceous Vegetation                                                & \begin{tabular}[c]{@{}l@{}}Bareland\\ Desert\end{tabular}                                           & Desert                                                                                              & Chaparral                                                                                            & \begin{tabular}[c]{@{}l@{}}Desert\\ Chaparral\end{tabular}                     & Bare Land                                                                                        \\ \midrule
Residential & Residential          & Residential & Residential                                                          & \begin{tabular}[c]{@{}l@{}}Dense Residential\\ Medium Residential\\ Sparse Residential\end{tabular} & \begin{tabular}[c]{@{}l@{}}Dense Residential\\ Sparse Residential\\ Medium Residential\end{tabular} & \begin{tabular}[c]{@{}l@{}}Dense Residential\\ Sparse Residential \\ Medium Residential\end{tabular} & \begin{tabular}[c]{@{}l@{}}Dense Residential\\ Medium Residential\end{tabular} & Sparse Residential Area                                                                          \\ \midrule
River       & River                & River       & River                                                                & River                                                                                               & River                                                                                               & River                                                                                                & River                                                                          & River                                                                                            \\ \midrule
Forest      & Forest               & Forest      & Forest                                                               & Forest                                                                                              & Forest                                                                                              & Forest                                                                                               & Forest                                                                         & \begin{tabular}[c]{@{}l@{}}Natural Dense Forest Land\\ Artificial Dense Forest Land\end{tabular} \\ \midrule
Airport     & -                    & Airport     & -                                                                    & Airport                                                                                             & \begin{tabular}[c]{@{}l@{}}Airport\\ Airplane\end{tabular}                                          & Airport                                                                                              & \begin{tabular}[c]{@{}l@{}}Airport\\ Airplane\\ Runway\end{tabular}            & \begin{tabular}[c]{@{}l@{}}Airport\\ Airplane\end{tabular}                                       \\ \midrule
Beach       & Beach                & Beach       & -                                                                    & Beach                                                                                               & Beach                                                                                               & Beach                                                                                                & Beach                                                                          & Beach                                                                                            \\ \midrule
Highway     & -                    & -     & Highway                                                              & -                                                                                             & Freeway                                                                                             & Freeway                                                                                              & Freeway                                                                        & Freeway                                                                                          \\ \midrule
Industrial  & -                    & Industrial  & Industrial                                                           & Storage Tanks                                                                                       & \begin{tabular}[c]{@{}l@{}}Storage Tank\\ Industrial Area\end{tabular}                              & Storage Tanks                                                                                        & Industrial Area                                                                & \begin{tabular}[c]{@{}l@{}}Refinery\\ Steel Smelter\end{tabular}                                 \\ \midrule
Port        & -                    & Port        & -                                                                    & Harbor                                                                                              & \begin{tabular}[c]{@{}l@{}}Harbor\\ Ship\end{tabular}                                               & Harbor                                                                                               & Harbor                                                                         & \begin{tabular}[c]{@{}l@{}}Dock\\ Ship\end{tabular}                                              \\ \midrule
Overpass    & -                    & Viaduct     & -                                                                    & Viaduct                                                                                             & Overpass                                                                                            & Overpass                                                                                             & Overpass                                                                       & Overpass                                                                                         \\ \midrule
Parkinglot  & -                    & Parking     & -                                                                    & Parking                                                                                             & Parking Lot                                                                                         & Parkinglot                                                                                           & Parking Lot                                                                    & Parking Lot                                                                                      \\ \midrule
Bridge      & -                    & Bridge      & -                                                                    & Bridge                                                                                              & Bridge                                                                                              & -                                                                                                    & Bridge                                                                         & Cross River Bridge                                                                               \\ \midrule
Mountain    & Mountain             & Mountain    & -                                                                    & Mountain                                                                                            & Mountain                                                                                            & -                                                                                                    & Mountain                                                                       & -                                                                                                \\ \midrule
Meadow      & -                    & Meadow      & Pasture                                                              & Meadow                                                                                              & Meadow                                                                                              & -                                                                                                    & Meadow                                                                         & Grassland                                                                                        \\ \bottomrule
\end{tabular}%
}
\end{sidewaystable}

\section{Benchmark Results}
This section provides supplementary results to those presented in the main paper.
\subsection{Local Training Results}

\begin{figure}[!tp] 
    \centering 
    \begin{subfigure}[b]{0.48\textwidth} 
        \centering
        \includegraphics[width=\linewidth]{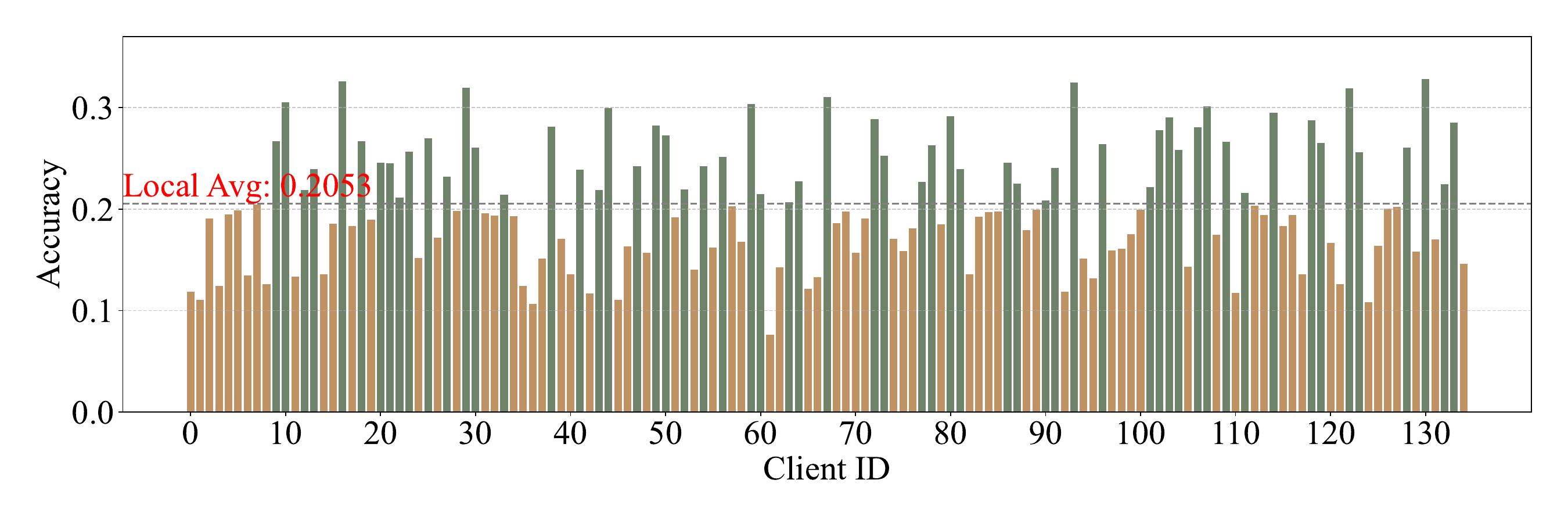} 
        \caption{Local accuracy of ResNet18 on FedRS NIID-1 partition. Testing is performed on $T_I$.}
    \end{subfigure}
        \hfill 
    \begin{subfigure}[b]{0.48\textwidth} 
        \centering
        \includegraphics[width=\linewidth]{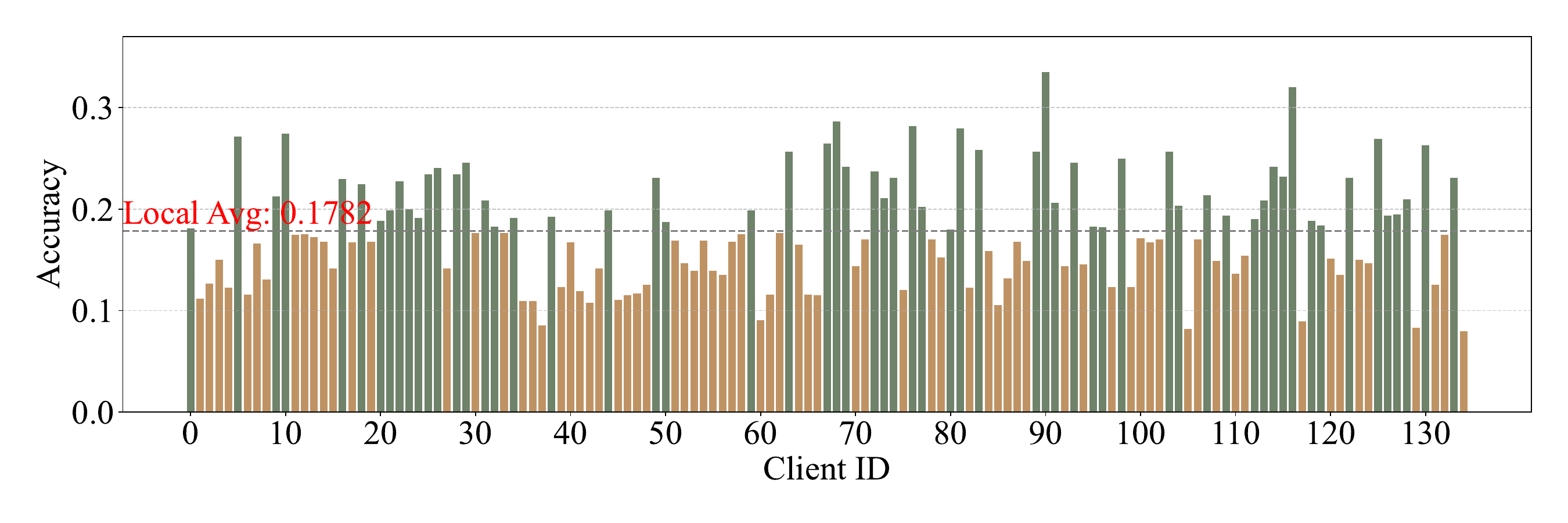} 
        \caption{Local accuracy of ResNet18 on FedRS NIID-1 partition. Testing is performed on $T_B$.}
    \end{subfigure}

    \vspace{\baselineskip} 
    \begin{subfigure}[b]{0.48\textwidth} 
        \centering
        \includegraphics[width=\linewidth]{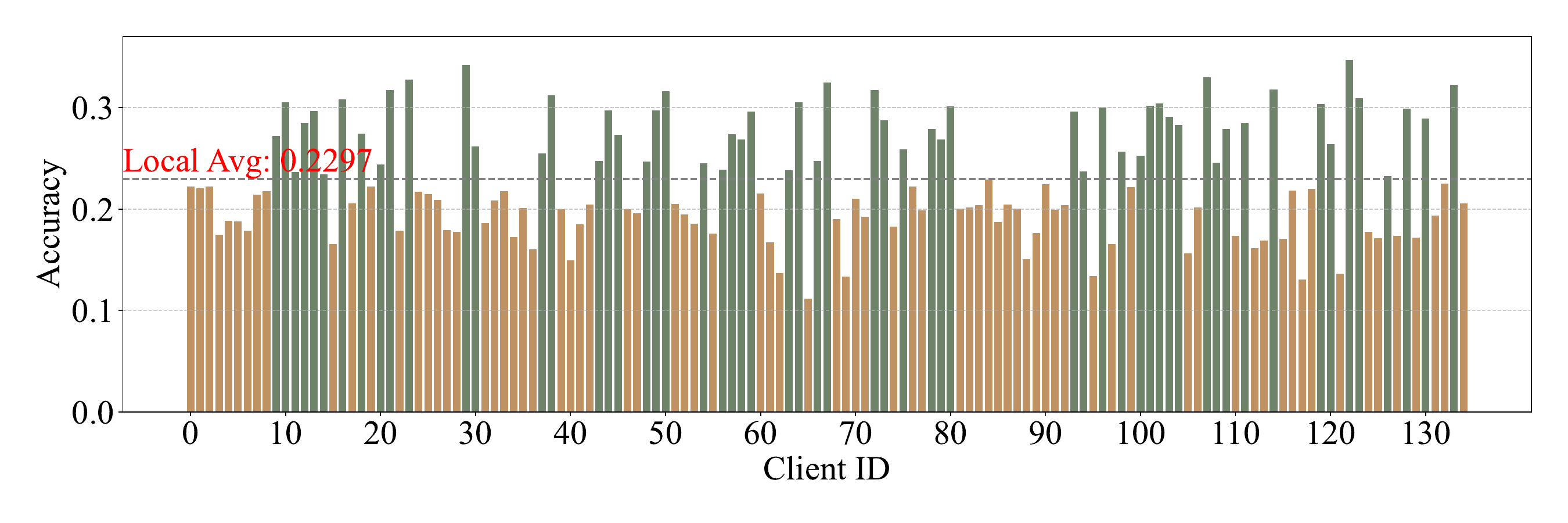} 
        \caption{Local accuracy of ResNet18 on FedRS NIID-2 partition. Testing is performed on $T_I$.}
    \end{subfigure}
        \hfill 
    \begin{subfigure}[b]{0.48\textwidth} 
        \centering
        \includegraphics[width=\linewidth]{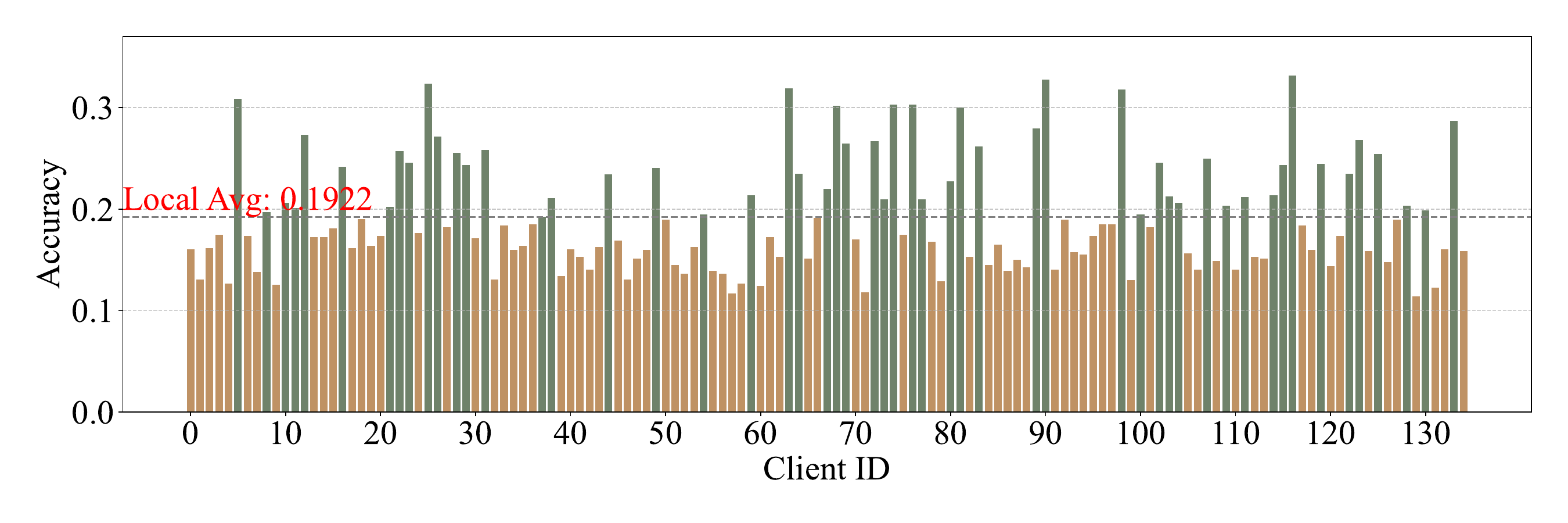} 
        \caption{Local accuracy of ResNet18 on FedRS NIID-2 partition. Testing is performed on $T_B$.}
    \end{subfigure}
    \vspace{\baselineskip} 
        \begin{subfigure}[b]{0.48\textwidth} 
        \centering
        \includegraphics[width=\linewidth]{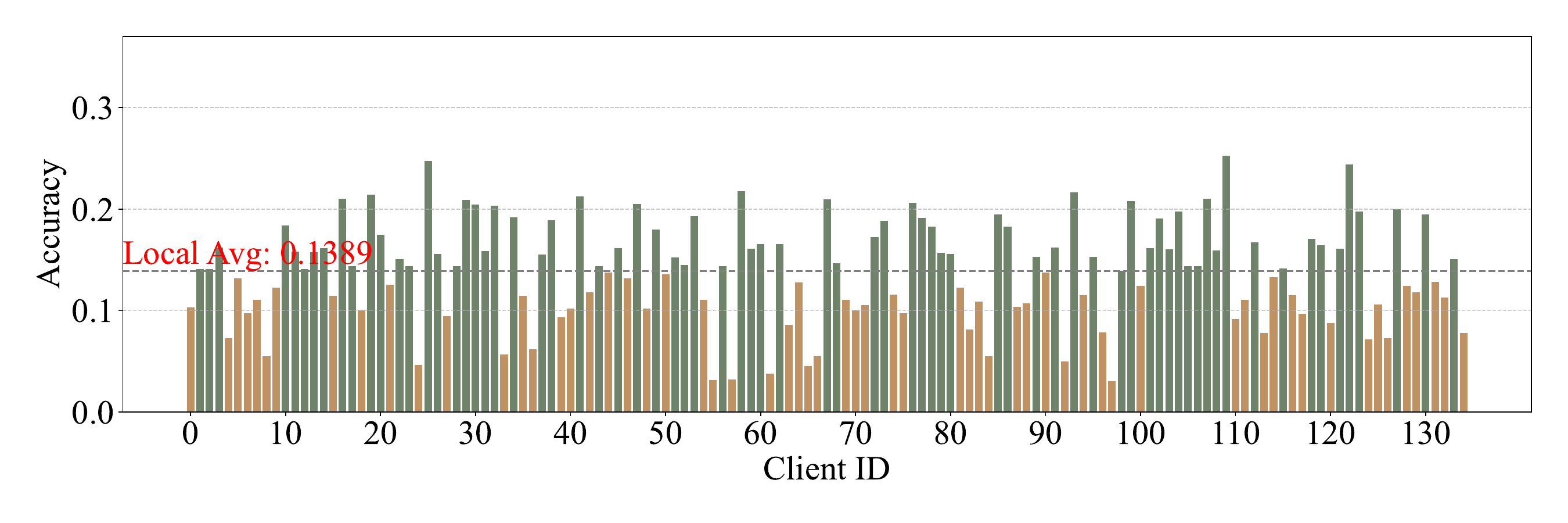} 
        \caption{Local accuracy of CNN on FedRS NIID-1 partition. Testing is performed on $T_I$.}
    \end{subfigure}
        \hfill 
    \begin{subfigure}[b]{0.48\textwidth} 
        \centering
        \includegraphics[width=\linewidth]{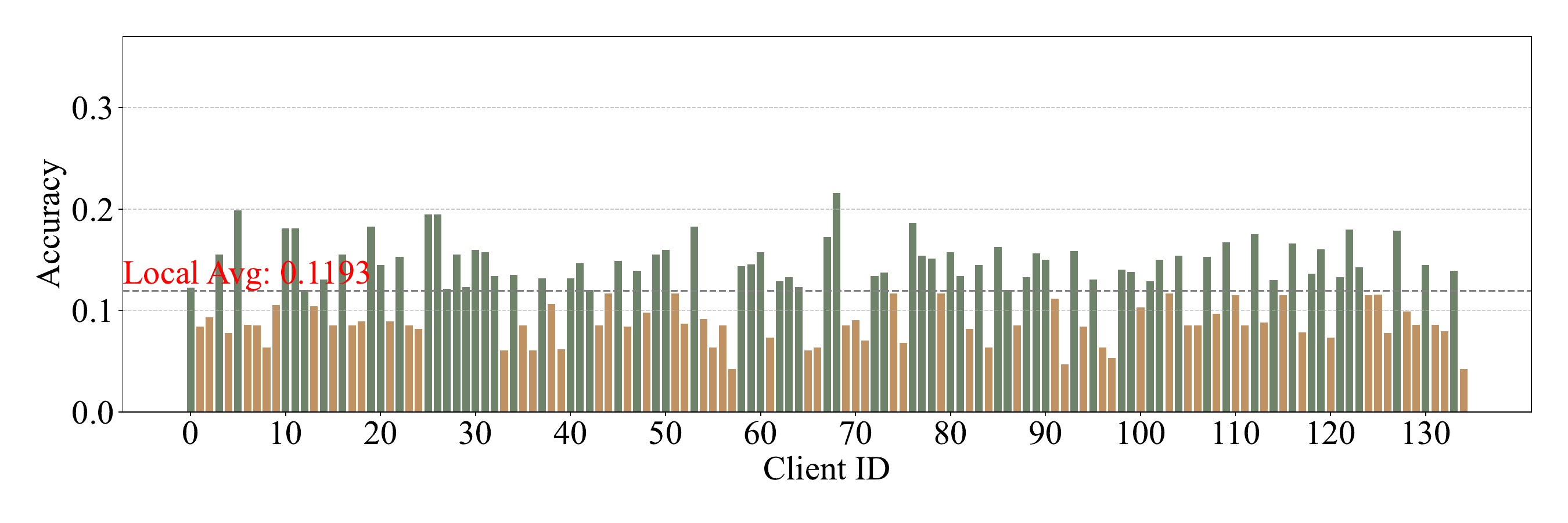} 
        \caption{Local accuracy of CNN on FedRS NIID-1 partition. Testing is performed on $T_B$.}
        \label{fig:source}
    \end{subfigure}

    \vspace{\baselineskip} 
    \begin{subfigure}[b]{0.48\textwidth} 
        \centering
        \includegraphics[width=\linewidth]{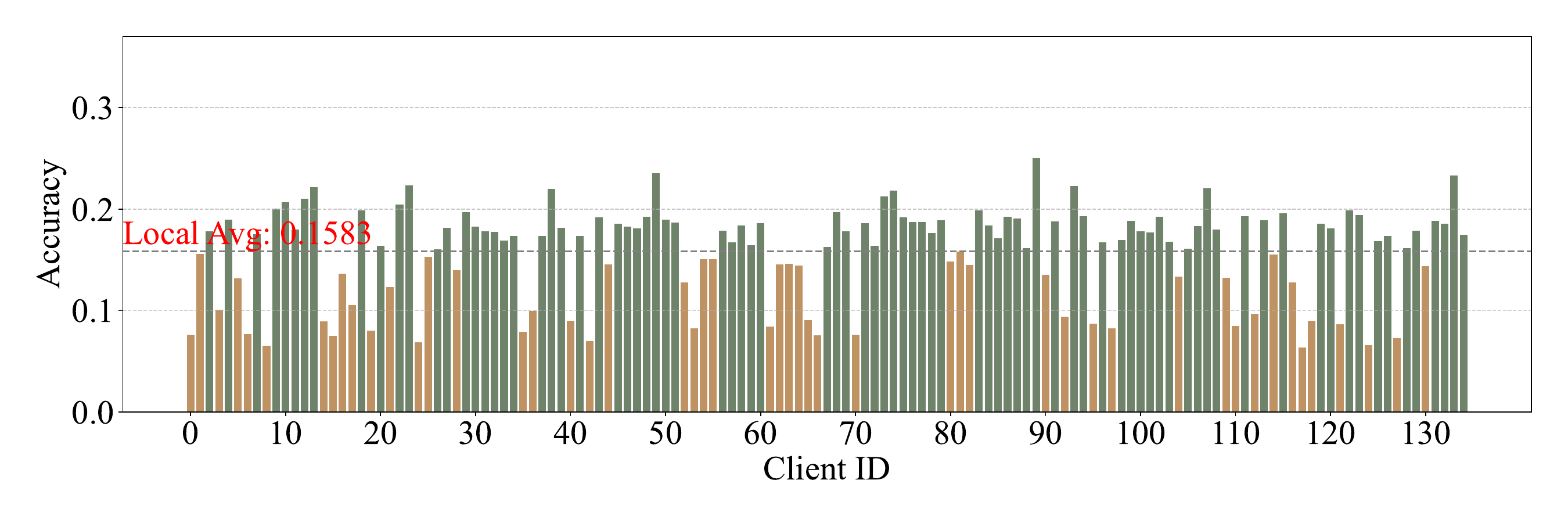} 
        \caption{Local accuracy of CNN on FedRS NIID-2 partition. Testing is performed on $T_I$.}
    \end{subfigure}
        \hfill 
    \begin{subfigure}[b]{0.48\textwidth} 
        \centering
        \includegraphics[width=\linewidth]{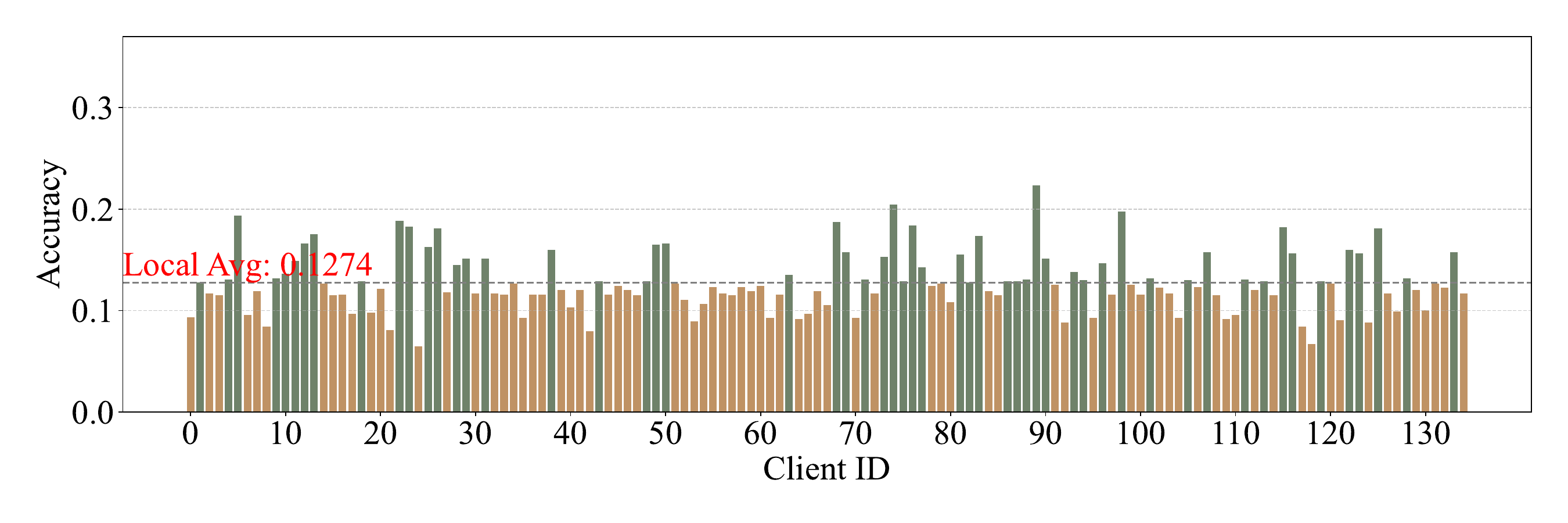} 
        \caption{Local accuracy of CNN on FedRS NIID-2 partition. Testing is performed on $T_B$.}
    \end{subfigure}
    \caption{Local training accuracies of 135 clients on FedRS.}
    \label{fig:local-rs15} 
\end{figure}

\begin{figure}[!tp] 
    \centering 
    \begin{subfigure}[b]{0.48\textwidth} 
        \centering
        \includegraphics[width=\linewidth]{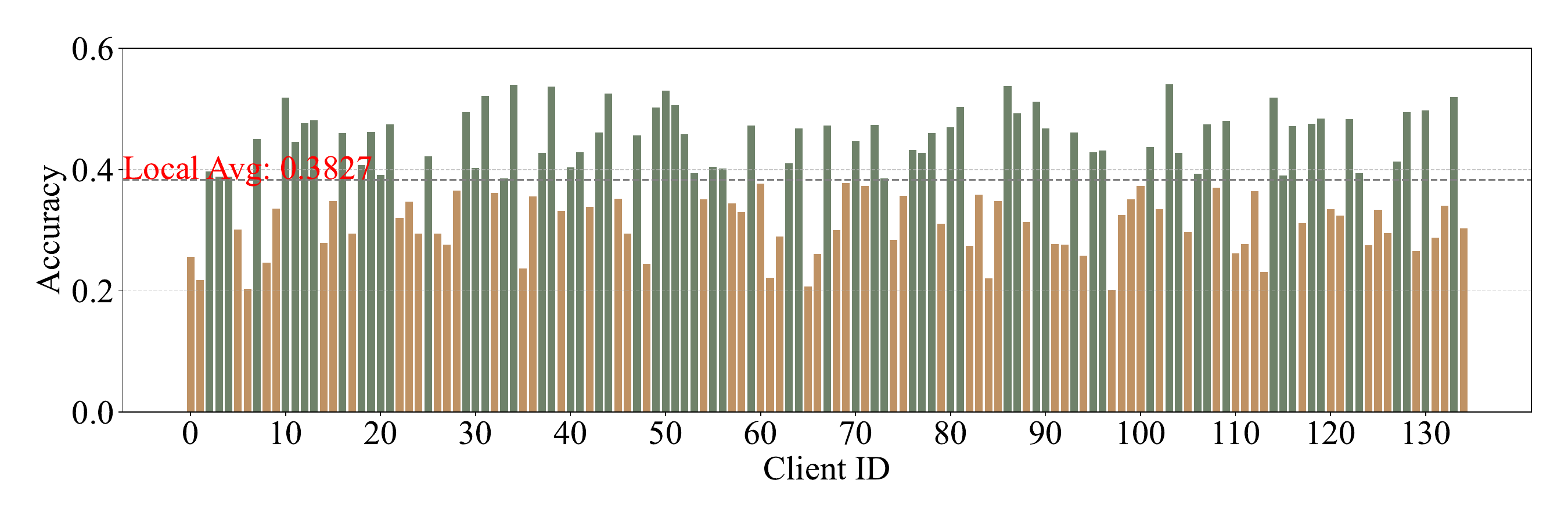} 
        \caption{Local accuracy of ResNet18 on FedRS-5 NIID-1 partition. Testing is performed on $T_I$.}
    \end{subfigure}
        \hfill 
    \begin{subfigure}[b]{0.48\textwidth} 
        \centering
        \includegraphics[width=\linewidth]{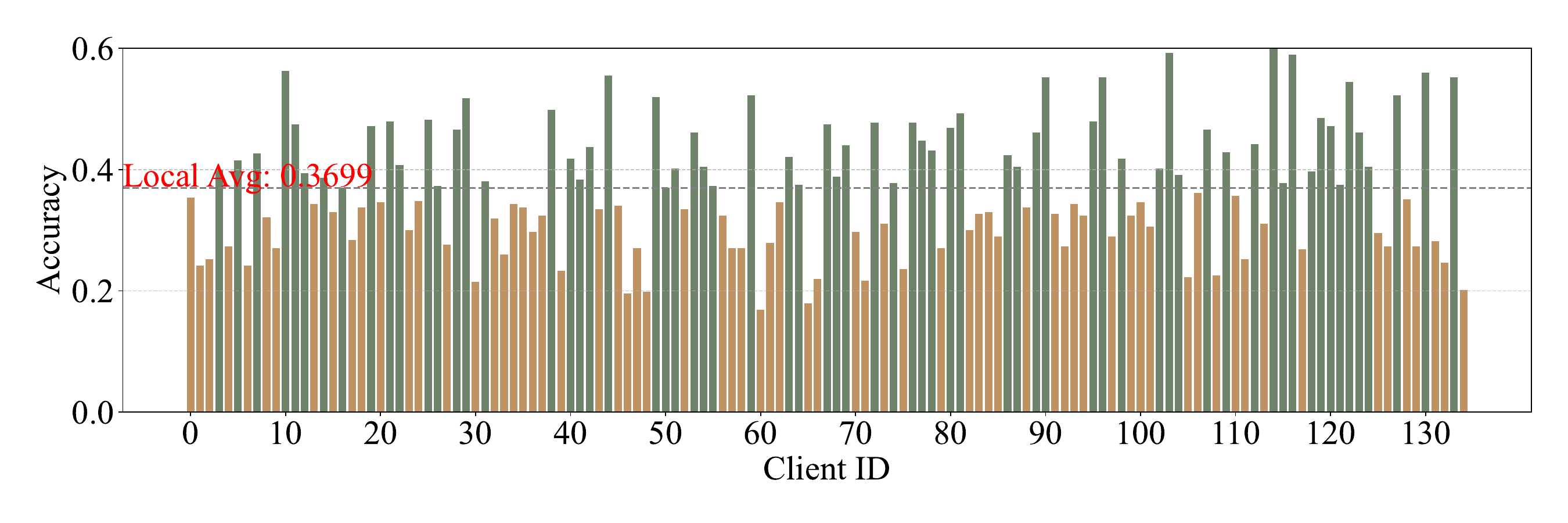} 
        \caption{Local accuracy of ResNet18 on FedRS-5 NIID-1 partition. Testing is performed on $T_B$.}
    \end{subfigure}

    \vspace{\baselineskip} 
    \begin{subfigure}[b]{0.48\textwidth} 
        \centering
        \includegraphics[width=\linewidth]{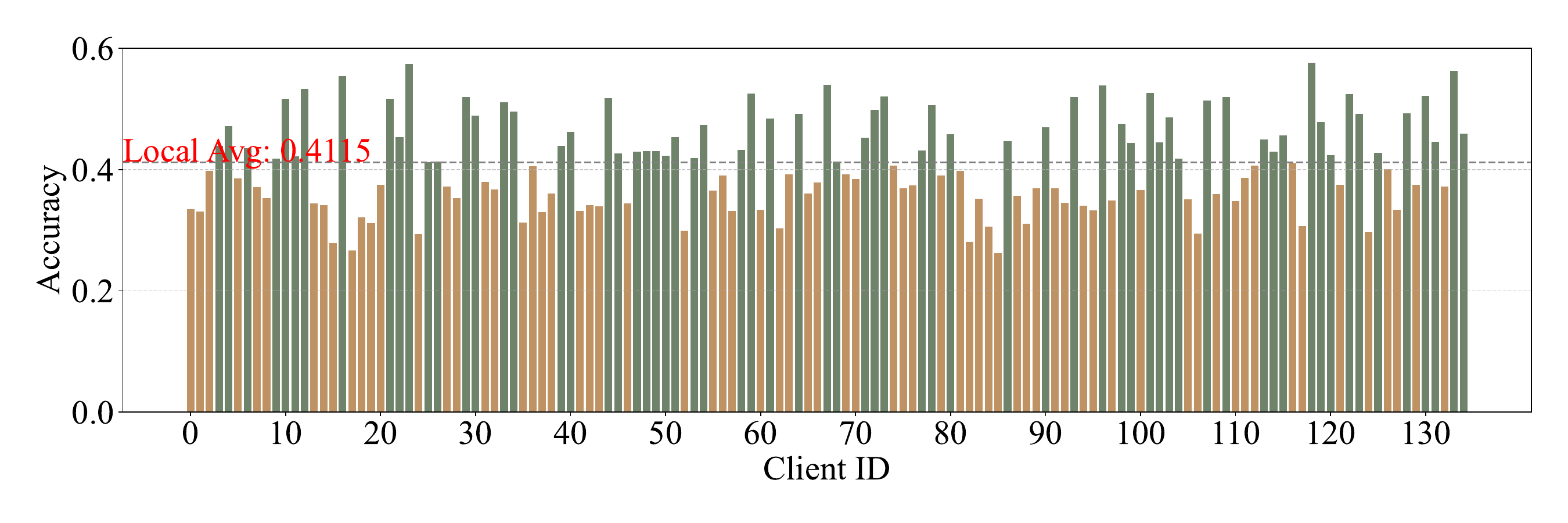} 
        \caption{Local accuracy of ResNet18 on FedRS-5 NIID-2 partition. Testing is performed on $T_I$.}
    \end{subfigure}
        \hfill 
    \begin{subfigure}[b]{0.48\textwidth} 
        \centering
        \includegraphics[width=\linewidth]{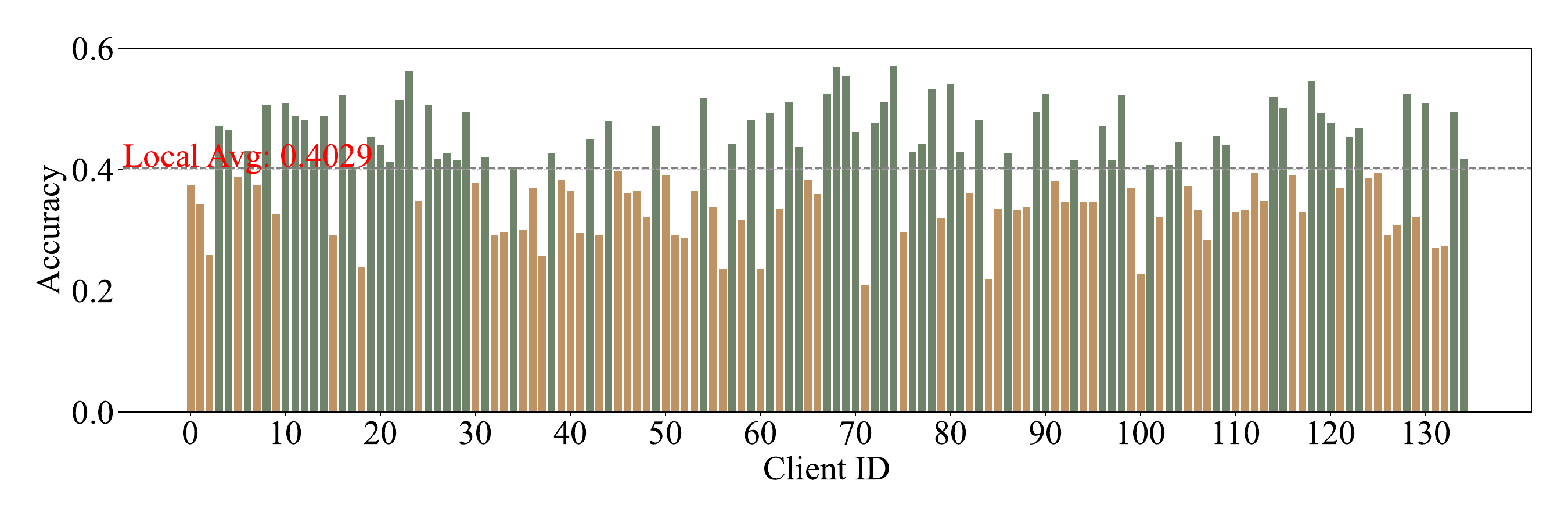} 
        \caption{Local accuracy of ResNet18 on FedRS-5 NIID-2 partition. Testing is performed on $T_B$.}
    \end{subfigure}
    \vspace{\baselineskip} 
        \begin{subfigure}[b]{0.48\textwidth} 
        \centering
        \includegraphics[width=\linewidth]{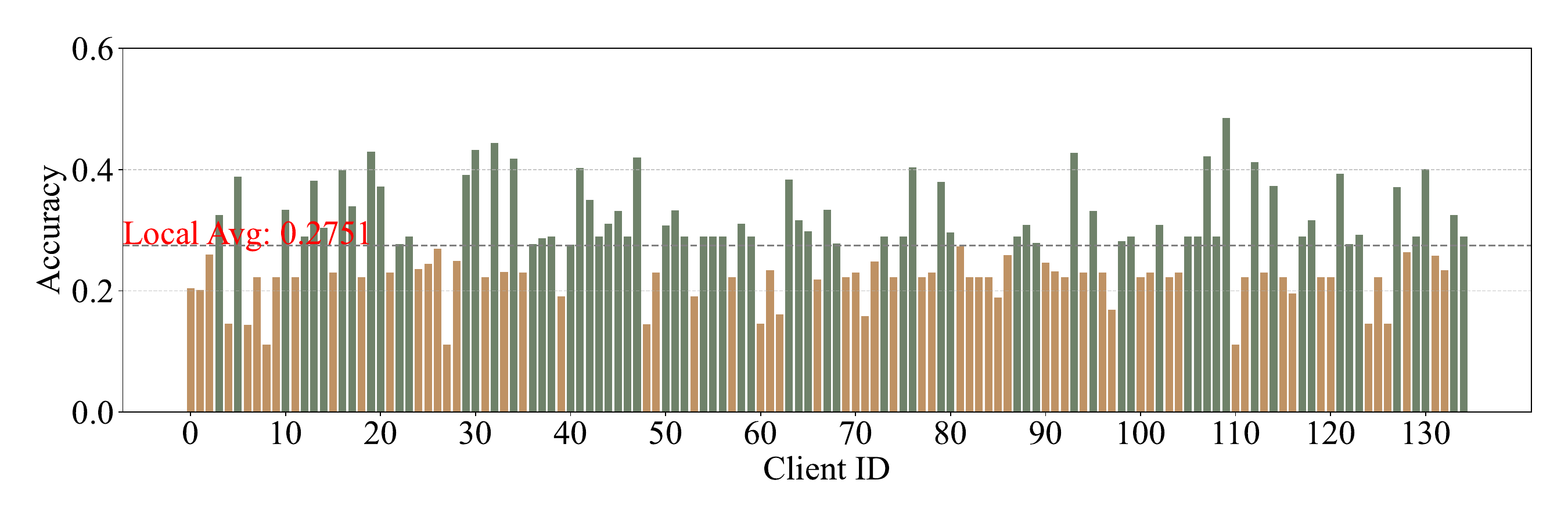} 
        \caption{Local accuracy of CNN on FedRS-5 NIID-1 partition. Testing is performed on $T_I$.}
    \end{subfigure}
        \hfill 
    \begin{subfigure}[b]{0.48\textwidth} 
        \centering
        \includegraphics[width=\linewidth]{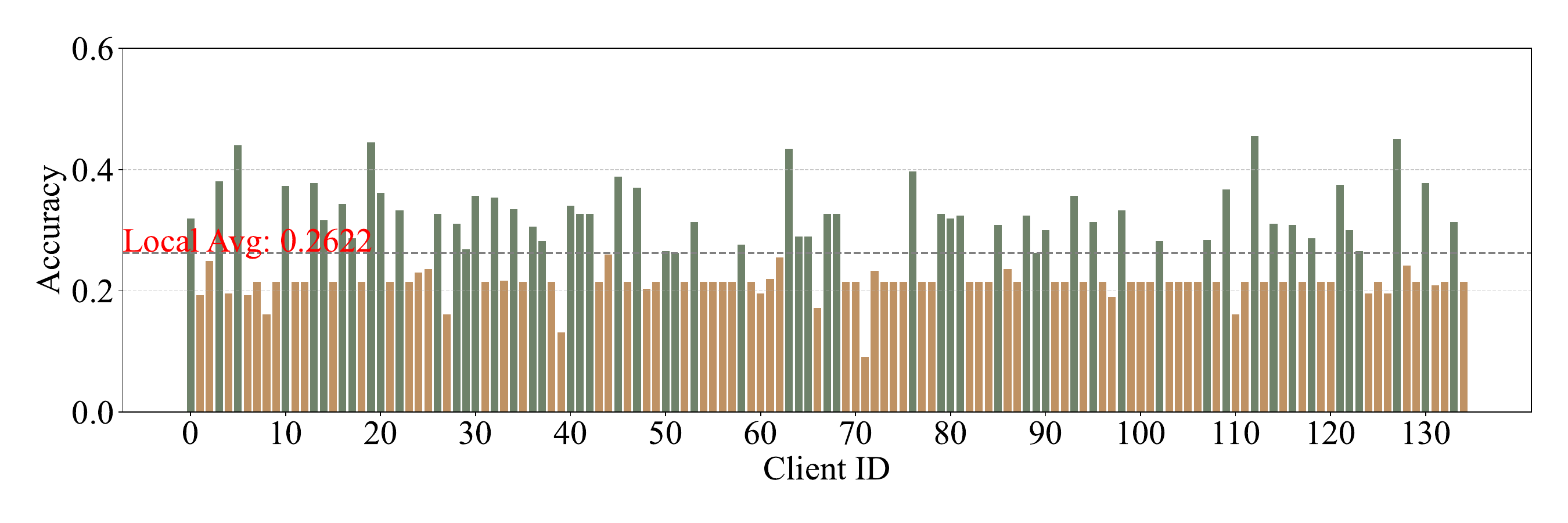} 
        \caption{Local accuracy of CNN on FedRS-5 NIID-1 partition. Testing is performed on $T_B$.}
        \label{fig:source}
    \end{subfigure}

    \vspace{\baselineskip} 
    \begin{subfigure}[b]{0.48\textwidth} 
        \centering
        \includegraphics[width=\linewidth]{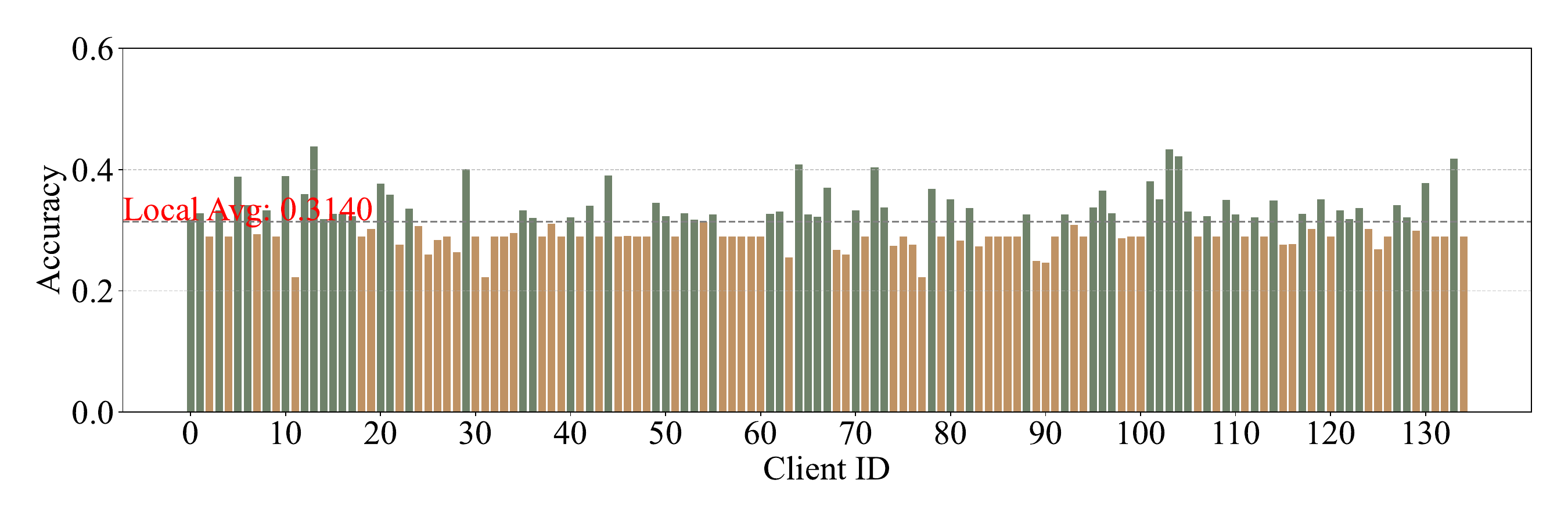} 
        \caption{Local accuracy of CNN on FedRS-5 NIID-2 partition. Testing is performed on $T_I$.}
    \end{subfigure}
        \hfill 
    \begin{subfigure}[b]{0.48\textwidth} 
        \centering
        \includegraphics[width=\linewidth]{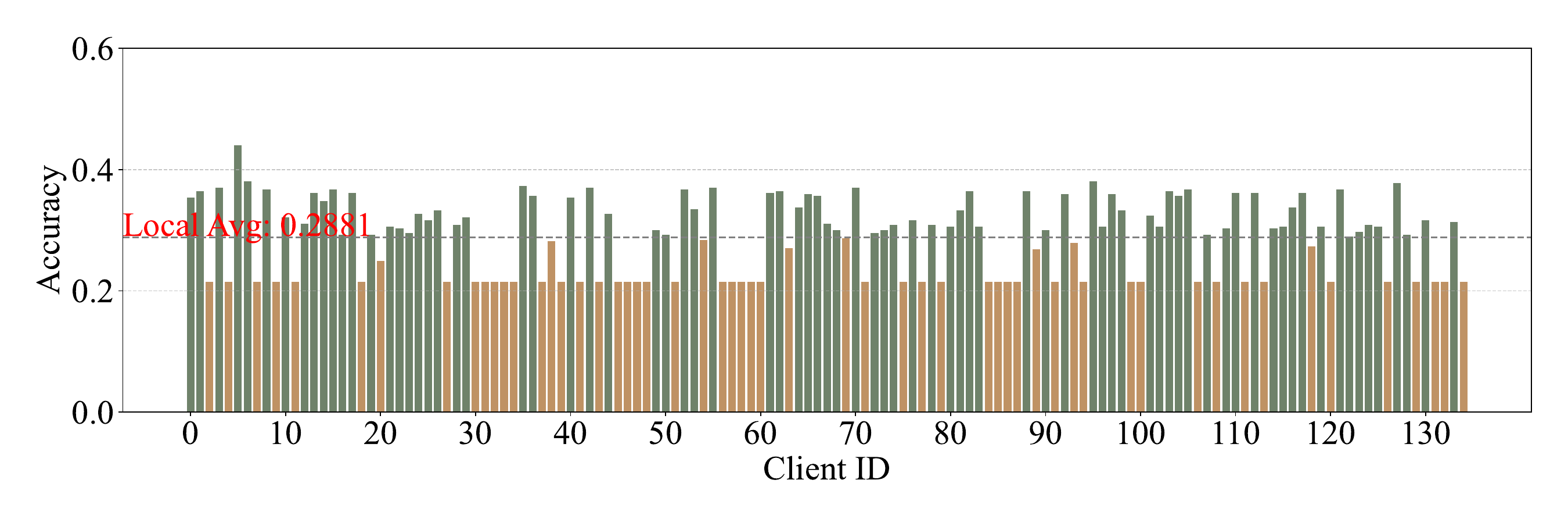} 
        \caption{Local accuracy of CNN on FedRS-5 NIID-2 partition. Testing is performed on $T_B$.}
    \end{subfigure}
    \caption{Local training accuracies of 135 clients on FedRS-5.}
    \label{fig:local-rs5} 
\end{figure}

Figure~\ref{fig:local-rs15} and Figure~\ref{fig:local-rs5} present the local training accuracies for each of the 135 clients on the FedRS and FedRS-5 datasets, respectively.  These figures cover experiments using both CNN and ResNet18 architectures, under NIID-1 and NIID-2 data partition settings, with evaluations performed on both the imbalanced test set $T_I$ and the balanced test set $T_B$. The results demonstrate that due to the significant data heterogeneity across clients (variations in data quantity, class distributions, and feature spaces), the performance of models trained solely on local data is generally limited. Furthermore, there is considerable variance in the performance of local models across different clients, indicating that the local optimization objectives and resulting model quality can differ substantially from one client to another.

These findings highlight two key points:
\begin{itemize}
    \item The necessity of federated learning to leverage diverse data from multiple clients for improved model generalization. 
    \item The significant challenge that federated learning algorithms face due to disparate local data distributions and potentially conflicting local optimization directions. 
\end{itemize}

\subsection{FL Learning Curves}
\label{sec:curves}
\begin{figure}[!tp] 
    \centering 
    \begin{subfigure}[b]{0.48\textwidth} 
        \centering
        \includegraphics[width=\linewidth]{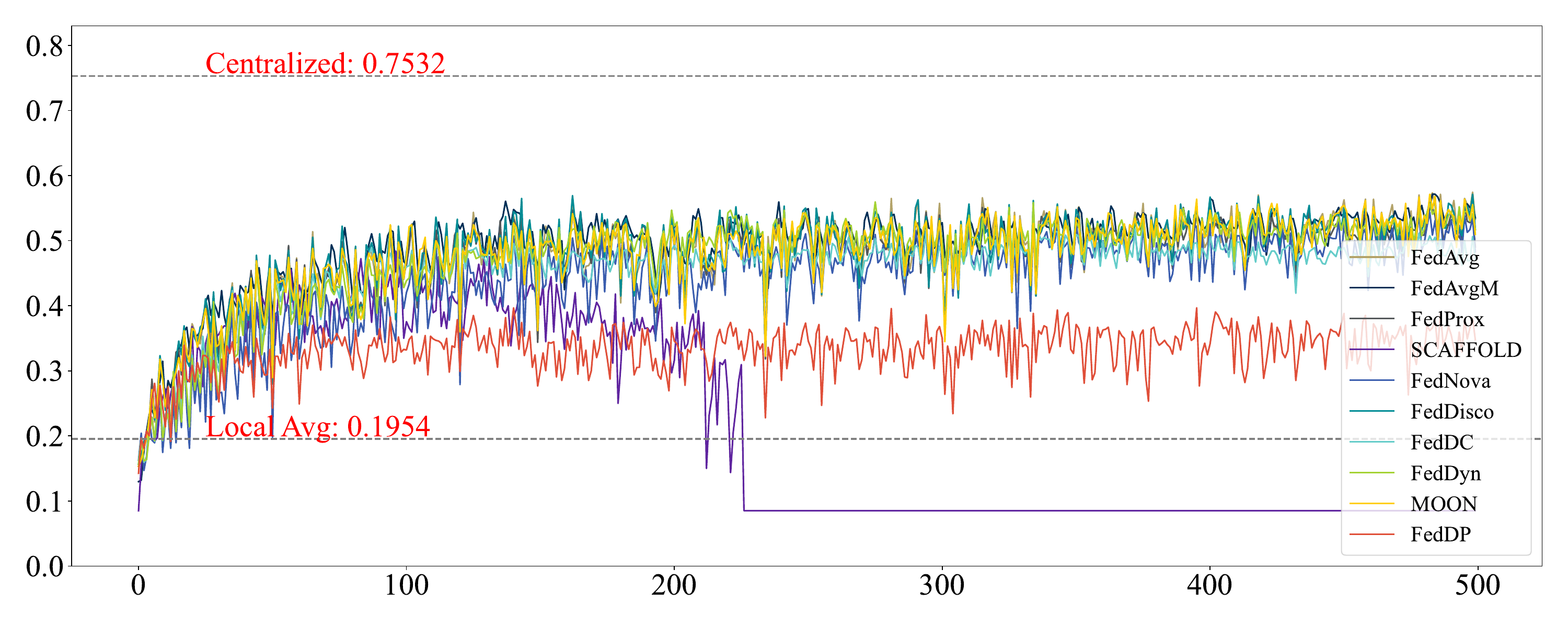} 
        \caption{Accuracy curve of CNN on FedRS NIID-1 partition. Testing is performed on $T_B$.}
        \label{fig:label} 
    \end{subfigure}
        \hfill 
    \begin{subfigure}[b]{0.48\textwidth} 
        \centering
        \includegraphics[width=\linewidth]{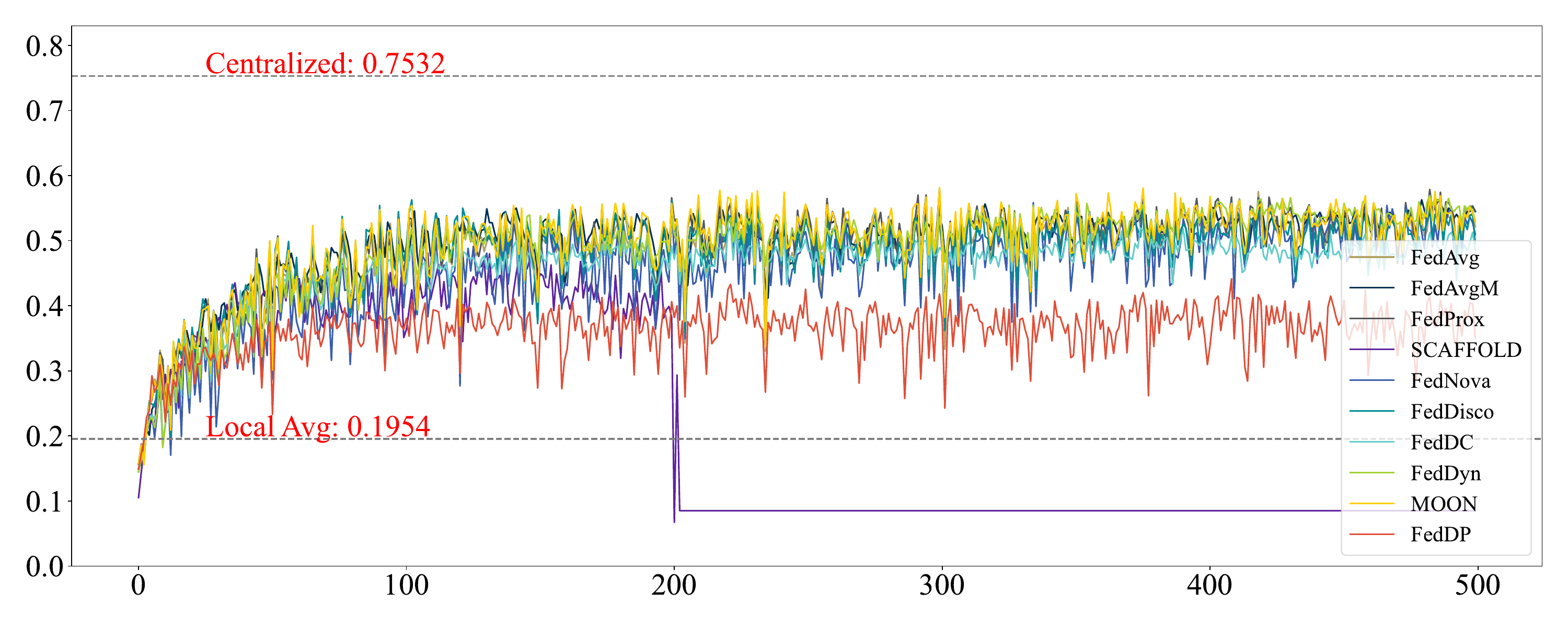} 
        \caption{Accuracy curve of CNN on FedRS NIID-2 partition. Testing is performed on $T_B$.}
        \label{fig:source}
    \end{subfigure}

    \vspace{\baselineskip} 
    \begin{subfigure}[b]{0.48\textwidth} 
        \centering
        \includegraphics[width=\linewidth]{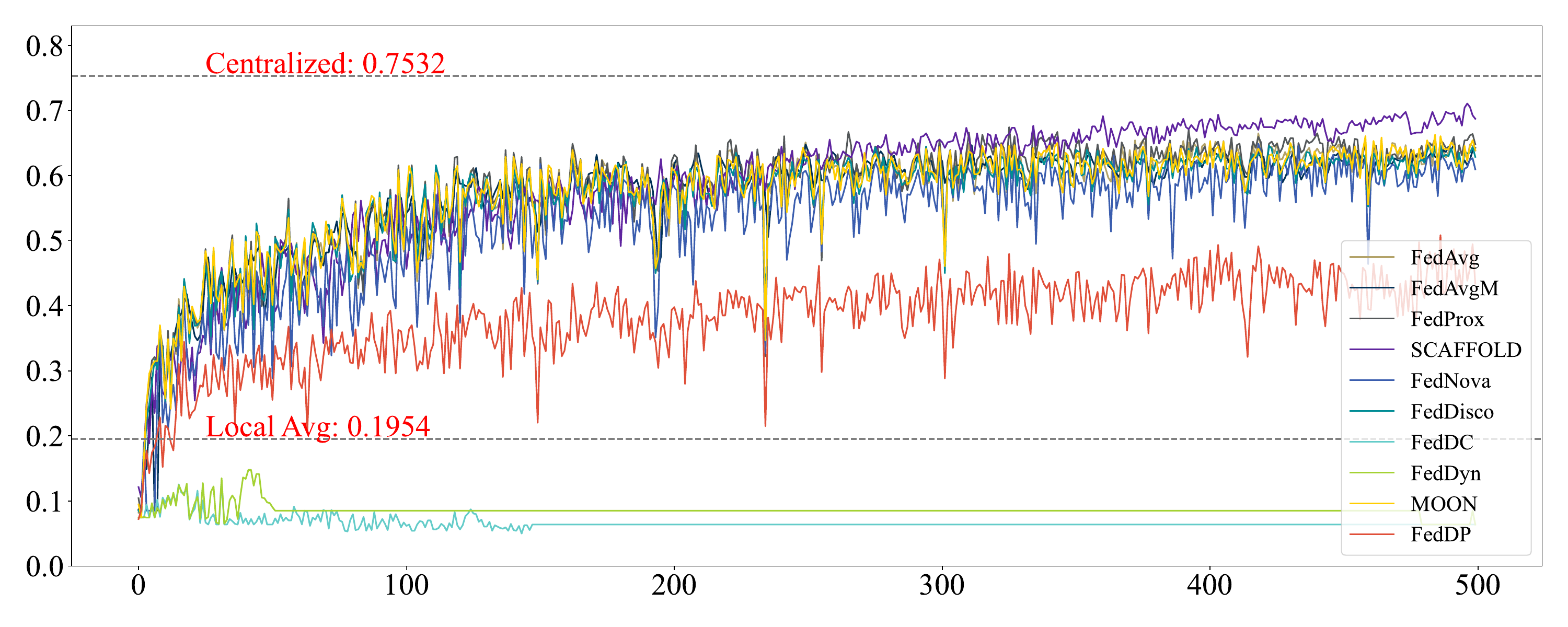} 
        \caption{Accuracy curve of ResNet18 on FedRS NIID-1 partition. Testing is performed on $T_B$.}
        \label{fig:label} 
    \end{subfigure}
        \hfill 
    \begin{subfigure}[b]{0.48\textwidth} 
        \centering
        \includegraphics[width=\linewidth]{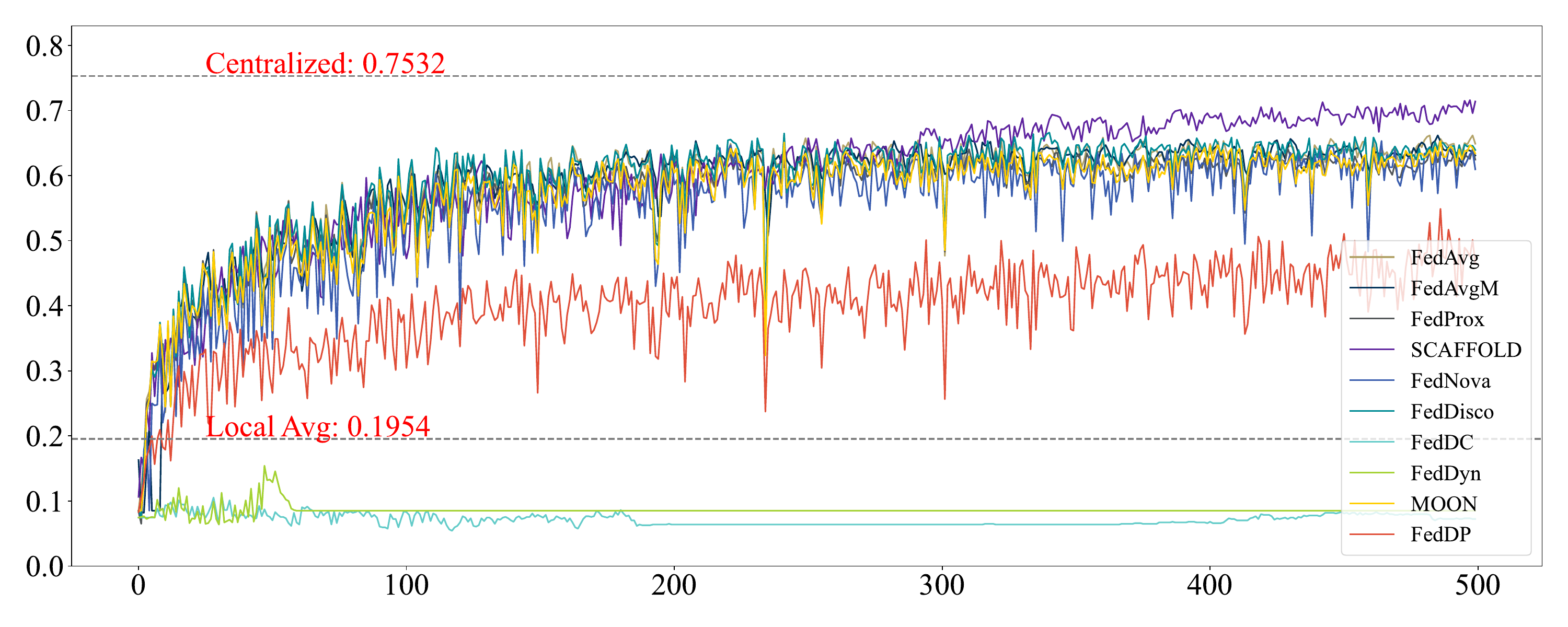} 
        \caption{Accuracy curve of ResNet18 on FedRS NIID-2 partition. Testing is performed on $T_B$.}
        \label{fig:source}
    \end{subfigure}
    \caption{Test accuracy curve of FL baseline method during the communication round of training.}
    \label{fig:curve-balance} 
\end{figure}

\begin{figure}[!tp] 
    \centering 
    \begin{subfigure}[b]{0.48\textwidth} 
        \centering
        \includegraphics[width=\linewidth]{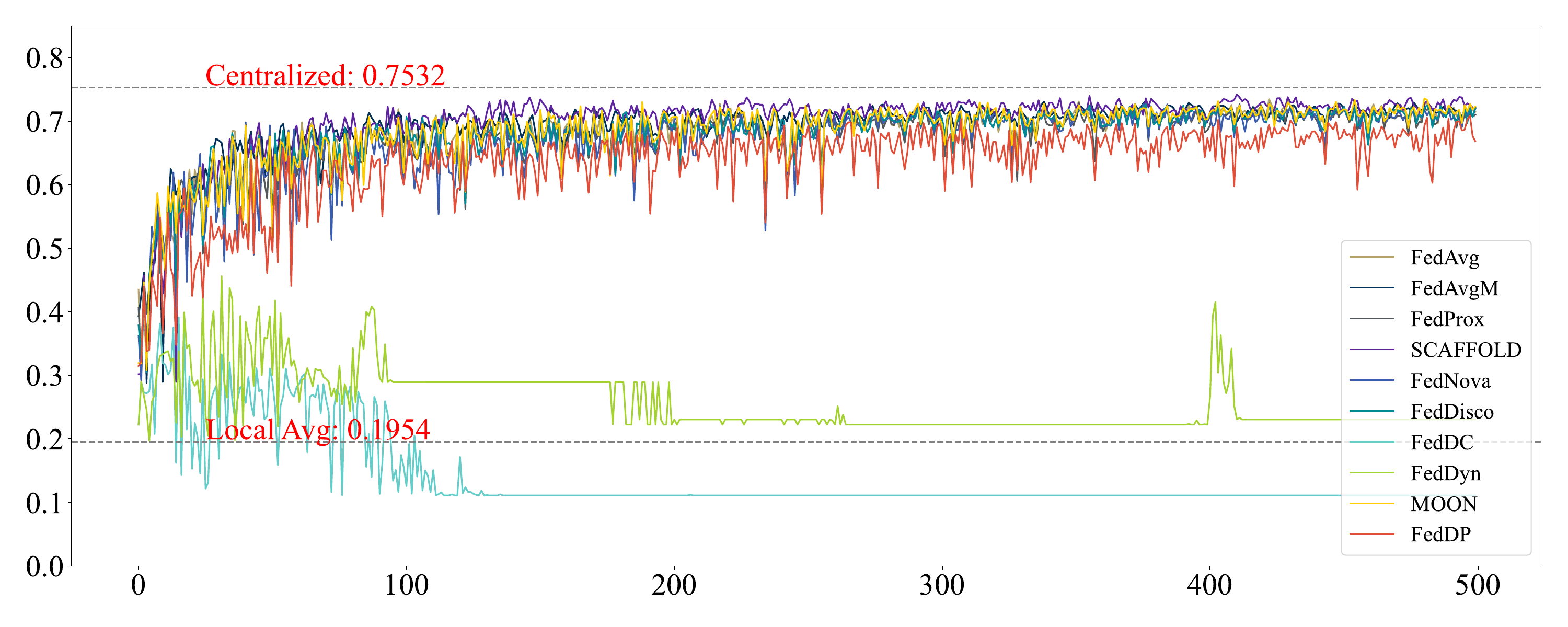} 
        \caption{Accuracy curve of ResNet18 on FedRS-5 NIID-1 partition. Testing is performed on $T_I$.}
        \label{fig:label} 
    \end{subfigure}
        \hfill 
    \begin{subfigure}[b]{0.48\textwidth} 
        \centering
        \includegraphics[width=\linewidth]{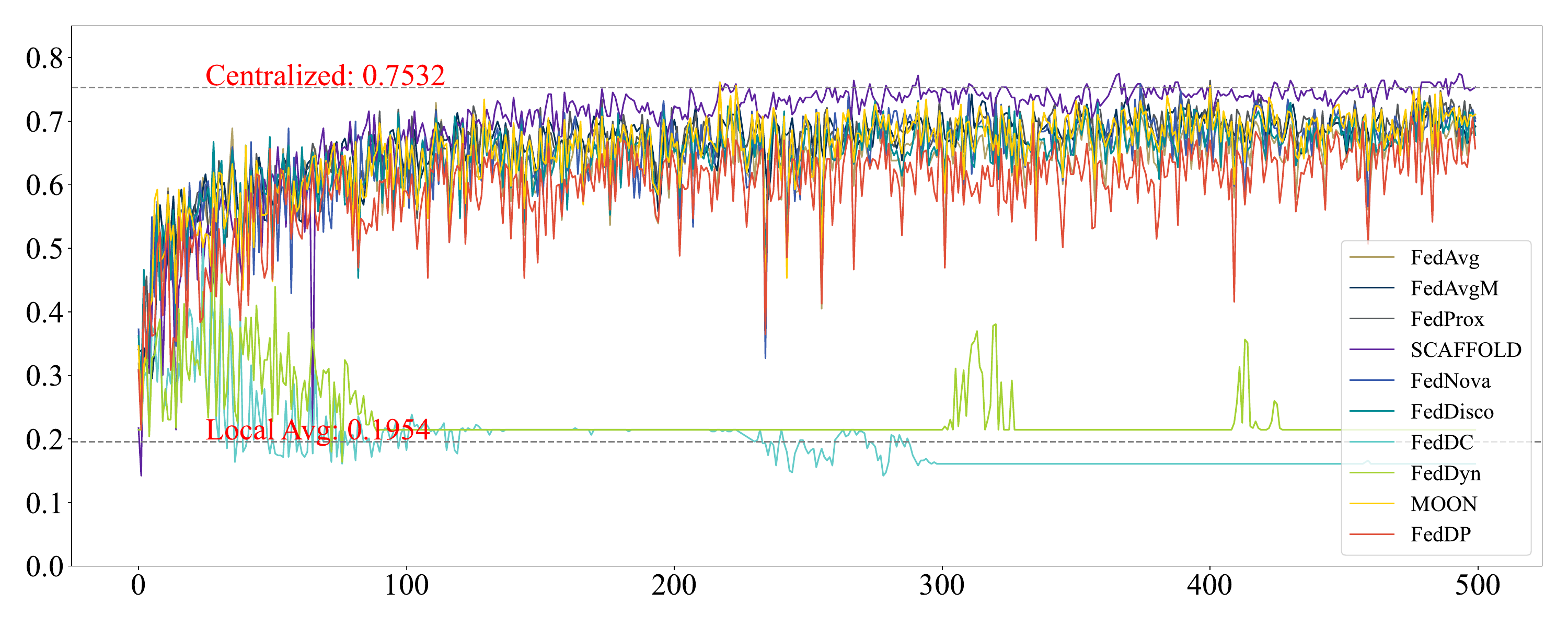} 
        \caption{Accuracy curve of ResNet18 on FedRS-5 NIID-1 partition. Testing is performed on $T_B$.}
        \label{fig:source}
    \end{subfigure}

    \vspace{\baselineskip} 
    \begin{subfigure}[b]{0.48\textwidth} 
        \centering
        \includegraphics[width=\linewidth]{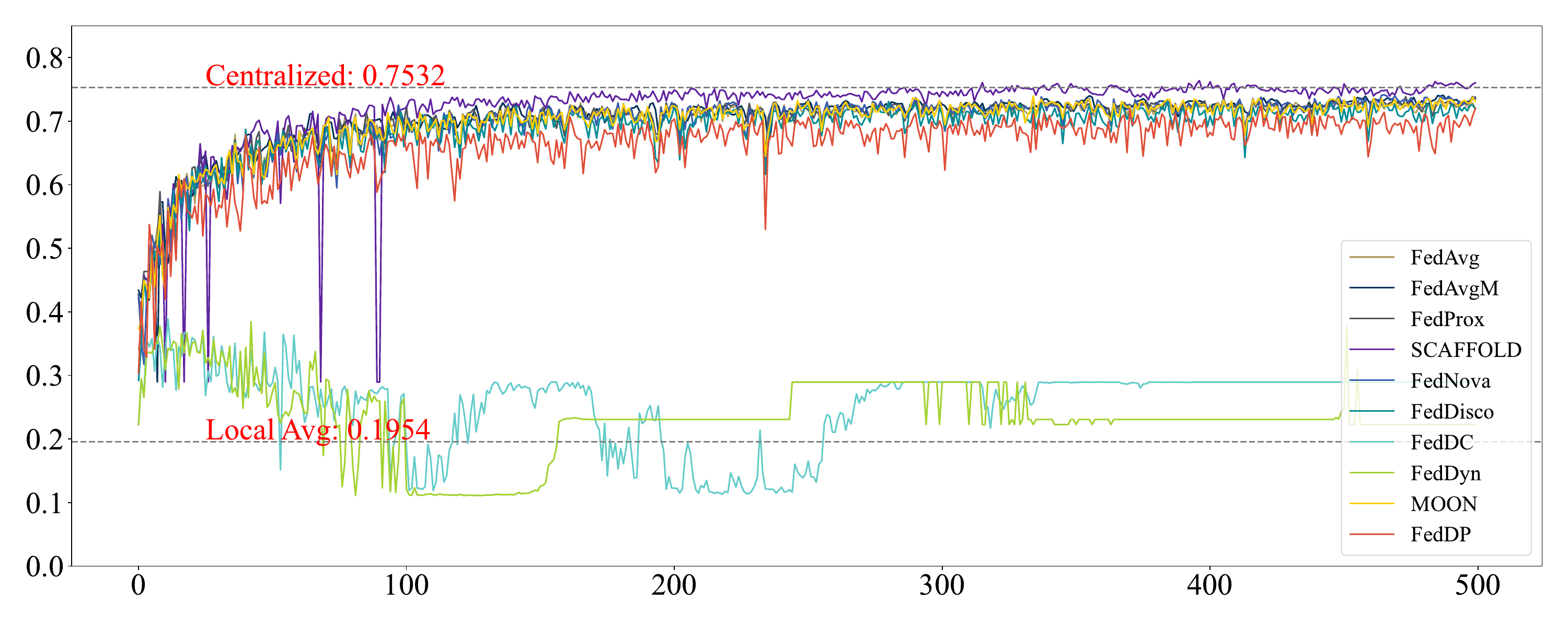} 
        \caption{Accuracy curve of ResNet18 on FedRS-5 NIID-2 partition. Testing is performed on $T_I$.}
        \label{fig:label} 
    \end{subfigure}
        \hfill 
    \begin{subfigure}[b]{0.48\textwidth} 
        \centering
        \includegraphics[width=\linewidth]{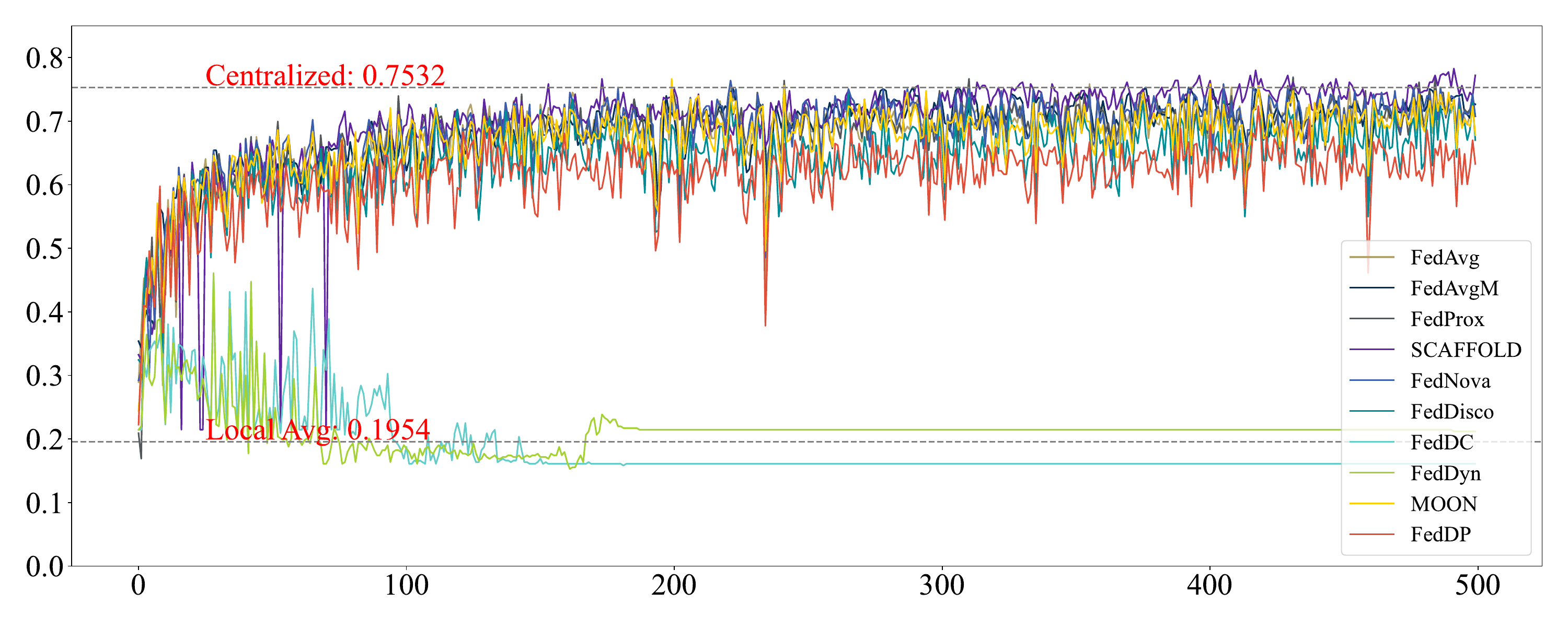} 
        \caption{Accuracy curve of ResNet18 on FedRS-5 NIID-2 partition. Testing is performed on $T_B$.}
        \label{fig:source}
    \end{subfigure}
    \vspace{\baselineskip} 
        \begin{subfigure}[b]{0.48\textwidth} 
        \centering
        \includegraphics[width=\linewidth]{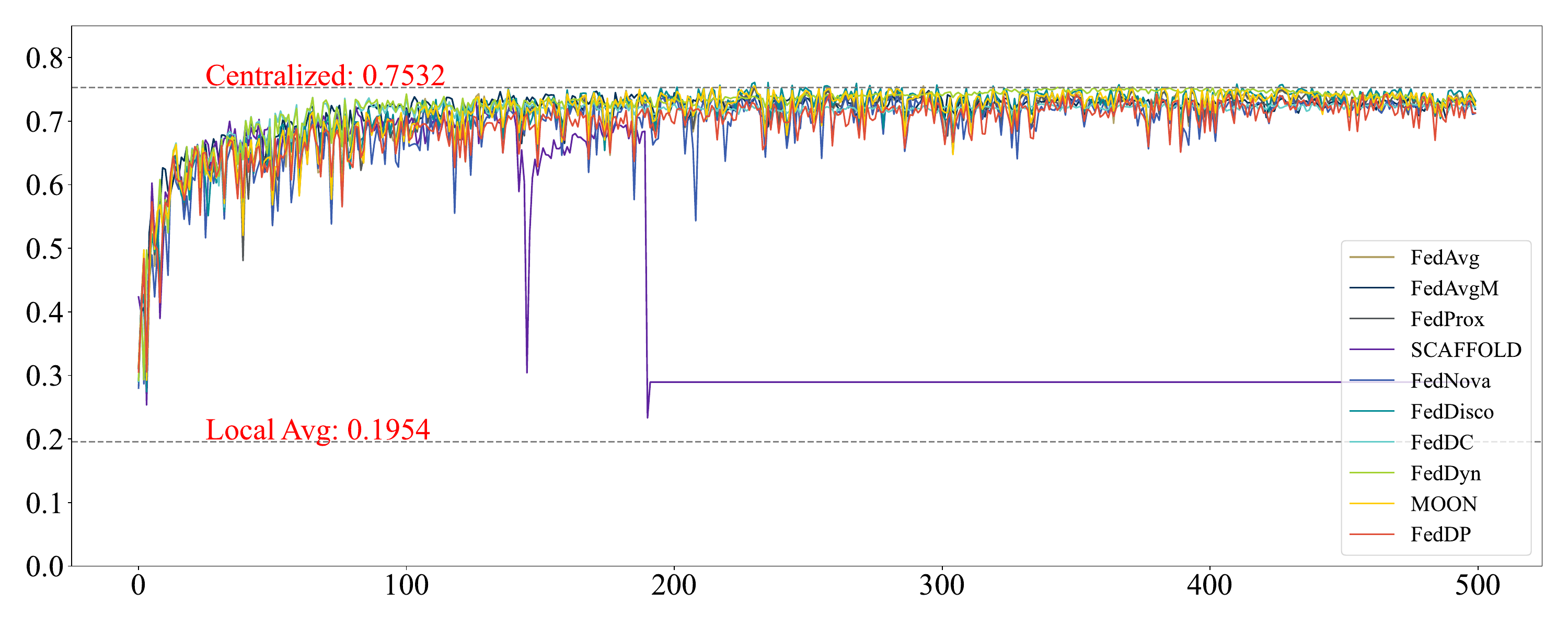} 
        \caption{Accuracy curve of CNN on FedRS-5 NIID-1 partition. Testing is performed on $T_I$.}
        \label{fig:label} 
    \end{subfigure}
        \hfill 
    \begin{subfigure}[b]{0.48\textwidth} 
        \centering
        \includegraphics[width=\linewidth]{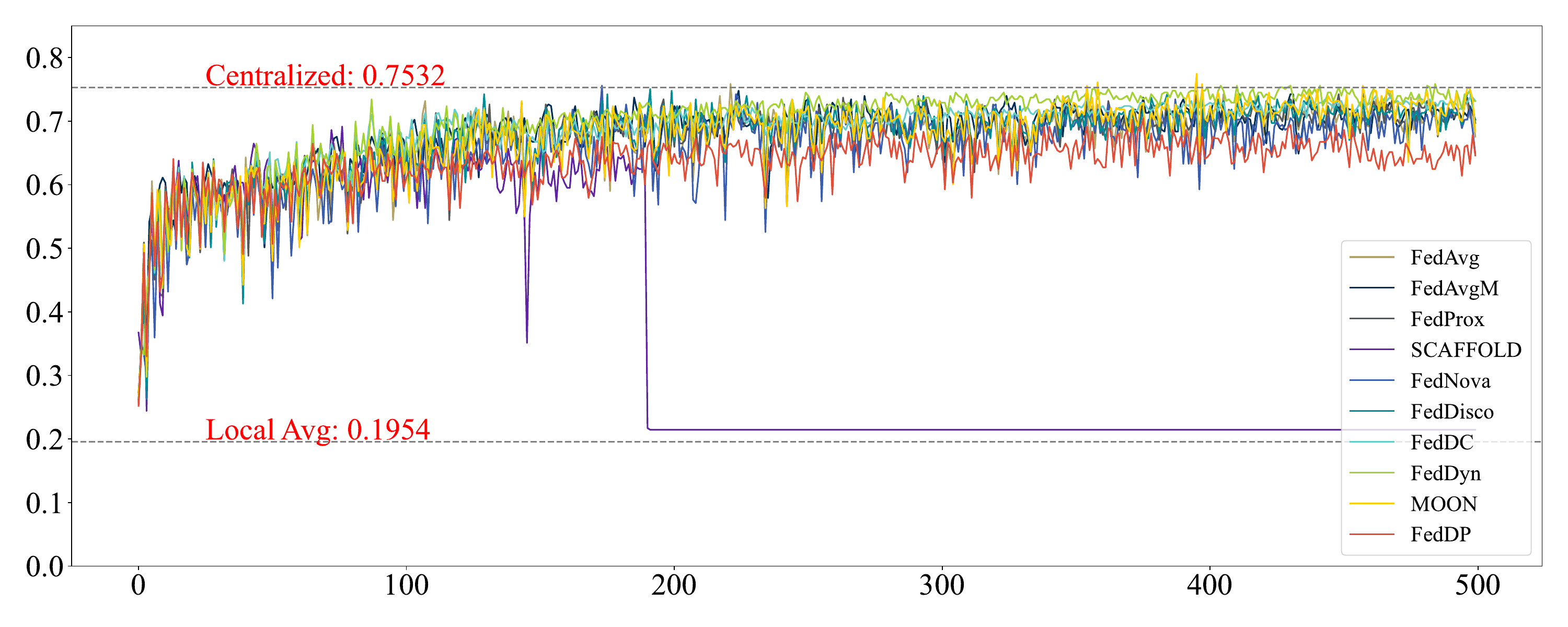} 
        \caption{Accuracy curve of CNN on FedRS-5 NIID-1 partition. Testing is performed on $T_B$.}
        \label{fig:source}
    \end{subfigure}

    \vspace{\baselineskip} 
    \begin{subfigure}[b]{0.48\textwidth} 
        \centering
        \includegraphics[width=\linewidth]{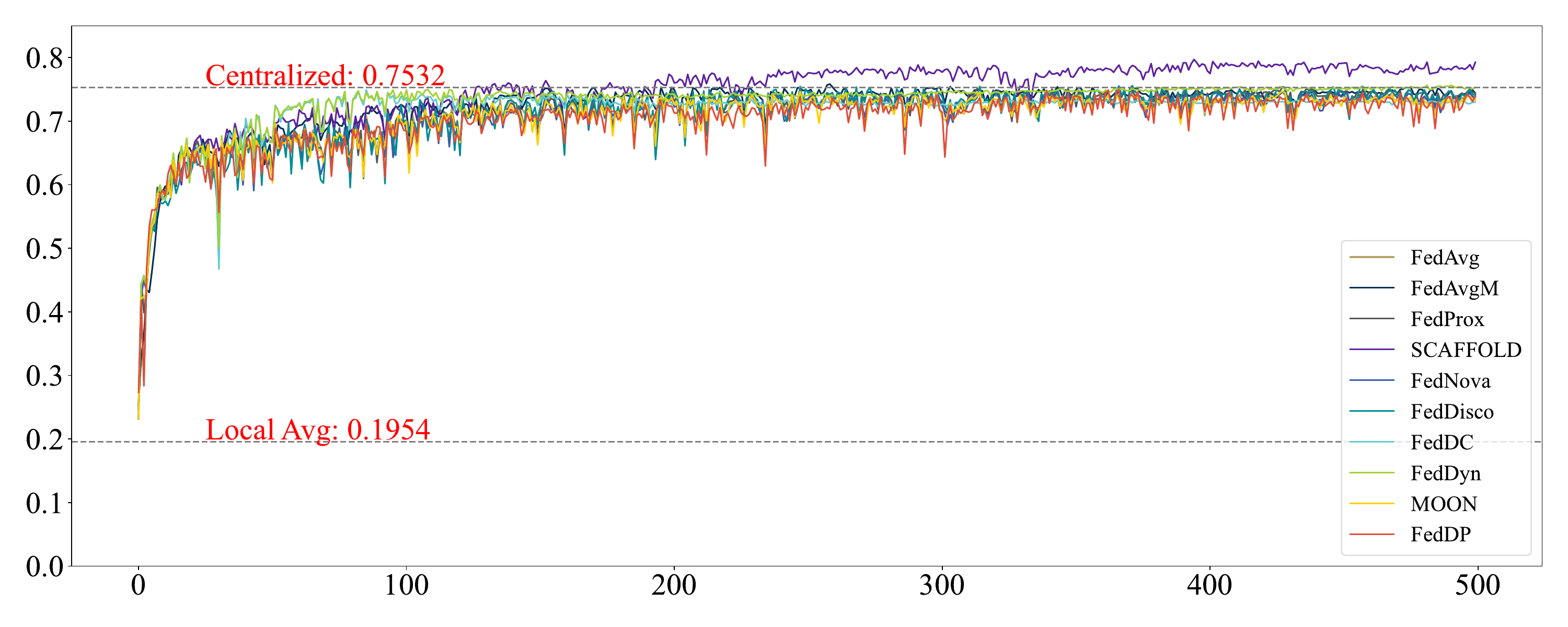} 
        \caption{Accuracy curve of CNN on FedRS-5 NIID-2 partition. Testing is performed on $T_I$.}
        \label{fig:label} 
    \end{subfigure}
        \hfill 
    \begin{subfigure}[b]{0.48\textwidth} 
        \centering
        \includegraphics[width=\linewidth]{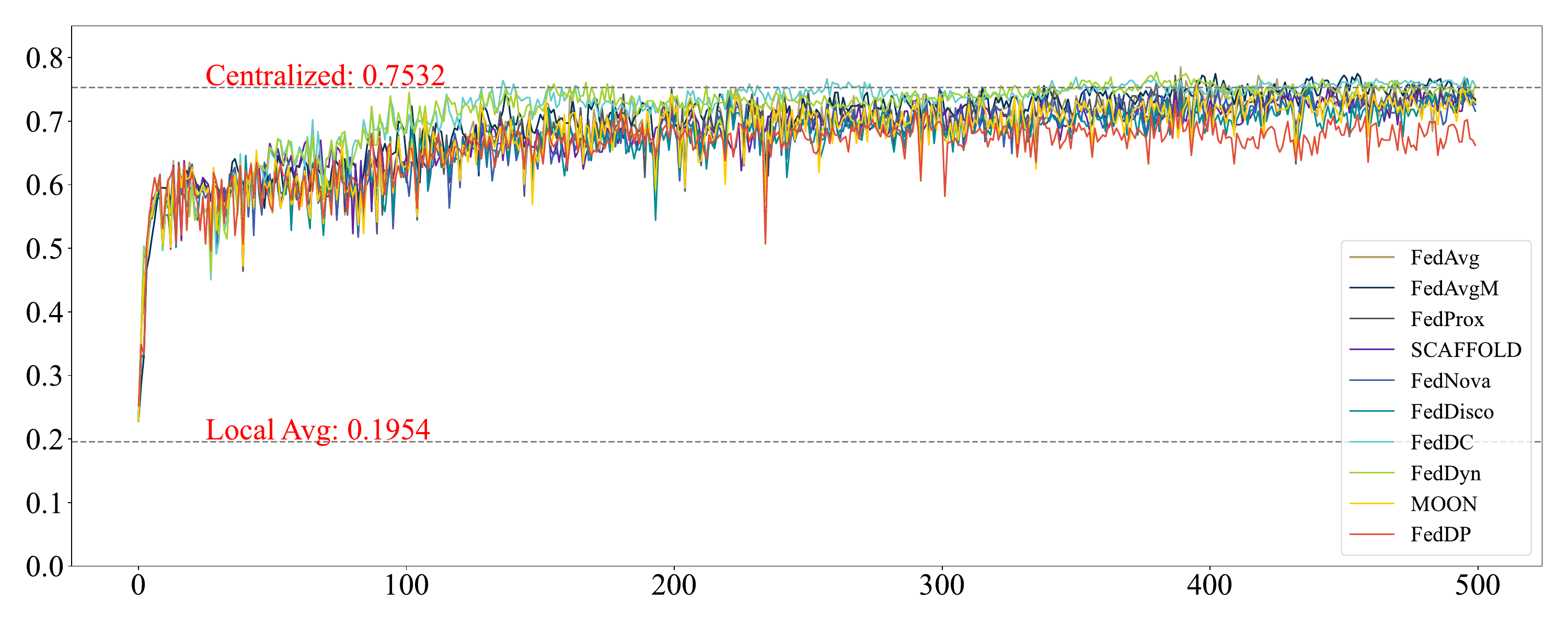} 
        \caption{Accuracy curve of CNN on FedRS-5 NIID-2 partition. Testing is performed on $T_B$.}
        \label{fig:source}
    \end{subfigure}
    \caption{Test accuracy curve of FL baseline method during the communication round of training.}
    \label{fig:curve-RS5} 
\end{figure}

Here we provide a detailed presentation and discussion of the learning curves (test accuracy vs. communication rounds) for the various FL algorithms benchmarked. These curves offer insights into the convergence behavior, stability, and learning speed of different FL methods under diverse experimental conditions. Figure~\ref{fig:curve-balance} and Figure~\ref{fig:curve-RS5} 

Figure~\ref{fig:curve-balance} complements Figure~\ref{fig:curve} by showing test accuracy curves for FL baseline methods on the full \texttt{FedRS} dataset when testing is performed on the balanced test set $T_B$. Figure~\ref{fig:curve-RS5} displays the test accuracy curves for FL baseline methods on the \texttt{FedRS-5} dataset across various settings.

\textbf{Discussion of FL Learning Curves:}
The learning curves presented in Figure~\ref{fig:curve}, Figure~\ref{fig:curve-balance}, and Figure~\ref{fig:curve-RS5} illustrate several key aspects of the federated learning process:

$\bullet$ \textbf{General Convergence Trends:} Most FL algorithms demonstrate an increase in test accuracy over communication rounds, indicating successful learning. The speed of convergence and stability vary among methods, with some exhibiting smoother curves and others more fluctuations, particularly under high heterogeneity or with more complex algorithms.

$\bullet$ \textbf{Comparative Performance:} Visual comparison of the convergent accuracy levels of different FL algorithms within each experimental setting reveals their relative effectiveness. Some algorithms consistently perform well across various settings, while others show greater sensitivity to the dataset, model, or data partition. For instance,  SCAFFOLD performs well with ResNet18 but not with the simpler CNN, whereas MOON shows strong performance with the CNN model. In contrast, FedDC and FedDyn exhibit notably lower performance with the ResNet18 model in several scenarios.

$\bullet$ \textbf{Impact of Model Architecture (CNN vs. ResNet18):} The choice of model architecture influences both convergence dynamics and final accuracy. While ResNet18, being a more powerful model, generally achieves higher absolute accuracy, its complexity can also lead to different convergence behaviors and stability compared to the simpler CNN.

$\bullet$ \textbf{Impact of Data Heterogeneity (NIID-1 vs. NIID-2):} The type of non-IID data distribution (Dirichlet-based NIID-1 vs. uniform source-based NIID-2) affects the learning curves. Higher degrees of heterogeneity, often associated with NIID-1, can lead to slower convergence, increased instability, or lower final accuracies for certain FL methods, testing their robustness.

$\bullet$ \textbf{Impact of Test Set ($T_I$ vs. $T_B$):} Comparing the performance in the imbalanced test set $T_I$ (reflecting the proportions of the source data) versus the balanced test set $T_B$ (uniform sampling per category-source combination) helps to assess the generalization of the model. Differences can indicate whether models are biased towards more prevalent classes/sources in the training distribution or achieve balanced performance across all categories.

$\bullet$ \textbf{Comparison with Baselines:} Each plot includes lines for "Centralized" training (often an empirical upper bound) and "Local Avg" (average of local training, typically a lower performance baseline). FL methods consistently outperform "Local Avg", demonstrating the benefits of collaboration. While most FL methods do not reach "Centralized" performance due to communication constraints and data heterogeneity, there are instances, particularly in simpler settings, where FL performance approaches or even slightly exceeds that of centralized training, potentially due to the FL process guiding the model out of poor local optima.

$\bullet$ \textbf{Specific Algorithm Behaviors:} The learning curves also highlight specific characteristics of certain algorithms. For example, FedDP consistently shows lower accuracy due to the computational overhead and information loss inherent in enforcing user-level differential privacy. Furthermore, it is observed that some advanced optimization-based FL algorithms like SCAFFOLD, FedDC, and FedDyn can experience convergence issues or performance collapse when faced with very high levels of data heterogeneity, underscoring the importance of algorithm selection and hyperparameter tuning in realistic, challenging FL scenarios.

\subsection{Partial parameters updates}
\begin{figure}[!tp] 
    \centering 
    \begin{subfigure}[b]{0.48\textwidth} 
        \centering
        \includegraphics[width=\linewidth]{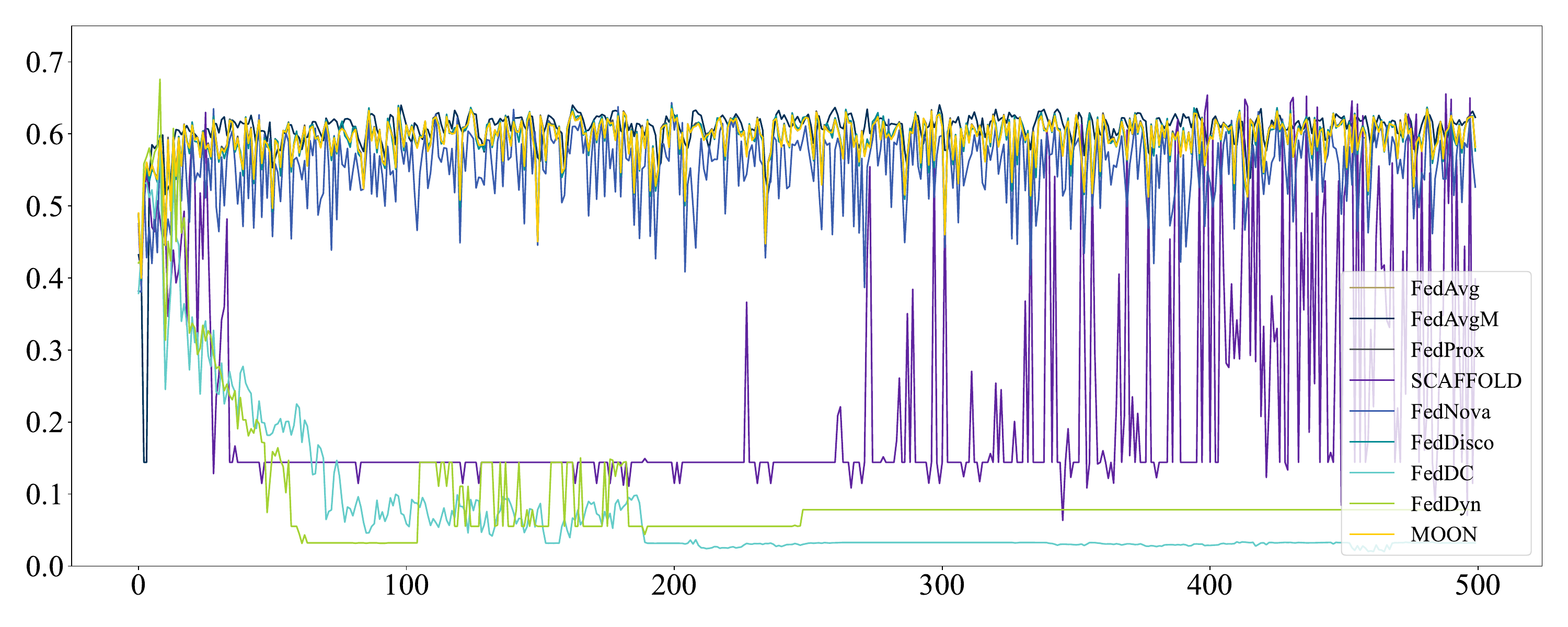} 
        \caption{Accuracy curve of pre-trained ResNet18 on FedRS NIID-1 partition. Testing is performed on $T_I$.}
        \label{fig:label} 
    \end{subfigure}
        \hfill 
    \begin{subfigure}[b]{0.48\textwidth} 
        \centering
        \includegraphics[width=\linewidth]{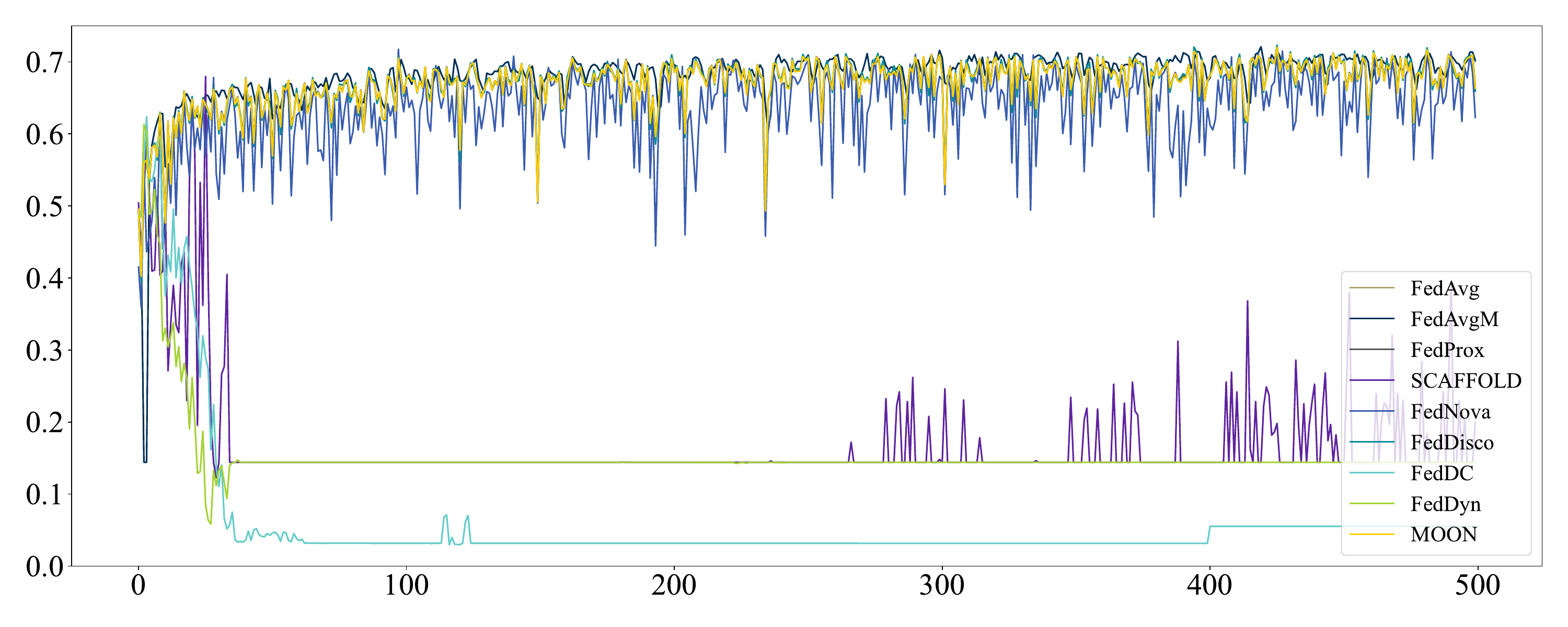} 
        \caption{Accuracy curve of pre-trained ResNet50 on FedRS NIID-2 partition. Testing is performed on $T_I$.}
        \label{fig:source}
    \end{subfigure}
    \caption{Test accuracy curve of FL baseline method during the communication round of training.}
    \label{fig:curve-pretrained} 
\end{figure}

Figure~\ref{fig:curve-pretrained} shows the accuracy curves of using partial parameters updates on pre-trained models. For most Fl algorithms, convergence is achieved in fewer rounds than with full parameter updates, and the best accuracy is comparable to that of full parameter updates. This fully demonstrates the advantage of using public pre-trained models in FL training. At the same time, it can be seen that for SCAFFOLD, FedDC, and FedDyn, their instability still exists.

\end{document}